\documentclass[10pt,logo,copyright]{nvidiatechreport}
\usepackage[authoryear,sort&compress,round]{natbib}

\usepackage[utf8]{inputenc} 
\usepackage[T1]{fontenc}    

\usepackage{parskip}        
\usepackage{url}            
\usepackage{booktabs}       
\usepackage{amsfonts}       
\usepackage{nicefrac}       
\usepackage{microtype}      
\usepackage{xcolor}         
\usepackage[dvipsnames]{xcolor} 
\usepackage{graphicx}
\usepackage{animate}        
\usepackage{subcaption}
\usepackage{tabularx}
\usepackage{makecell}
\usepackage{adjustbox}
\usepackage{setspace}
\newcolumntype{M}[1]{>{\centering\arraybackslash}m{#1}}
\usepackage{float}
\usepackage{tikz}
\usetikzlibrary{positioning,shapes,arrows}
\usepackage{amsmath,amsfonts,bm, bbm,leftindex}
\usepackage{multirow}
\usepackage{comment}
\usepackage{gensymb}
\usepackage{lipsum}
\usetikzlibrary{arrows.meta, positioning, fit}
\usepackage[para]{threeparttable}
\usepackage{tikz}
\usepackage{fvextra}
\usepackage[most]{tcolorbox}
\usepackage{adjustbox} 
\usepackage{pgfplots}                                   \usepackage{marginnote} 
\pgfplotsset{compat=1.18}
\usepackage{subcaption}
\usetikzlibrary{tikzmark}

\usepackage[symbol]{footmisc}

\renewcommand{\today}{2026-03-16}

\usepackage[group-separator = {,},  
            group-minimum-digits = 4,  
            group-digits = integer     
           ]{siunitx}

\definecolor{darkred}{rgb}{0.7, 0.0, 0.0}

\definecolor{golden}{RGB}{229, 184, 11}

\usepackage{pifont}

\usepackage[nameinlink]{cleveref}
\crefname{equation}{Eq.}{Eqs.}
\crefname{figure}{Fig.}{Figs.}
\crefname{section}{Sec.}{Sec.}
\crefname{appendix}{App.}{App.}
\crefname{table}{Tab.}{Tabs.}
\crefname{algorithm}{Algo}{Algo}
\crefname{thm}{Thm}{Thm}
\Crefname{thm}{Thm}{Thm}
\crefname{prop}{Prop}{Prop}
\usepackage{xcolor}
\usepackage{tcolorbox}
\usepackage{wrapfig}
\usepackage{twemojis}
\tcbuselibrary{breakable} 
\tcbset{
  colback=gray!5!white,
  colframe=gray!60!black,
  boxrule=0.6pt,
  arc=3pt
}

\newtcolorbox{promptbox}[1][]{%
  title=#1,
  coltitle=black,
  fonttitle=\bfseries,
  colbacktitle=gray!15,
  breakable,
}

\usepackage{listings}
\usepackage{courier}
\usepackage[dvipsnames]{xcolor}

\usepackage{microtype}
\usepackage{tcolorbox}
\usepackage{adjustbox}
\usepackage{threeparttable}
\usepackage{graphicx}
\definecolor{gold}{HTML}{FFD700}

\definecolor{metallicgold}{HTML}{D4AF37}

\newcommand{\crefnames}[3]{%
  \@for\next:=#1\do{%
    \expandafter\crefname\expandafter{\next}{#2}{#3}%
  }%
}

\newcommand{\hflogo}{\raisebox{-0.2\height}{\includegraphics[height=1em]{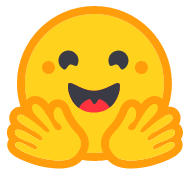}}}

\title{Nemotron-Cascade 2: Post-Training LLMs with Cascade RL and Multi-Domain On-Policy Distillation}

\definecolor{c0}{cmyk}{1,0.3968,0,0.2588} 
\definecolor{c1}{cmyk}{0,0.6175,0.8848,0.1490} 
\definecolor{c2}{cmyk}{0.1127,0.6690,0,0.4431} 
\definecolor{c3}{cmyk}{0.3081,0,0.7209,0.3255} 
\definecolor{nvg}{HTML}{33CC00}
\newtcbox{\hlprimary}{on line,colback=c0!10,colframe=white,size=fbox,arc=3pt, box align=base,before upper=\strut, top=-2pt, bottom=-4pt, left=-1pt, right=-1pt, boxrule=0pt}
\newtcbox{\hlprimarytab}{on line, box align=base, colback=c0!10,colframe=white,size=fbox,arc=3pt, before upper=\strut, top=-2pt, bottom=-4pt, left=-2pt, right=-2pt, boxrule=0pt}
\newtcbox{\hlsecondary}{on line,colback=c1!10,colframe=white,size=fbox,arc=3pt, box align=base,before upper=\strut, top=-2pt, bottom=-4pt, left=-1pt, right=-1pt, boxrule=0pt}
\newtcbox{\hlsecondarytab}{on line, box align=base, colback=c1!10,colframe=white,size=fbox,arc=3pt, before upper=\strut, top=-2pt, bottom=-4pt, left=-2pt, right=-2pt, boxrule=0pt}
\newtcolorbox{hlmultiline}{on line,colback=decentgrey!75,colframe=white,size=fbox,arc=3pt, box align=base, top=0pt, bottom=2pt, boxrule=0pt, before=\adjustbox{valign=c}\bgroup, after=\egroup, before upper=\strut}
\newtcbox{\hlmaintab}{on line, box align=base, colback=nvg!25,colframe=white,size=fbox,arc=3pt, before upper=\strut, top=-2pt, bottom=-4pt, left=-2pt, right=-2pt, boxrule=0pt}

\newcolumntype{Y}{>{\centering\arraybackslash}X}
\newcolumntype{Z}{>{\raggedleft\arraybackslash}X}

\definecolor{c4}{cmyk}{0.6765,0.2017,0,0.0667} 
\definecolor{c5}{cmyk}{0,0.8765,0.7099,0.3647} 

\definecolor{darkgrey}{RGB}{149,149,149}
\definecolor{decentgrey}{RGB}{242,242,242}

\newcommand{\model}{Nemotron-Cascade 2\xspace}

\author{
Zhuolin Yang\footnote[1]{Equal contribution, with authors listed in reverse alphabetical order by first name.},
~Zihan Liu$^*$, ~Yang Chen$^*$,  ~Wenliang Dai$^*$, ~Boxin Wang$^*$,
~Sheng-Chieh Lin,  ~Chankyu Lee,  ~Yangyi Chen, ~Dongfu Jiang, 
~Jiafan He\footnote[3]{Reviewed and scored our model-generated solutions for IMO 2025 as a gold medalist at the IMO 2015. Correspondence to: <jiafanhe19@ucla.edu>.}, 
~Renjie Pi, ~Grace Lam, ~Nayeon Lee,   ~Alexander Bukharin, ~Mohammad Shoeybi,  ~Bryan Catanzaro, ~Wei Ping$^*$\footnote[2]{Leads the effort. Correspondence to: <wping@nvidia.com>.}
}

\begin{abstract}
{\normalfont
We introduce Nemotron-Cascade 2, an open 30B MoE model with 3B activated parameters that delivers best-in-class reasoning and strong agentic capabilities. 
Despite its compact size, its mathematical and coding reasoning performance approaches that of frontier open models. 
It is the second open-weight LLM, after DeepSeek-V3.2-Speciale-671B-A37B, to achieve \boldsymbol{\textcolor{golden}{Gold Medal}}-level performance in the 2025 International Mathematical Olympiad (IMO), the International Olympiad in Informatics (IOI), and the ICPC World Finals, demonstrating remarkably high intelligence density with 20× fewer parameters.
In contrast to Nemotron-Cascade 1, the key technical advancements are as follows.
After SFT on a meticulously curated dataset, we substantially expand Cascade RL to cover a much broader spectrum of reasoning and agentic domains. 
Furthermore, we introduce multi-domain on-policy distillation from the strongest intermediate teacher models for each domain throughout the Cascade RL process, allowing us to efficiently recover benchmark regressions and sustain strong performance gains along the way.
We release the collection of model checkpoint
and training data.
}
\end{abstract}

\begin{document}

\maketitle

\abscontent

\begin{itemize}
    \item[\hflogo] \href{https://huggingface.co/nvidia/Nemotron-Cascade-2-30B-A3B}{Nemotron-Cascade-2-30B-A3B}: the post-trained model based on Nemotron-3-Nano-30B-A3B-Base.
    \item[\hflogo] \href{https://huggingface.co/datasets/nvidia/Nemotron-Cascade-2-SFT-Data}{Nemotron-Cascade-2-SFT-Data}: collection of SFT datasets for Nemotron-Cascade-2.
    \item[\hflogo] \href{https://huggingface.co/datasets/nvidia/Nemotron-Cascade-2-RL-Data}{Nemotron-Cascade-2-RL-Data}: collection of RL datasets for Nemotron-Cascade-2.
\end{itemize}

\begin{figure}[h]
\centerline{\includegraphics[width=1.05\linewidth]
{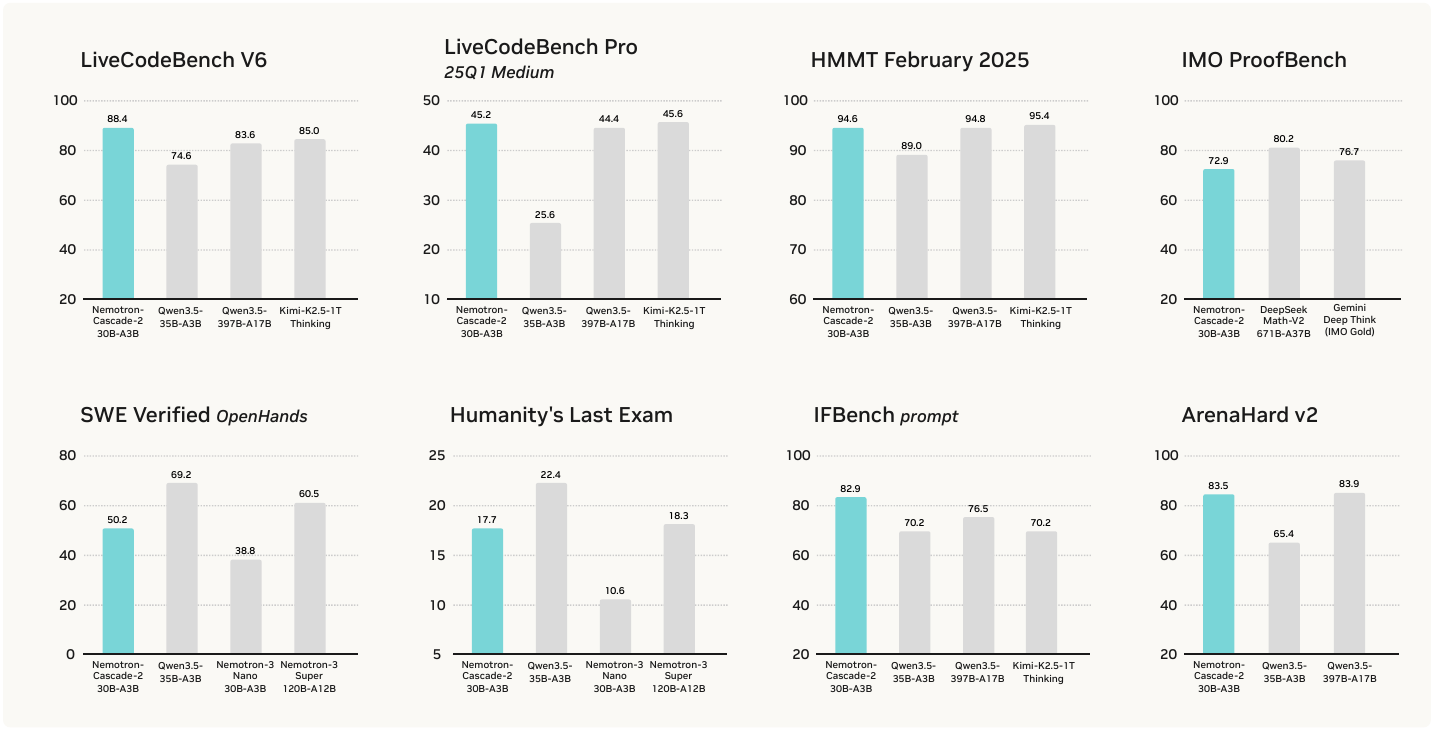}}
\label{fig:first_page}
\vspace{-.19cm}
\end{figure}

\newpage

\tableofcontents

\newpage

\renewcommand{\thefootnote}{\arabic{footnote}}

\section{Introduction}
\label{sec:introduction}

Reinforcement Learning (RL) \citep{ouyang2022training,guo2025deepseek} has emerged as the cornerstone of LLM post-training, driving advances in reasoning, agentic capabilities, and real-world problem-solving. 
As models are tasked with increasingly sophisticated requirements, the primary challenge lies in successfully incorporating a broader array of RL environments and very diverse reasoning and agentic tasks. 
Scaling RL to encompass multifaceted, real-world applications necessitates robust frameworks capable of handling varied reward signals and complex environmental feedback without destabilizing the training process.

 Our previous work, Nemotron-Cascade 1~\citep{wang2025nemotron}, introduced Cascade RL, a framework that orchestrates sequential, domain-wise RL training across specialized task domains. Cascade RL significantly simplifies the engineering complexity associated with multi-domain RL while achieving state-of-the-art performance across a wide range of benchmarks. The advantages of Cascade RL are threefold.
First, domain-specific RL stages are remarkably resistant to \emph{catastrophic forgetting}. They rarely degrade benchmark performance attained in earlier domains and may even improve it.
Second, it allows RL hyperparameters and the training curriculum to be carefully tailored to each specific domain, enabling optimized learning dynamics and improved final performance.
Third, task homogeneity within each RL stage also yields substantial compute savings, as response lengths and verification wall-clock times are more uniform within a domain than across multiple domains trained jointly.

In this work, we introduce Nemotron-Cascade 2, an open 30B Mixture-of-Experts (MoE) model with 3B activated parameters.
Similar to its predecessor, Nemotron-Cascade 2 further scales Cascade RL on high-priority domains to preserve the benefits of domain-wise training,  enabling us to push the limits of reasoning performance in key domains to state-of-the-art levels.
Furthermore, we incorporate on-policy distillation~\citep{xiao2026mimo, zeng2026glm} into Cascade RL training stages.
By distilling knowledge from the best-performing intermediate teacher models within each specific domain during Cascade RL, this mechanism effectively recovers any benchmark regressions that can occur when training in increasingly complex RL environments.
In addition, we integrate multi-domain RL into Cascade RL for groups of tasks with similar response formats and comparable verification costs, allowing them to be trained jointly to scale up for more RL environments and improve training efficiency when cross-task interference is minimal.

Our Nemotron-Cascade-2-30B-A3B achieves breakthrough performance in mathematical and coding reasoning, securing gold-medal results in both the 2025 International Mathematical Olympiad (IMO) and the International Olympiad in Informatics (IOI) despite being only a 30B MoE model,~\footnote{Our model is the second open-weight LLM, after DeepSeek-V3.2-Speciale-671B-A37B~\citep{liu2025deepseek}, to achieve gold-medal performance in both the IMO and IOI.}
while also delivering best-in-class performance across a broad range of benchmarks, including alignment, instruction-following, long context~(e.g., 1M context window), and agentic tasks. See Table~\ref{tab:main_results} for the full results.
We fully open source the model weights, training data, and methodological details, enabling the research community to reproduce, analyze, and extend the proposed Cascade RL training paradigm.

We organize the remainder of this report as follows. Section \S\ref{sec:main_results} summarizes the main results. Section \S\ref{sec:sft} describes the supervised fine-tuning~(SFT) with details on data curation. 
Section \S\ref{sec:rl} presents Cascade RL framework intergrated with the multi-domain on-policy distillation. 
Section~\S\ref{sec:imo} details the evaluation setup and results on IMO, while Section~\S\ref{sec:ioi} presents the evaluation setup and results on IOI and the ICPC World Finals.

\section{Main Results}
\label{sec:main_results}

\begin{table}[t!]
\centering
\footnotesize
\renewcommand{\arraystretch}{1.15}
\caption{\textbf{Main results}. 
Nemotron-Cascade-2-30B-A3B achieves gold-medal performance in both the IMO 2025 and IOI 2025, which demonstrate remarkably high intelligence density. 
\dag~Numbers in brackets refers to Tool-Integrated Reasoning (TIR) results.
$^\ddag$ For the baseline models, we use official numbers when available, otherwise evaluate them using the recommended settings.
}
\label{tab:main_results}
\begin{adjustbox}{width={1.0\textwidth}}
\begin{tabular}{lccc|c}
\toprule
\shortstack[l]{\textbf{Benchmark}\\\textbf{}}
 & \shortstack{\textbf{Nemotron-3-Nano}\\\textbf{30B-A3B}}
 & \shortstack{\textbf{Nemotron-3-Super}\\\textbf{120B-A12B}}
  & \shortstack{\textbf{Qwen3.5}\\\textbf{35B-A3B}}
 & \shortstack{\textbf{Nemotron-Cascade-2}\\\textbf{30B-A3B}} \\
\midrule
\multicolumn{5}{l}{\textbf{Math}} \\
IMO 2025                     & -- & -- & -- & \raisebox{-0.6ex}{\includegraphics[height=2.8ex]{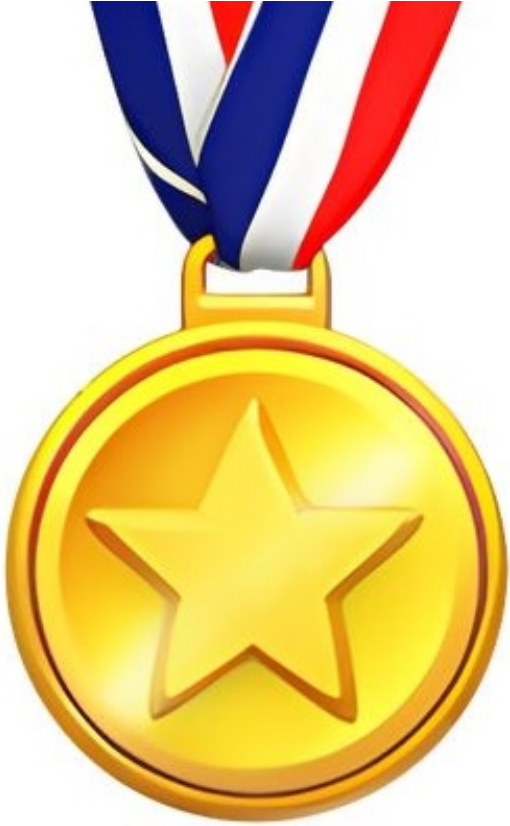}} \textbf{35 pts}  \\
IMO AnswerBench             & \hspace{1.35mm}70.4$^\ddag$ & \hspace{1.35mm}77.2$^\ddag$ & \hspace{1.35mm}74.8$^\ddag$ & 79.3 \\
IMO ProofBench              & -- & -- & -- & 72.9 \\
AIME 2025 & 89.1 & 90.2 & \hspace{1.35mm}91.9$^\ddag$ & 92.4 (98.6)\dag \\
AIME 2026 & \hspace{1.35mm}89.9$^\ddag$ & \hspace{1.35mm}89.8$^\ddag$  & \hspace{1.35mm}91.1$^\ddag$ & 90.9 (95.0)\dag \\
HMMT Feb25 & \hspace{1.35mm}84.6$^\ddag$ & 93.7 & 89.0 & 94.6 \\
\midrule
\multicolumn{5}{l}{\textbf{Code Reasoning}} \\
IOI 2025                    & -- & -- & \hspace{1.35mm}348.6$^\ddag$ & \raisebox{-0.6ex}{\includegraphics[height=2.8ex]{figures/gold_medal_emoji.pdf}} \textbf{439.28} \\
ICPC World Finals 2025                 & -- & -- & -- & \raisebox{-0.6ex}{\includegraphics[height=2.8ex]{figures/gold_medal_emoji.pdf}} \bf 10/12 \\
LiveCodeBench v6 (2408-2505)        & 68.3  & 78.7  & 74.6  & 87.2 (88.4)\dag \\
LiveCodeBenchPro 25Q2 (Easy)                   & \hspace{1.35mm}54.5$^\ddag$ & \hspace{1.35mm}81.7$^\ddag$ & \hspace{1.35mm}81.1$^\ddag$ & 87.0 (89.3)\dag  \\
LiveCodeBenchPro 25Q2 (Med)          & \hspace{1.35mm}3.50$^\ddag$ & \hspace{1.35mm}23.2$^\ddag$ & \hspace{1.35mm}17.8$^\ddag$ & 27.6 (36.8)\dag  \\
SciCode                & 33.3 & 42.1 & 38.0 & 36.4 \\
\midrule
\multicolumn{5}{l}{\textbf{Knowledge \& STEM}} \\
MMLU-Redux              & -- & -- & 93.3 & 86.3 \\
MMLU-Pro                & 78.3 & 83.7 & 85.3 & 79.8 \\
GPQA-Diamond           & 73.0 & 79.2 & 84.2  & 76.1  \\
HLE (no tool)          & 10.6 & 18.3 & 22.4 & 17.7 \\
\midrule
\multicolumn{5}{l}{\textbf{Alignment \& Instruction Following}} \\
ArenaHard v2 (Avg.)        & 67.7 & -- & \hspace{1.35mm}65.4$^\ddag$  & 83.5 \\
\ \ -- Hard Prompt        & 72.1 & 73.9 & \hspace{1.35mm}64.5$^\ddag$  & 88.2 \\
\ \ -- Creative Writing        & 63.2 & -- & \hspace{1.35mm}66.3$^\ddag$  & 78.7 \\
IFBench (prompt)    & 71.5 & 72.6 &  70.2 & 82.9 \\
Scale AI Multi-Challenge & 38.5 & 55.2 & 60.0 & 45.3 \\
\midrule
\multicolumn{5}{l}{\textbf{Long Context \& Context Learning}} \\
AA-LCR                      & 35.9 & 58.3 & 58.5 & 39.1 \\
LongBench v2                & 39.6 & -- & 59.0 & 40.3 \\
NIAH@1M (RULER Subset)                        & 94.8 & 98.3 & \hspace{1.35mm}94.3$^\ddag$ & 99.0 \\
CL-Bench                        & \hspace{1.35mm}12.0$^\ddag$ & -- & \hspace{1.35mm}15.5$^\ddag$  & 12.2 \\ 
\midrule
\multicolumn{5}{l}{\textbf{Agentic}} \\
BFCL v4                        & 53.8 & -- & 67.3 & 52.9 \\
$\tau^2$-Bench                   & 49.0 & 61.2 & 81.2 & 58.9 \\
Terminal Bench 2.0             & 8.5 & 31.0 & 40.5 & 21.1 \\
SWE Verified~(OpenHands)       & 38.8 & 60.5 & 69.2 & 50.2 \\
\midrule
\multicolumn{5}{l}{\textbf{Multilingual}} \\
MMLU-ProX                       & 59.5 & 79.4 & 81.0  & 72.5 \\ 
WMT24++ (en -> xx)              & 86.2 & 86.7 & \hspace{1.35mm}87.6$^\ddag$  & 84.1 \\ 
\bottomrule
\end{tabular}
\end{adjustbox}
\end{table}

\begin{table}[h!]
\centering
\begin{minipage}{\textwidth} 
\centering
\caption{Performance of Nemotron-Cascade-2-30B-A3B model on IMO 2025, IOI 2025, and ICPC World Finals 2025 competitions. Nemotron-Cascade-2 model achieved solid gold medal on all these top-tier competitions. Our IMO 2025 solutions are evaluated by human expert (IMO 2015 Gold medalist) while IOI 2025 and ICPCWF 2025 solutions are verified through OnlineJudge with official testcases.}
\begin{tabular}{lcccccccc}
\toprule
 \textbf{Competition} & P1 & P2 & P3 & P4 & P5 & P6 & Overall & Medal\\
 \midrule
IMO 2025 & 7 & \textcolor{gray}{7$^{\dag}$} & 7 & 7 & 7 & 0 & 35/42 & \textcolor{metallicgold}{Gold} \\
IOI 2025 & 39 & 88.53 & 100 & 100 & 28.75 & 83 & 439.28/600 & \textcolor{metallicgold}{Gold} \\
\bottomrule
\end{tabular}
\\[2ex]
\begin{tabular}{lcccccccccccccc}
\toprule
 \textbf{Competition} & A & B & C & D & E & F & G & H & I & J & K & L & Overall & Medal\\
 \midrule
ICPC World Finals 2025 & \color{ForestGreen} \bf + & \color{BrickRed} \bf - & \color{ForestGreen}\bf + & \color{ForestGreen}\bf + &  \color{ForestGreen}\bf + & \color{ForestGreen} \bf+ & \color{BrickRed} \bf - &\color{ForestGreen} \bf + & \color{ForestGreen} \bf + & \color{ForestGreen}\bf + & \color{ForestGreen} \bf+ & \color{ForestGreen}\bf + & 10/12 & \textcolor{metallicgold}{Gold} \\
\bottomrule
\end{tabular}
\vspace{0.5em}
\begin{flushleft}
\small
\setlength{\hangindent}{1em} 
$^{\dag}$ For IMO 2025 P2, we use LLM grader with reference solution and marking schema from ProofBench~\citep{ma2025reliable} due to the extensive analytic geometry approach of the model, which human expert could be hard to verify all the intermediate derivation steps.
\end{flushleft}
\label{tab:main_IMO_IOI_ICPC_2025}
\end{minipage}
\end{table}

We evaluate Nemotron-Cascade 2 on a comprehensive suite of benchmarks covering mathematical and coding reasoning, knowledge and STEM, alignment and instruction following, long-context understanding and in-context learning, multilingual capabilities, and agentic tasks.
The main results are shown in Table~\ref{tab:main_results}, and the benchmarks and detailed evaluation setups are described in Appendix~\ref{appendix:benchmarks}.

From Table~\ref{tab:main_results}, Nemotron-Cascade-2-30B-A3B outperforms both the latest released Qwen3.5-35B-A3B~(2026-02-24)~\citep{qwen3.5} and the larger Nemotron-3-Super-120B-A12B~(2026-03-11)~\citep{blakeman2025nvidia}, and achieves best-in-class performance across benchmarks in mathematics, code reasoning, general alignment, and instruction following.

Notably, despite being only a 30B MoE model, Nemotron-Cascade 2 achieves \textbf{gold-medal performance on IMO 2025, IOI 2025 and ICPC World Finals 2025}, results previously thought to be attainable only by frontier proprietary models~\citep{googledeepmind2025imogold} (i.e., Gemini Deep Think) and frontier-sized open models~\citep{liu2025deepseek} (i.e., DeepSeek-V3.2-Speciale-671B-A37B).
The detailed performance of our model is reported in Table~\ref{tab:main_IMO_IOI_ICPC_2025}.
For IMO 2025, the model-generated solutions and the human expert reviews and scores are provided in Appendix~\ref{appendix:imo2025-model-solutions}.

Nemotron-Cascade-2-30B-A3B also outperforms Nemotron-3-Nano-30B-A3B on nearly all benchmarks, even though both models are post-trained from the same pretrained model, Nemotron-3-Nano-30B-A3B-Base~\citep{nano_v3_base}.
This result further demonstrates the effectiveness of our Cascade RL plus MOPD training pipeline.
It underperforms Qwen3.5-35B-A3B on knowledge-intensive and agentic benchmarks, highlighting the importance of stronger knowledge-intensive pretraining and agentic RL in future work.

\section{Supervised Fine-Tuning}
\label{sec:sft}

In this section, we describe the training framework and data curation process for supervised fine-tuning (SFT), the first stage of our post-training pipeline. This stage equips the model with foundational capabilities, including reasoning, conversational ability, instruction following, and agentic and software engineering skills.

\subsection{Training Framework}

\subsubsection{Overview}
Our SFT data spans a broad range of domains, including mathematics, coding, science, tool use, agentic tasks, and software engineering, as well as more general domains such as multi-turn dialogue, knowledge-intensive question answering, creative writing, role-playing, safety, and instruction following.

We pack all SFT samples into sequences of up to 256K tokens and train the model in a single stage. Empirically, we find that the SFT model reaches optimal performance after approximately 1.5 epochs.
The SFT training hyperparameters can be found in Appendix~\ref{appendix:training_hyperparams}.

\subsubsection{Chat Template}

\begin{figure}[t!]
\centerline{\includegraphics[width=1.0\linewidth]{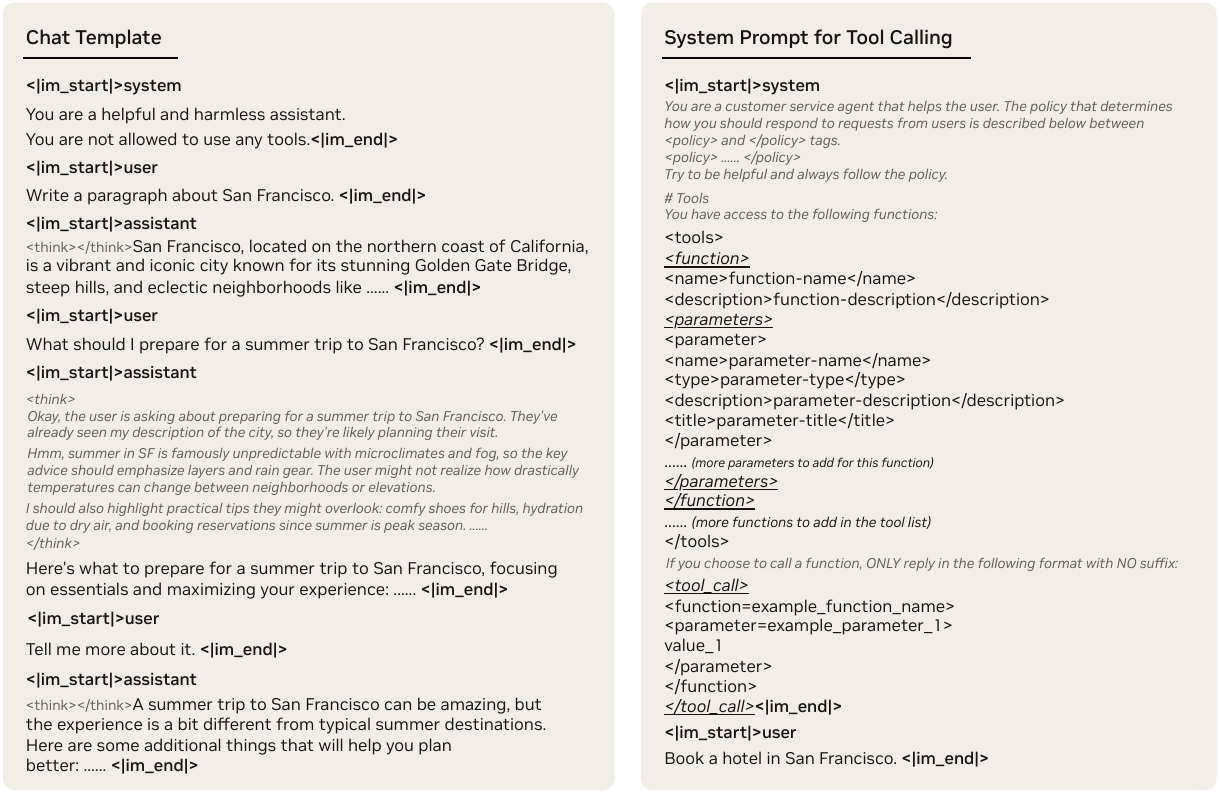}}
\caption{(Left)~The chat template uses adjacent \emph{<think></think>} tokens to indicate non-thinking mode, and a single \emph{<think>} followed by \emph{\textbackslash n} to indicate thinking mode.
(Right)~For tool calling, the available tools are listed in the system prompt. The model is instructed to call tools within the \emph{<tool\_call>} and \emph{</tool\_call>} tags.
}
\label{fig:chat_templates}
\end{figure}

Our chat template is depicted in Figure~\ref{fig:chat_templates}. 
There are two changes to the chat template compared with Nemotron-Cascade~\citep{wang2025nemotron}. First, we remove the \texttt{/think} and \texttt{/no\_think} tags for simplicity. Second, we prepend an empty \texttt{<think></think>} block to activate the non-thinking mode.

For tool calling task, we specify all available tools in the system prompt within the \texttt{<tools>} and \texttt{</tools>} tags, and instruct the model to perform tool calls wrapped within the \texttt{<tool\_call>} and \texttt{</tool\_call>} tags.

\subsection{SFT Data Curation}

\subsubsection{Math}


Our non-proof math prompts are primarily sourced from Nemotron-Cascade~\citep{wang2025nemotron} and Nemotron-Math-v2~\citep{du2025nemotron}, from which we collect 1.8M tool calling (i.e., python) samples and 1.9M non-tool samples, with responses generated by DeepSeek-V3.2 and DeepSeek-V3.2-Speciale~\citep{liu2025deepseek}, respectively.
In addition, we collect 676K samples from the generation-selection category (without tool calling) of Nemotron-3-Nano~\citep{blakeman2025nemotron}, with responses generated by GPT-OSS-120B~\citep{agarwal2025gpt}. 
In total, the competition math SFT comprises 1.8M tool-calling samples and 2.6M samples without tool use.

For mathematical natural language proof, we collect 98K mathematical proof problems from the AOPS split of Nemotron-Math-Proofs-v1~\citep{du2025nemotron}. We generate multiple samples per problem to cover two capabilities including proof generation (410K) and proof verification (400K) using DeepSeek-V3.2-Speciale~\citep{liu2025deepseek}, resulting in a total of 816K samples.

\subsubsection{Code Reasoning}
Built on Nemotron-Cascade 1~\citep{wang2025nemotron},
we curate approximately 165K unique coding prompts from several open-source datasets, including OpenCode-Stage2~\citep{huang2024opencoder}, OpenCodeReasoning~\citep{ahmad2025opencodereasoning}, and HardTests~\citep{he2025hardtests}. These prompts are originally sourced from competitive programming platforms such as Codeforces, AtCoder, AIZU, and CodeChef. To encourage prompt diversity and reduce redundancy in our SFT training set, we apply strict deduplication using two methods: (1) sample I/O fingerprinting and (2) n-gram-based text analysis. This process removes approximately 24.2\% of self-duplicated coding prompts.

We choose GPT-OSS-120B~\citep{agarwal2025gpt} as our SFT teacher model due to its strong code reasoning capabilities. For each coding prompt with verifiable test cases, we apply correctness filtering to the teacher’s reasoning traces, retaining only those that generate correct code. For prompts without verifiable test cases, we generally select longer reasoning traces under the assumption that they reflect more thorough problem analysis. This pipeline yields a final dataset comprising 1.9M Python reasoning traces, 1.0M C++14 reasoning traces, and 1.3M Python tool-calling reasoning traces for competitive coding.

\textbf{Scientific Coding:}
We further collect scientific research coding prompts spanning the domains of biology, material science, physics, chemistry, and mathematics. The responses to these prompts are generated by GPT-OSS-120B~\citep{agarwal2025gpt}, resulting in a total of 1.1M SFT samples.

\subsubsection{Science}
The science prompts we collect span physics, chemistry, and biology. We use 1.4M science SFT samples from Nemotron-Cascade~\citep{wang2025nemotron} and an additional 1.3M samples from Nemotron-3-Nano~\citep{blakeman2025nemotron}. Responses in both datasets are generated by GPT-OSS-120B~\citep{agarwal2025gpt}.


\subsubsection{Long Context}
We adopt the 160K long context SFT data from Nemotron-3-Nano~\citep{blakeman2025nemotron}, which has an average sequence length of 128K tokens. In addition, we collect another 74K long context SFT from ChatQA-2~\citep{xu2024chatqa}, which has an average length of 29K tokens.

\subsubsection{General Chat}
We source prompts from Nemotron-Cascade 1~\citep{wang2025nemotron} and construct 4.9M reasoning-on and 372K reasoning-off samples. Responses for reasoning-on samples are generated by GPT-OSS-120B~\citep{agarwal2025gpt}. For reasoning-off samples, 300K responses are drawn from high-quality annotated short answers within the dataset itself, while an additional 330K are generated by DeepSeek-V3-0324~\citep{liu2024deepseek} to improve response quality.

To enhance multi-turn dialogue capabilities, we synthesize approximately 700K multi-turn conversation samples using two GPT-OSS-120B~\citep{agarwal2025gpt} instances in a role-playing setup, where one instance plays the user and the other the assistant. The user-side model may terminate the conversation at any point to prevent repetitive exchanges.

We additionally incorporate 4.6M reasoning-on chat samples from Nemotron-3-Nano~\citep{blakeman2025nemotron}, with prompts drawn from LMSYS~\citep{zheng2023lmsyschat1m} and WildChat~\citep{zhao2024wildchat}. Responses are generated by GPT-OSS-120B~\citep{agarwal2025gpt}, Qwen3-235B-A22B-Thinking-2507, and Qwen3-235B-A22B-Instruct-2507~\citep{yang2025qwen3}.

\subsubsection{Instruction Following}
We source prompts from Nemotron-Cascade 1~\citep{wang2025nemotron} and generate approximately 230K reasoning-on responses using GPT-OSS-120B~\citep{agarwal2025gpt} and 64K reasoning-off responses using DeepSeek-V3-0324~\citep{liu2024deepseek}.
In addition, we incorporate 497K instruction-following samples from Nemotron-3-Nano~\citep{blakeman2025nemotron}, including 457K reasoning-on and 40K reasoning-off responses. These responses are generated by GPT-OSS-120B~\citep{agarwal2025gpt}, Qwen3-235B-A22B-Thinking-2507, and Qwen3-235B-A22B-Instruct-2507~\citep{yang2025qwen3}.

\subsubsection{Safety}
We collect 4K safety SFT samples from Nemotron-3-Nano~\citep{blakeman2025nemotron} to enable models to exhibit appropriate refusal behavior when encountering unsafe inputs. The SFT prompts are originally sourced from Nemotron Content Safety v2~\citep{ghosh2025aegis2}, Gretel Safety Alignment v1~\citep{gretelai_gretel_safety_alignment}, Harmful Tasks~\citep{hasan2024pruning}, and Red-Team-2K~\citep{luo2024jailbreakv}.

\subsubsection{Conversational Agent}
Aside from the Python tool-use data for math and code reasoning, we further gather tool-use samples in multi-turn conversational settings, where multiple tools are available and the assistant must determine which tools to invoke and how to use them effectively.
We collect 822K conversational tool-use samples from Nemotron-3-Nano~\citep{blakeman2025nemotron}, with responses generated by Qwen3-235B-A22B-Thinking-2507, Qwen3-32B, Qwen3-235B-A22B-Instruct-2507~\citep{yang2025qwen3}, and GPT-OSS-120B~\citep{agarwal2025gpt}.

\subsubsection{Software Engineering Agent}
We curate the software engineering (SWE) data using various agentic scaffolds, including OpenHands~\citep{wang2025openhands}, SWE-Agent~\citep{NEURIPS2024_5a7c9475}, Mini-SWE-Agent, and the agentless scaffold proposed by \citet{wei2025swe}, to enhance the models' agentic software engineering  capabilities.
First, we utilize the data from Nemotron 3 Nano~\citep{blakeman2025nemotron} and Super~\citep{blakeman2025nvidia}, which includes SWE agentic trajectories generated using Qwen3-Coder-480B-A35B-Instruct~\citep{yang2025qwen3}. 
The problem instances are drawn from SWE-Gym~\citep{training-software}, SWE-rebench~\citep{badertdinov2025swe}, and R2E-Subset~\citep{jain2025r2e}.
Second, we employ SWE agentless data from Nemotron-Cascade 1~\citep{wang2025nemotron}, which includes three main tasks: (1) buggy code localization, (2) code repair, and (3) test case generation. Following the established procedure in \citet{wang2025nemotron}, we reconstruct the code repair data using DeepSeek-V3.2~\citep{liu2025deepseek}.

Our preliminary study shows that incorporating SWE agentless data improves models’ effectiveness on SWE agentic tasks. 
For example, fine-tuning solely on agentic data achieves Pass@1 of 48.9 and Pass@4 of 62.8, whereas fine-tuning on a combination of agentic and agentless data improves performance to Pass@1 of 49.9 and Pass@4 of 65.2 on SWE-bench Verified using OpenHands.
Based on this observation, we combine 125K agentic samples and 389K agentless samples as the supervised fine-tuning (SFT) data for SWE tasks. Our models are trained in non-thinking mode on SWE agentic data and in thinking mode on SWE agentless data.

\subsubsection{Terminal Agent}
To enhance agentic capabilities for terminal use, we adopt the \textit{Terminal-Task-Gen} methodology \citep{pi2026dataengineeringscalingllm} to curate our training tasks. This framework consists of (1) dataset adapters that transform static data into interactive terminal formats, and (2) synthetic tasks generated from both diverse seed prompts and a structured terminal skill taxonomy. Using this framework, we curate 490K samples in total. Specifically, we first adapt 162K math, 32K code, and 32K SWE-specific samples from existing high-quality sources \citep{wang2025nemotron}, which establishes broad foundational coverage. To further improve targeted skill refinement, we synthesize 120K seed-based and 140K skill-based tasks. For trajectory construction, we leverage the tasks curated from above, and employ DeepSeek-V3.2~\citep{liu2025deepseek} as the core engine to generate step-by-step solution traces via an execution-feedback loop within isolated Docker environments. The Terminus 2 agent framework \citep{merrill2026terminalbenchbenchmarkingagentshard} serves as the underlying scaffolding and tool-use protocol, enabling the model to interact with the terminal and complete complex tasks.

\begin{figure}[t]
    \centering
    \includegraphics[width=\linewidth]{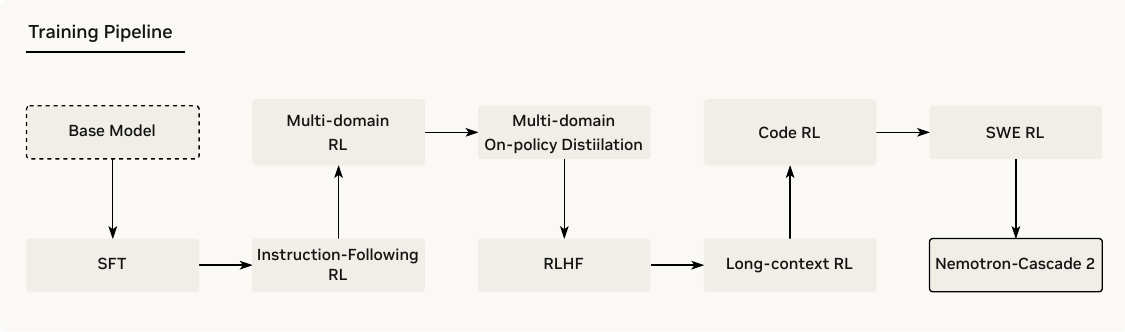}
    \caption{\model applies Cascade RL with the sequential, domain-wise ordering after SFT, leading to substantial
improvements across the corresponding domains.}
    \label{fig:pipeline}
\end{figure}

\section{Cascade RL and Multi-Domain On-Policy Distillation}
\label{sec:rl}

Following a similar approach to Nemotron-Cascade 1 \citep{wang2025nemotron}, we apply Cascaded Reinforcement Learning (Cascade RL) as our post training pipeline. 
In particular, we integrated the Multi-Domain On-Policy Distillation~(MOPD) along the Cascade RL process.

\subsection{Training Framework}

We illustrate our training process in Figure \ref{fig:pipeline}.
In this work, we start the Cascade RL process with IF-RL (\S \ref{sec:ifrl}) to establish foundational instruction adherence, followed by multi-domain RL (\S \ref{sec:workbench}) to enhance the model's tool-calling capabilities, STEM reasoning, and response format adherence. 
We then transition to Multi-domain On-policy Distillation (\S \ref{sec:mopd}) to unify specialized expertise into a single, cohesive policy to mitigate performance degradation. 
We continue with RLHF (\S \ref{sec:rlhf}) for human alignment, Long-context RL (\S \ref{sec:lcrl}) to enhance reasoning over massive input sequences, Code RL (\S \ref{sec:coderl}) for competitive coding problems, and finally SWE RL (\S\ref{sec:swerl}) for mastering agentic software interactions.

\subsubsection{What determines the ordering of Cascade RL}

The optimal ordering of stages within a Cascade RL pipeline is not a universal constant; rather, it is a dynamic function of the model's underlying behaviors and learning trajectories. 
In contrast to the original Nemotron Cascade \citep{wang2025nemotron}, our current work \model introduces significant improvements in SFT data quality and substantially scales the complexity of the RL environments and tasks. 
These advancements have fundamentally altered the model's behavioral dynamics, which require us to adopt a different order to better accommodate the evolving capabilities of LLMs.

\textbf{Rule of thumb: Mitigating Inter-Domain Interference.} Specifically, the rationale for this ordering is primarily driven by the need to mitigate catastrophic forgetting as the model interacts with increasingly diverse environments. 
Cascade RL provides a granular lens through which we can observe how specific domains compete or conflict, such as strict instruction adherence in IF-RL versus human preference alignment in RLHF. 
Our core design principle is to identify an ordering that minimizes negative interference across domains while thoroughly optimizing the highest-priority domains.
By identifying which tasks serve as foundational priors and which act as specialized refinements, we can mitigate inter-domain interference.

\textbf{Scaling via Multi-Domain Integration.} 
Following this principle, the Cascade RL pipeline can incorporate multi-domain RL stages when specific domains are found to be non-conflicting or beneficial to the overall performance. 
This integrated approach is particularly effective as RL environments and datasets grow in complexity, while ensuring that the model maintains a broad performance profile across various benchmarks, as detailed in \S\ref{sec:workbench}.

\textbf{Stabilization through On-policy Distillation.} Furthermore, We find that Multi-domain On-policy Distillation (\S\ref{sec:mopd}) serves as a critical stabilization point in this ordering. It is effective at recovering benchmark performance that may have regressed during earlier, more specialized stages of the cascade RL, leading to a more balanced and robust final policy model.

\subsubsection{RL Training Configuration}
\label{subsec:rl_config}
Throughout the entire Cascade RL process, we use Group Relative Policy Optimization (GRPO) algorithm~\citep{shao2024deepseekmath} with strict \textbf{on-policy} training following Nemotron Cascade~\citep{wang2025nemotron}.
We adopt on-policy training for improved stability and higher accuracy.
We conduct our training using the Nemo-RL repository~\citep{nemo-rl}.

At each iteration, we generate a group of $G$ rollouts from the current policy $\pi_{\theta}$ and then perform a \emph{single} gradient update. This ensures that the policy used for data collection always matches the one being updated, making the importance sampling ratio exactly 1. This on-policy setup contributes to stable RL training and mitigates entropy collapse. In addition, we {remove KL divergence} term entirely, which simplifies the GRPO objective to the standard \emph{REINFORCE} objective~\citep{williams1992simple} with group-normalized rewards and {token-level} loss~\citep{yu2025dapo}:
%
\begin{align}
\mathcal{J}_\text{GRPO}(\theta)& = \mathbb{E}_{(q,a)\sim \mathcal{D},~ \{o_i\}_{i=1}^G\sim \pi_{\theta}(\cdot\mid q)}
\Bigg[
\frac{1}{\sum_{i=1}^{G}|o_i|}\sum_{i=1}^{G}\sum_{t=1}^{|o_i|}
\hat{A}_{i,t}
\Bigg], 
~\text{where}~
    \hat{A}_{i,t} = \frac{r_i - \text{mean}(\{r_i\}_{i=1}^G)}{\text{std}(\{r_i\}_{i=1}^G)} ~\text{for all } t,
\label{equation:grpo_objective}
\end{align}
%
and $\{r_i\}_{i=1}^{G}$ denotes the group of G rewards assigned to the sampled responses $\{o\}_{i=1}^G$ for a given question $q$ drawn from the dataset $\mathcal{D}$, verified against the ground-truth answer $a$ in RLVR.
For RLHF, $r_i$ is the aggregated reward score from the generative reward model for response $o_i$ and question $q$.
Details of the reward functions for different domains will be provided in the corresponding subsections.

\subsection{Instruction-Following Reinforcement Learning (IF-RL)}
\label{sec:ifrl}

In this subsection, we describe our instruction-following RL recipe, which serves as the first stage of our Cascade RL.
We demonstrate that applying verifiable IF-RL significantly improves instruction adherence, achieving a state-of-the-art accuracy of 83.13\% on IFBench~\citep{pyatkin2025generalizingverifiableinstructionfollowing}.

\subsubsection{Dataset}
We use the same instruction-following training data used for NVIDIA Nano-v3 post-training \citep{blakeman2025nemotron}.
The instructions in this dataset are designed for objective verifiability, for instance, requiring a response to be under 200 words.
This making the dataset well-suited for training and evaluating models on strict adherence. 
Given the high baseline quality of the data, our curation process mainly resolves formatting inconsistencies within the keyword arguments for certain instruction types (e.g., \texttt{count\_increment\_word}).

\subsubsection{Training recipe}
Following \citep{wang2025nemotron}, we also apply dynamic filtering \citep{yu2025dapo}.
This technique filters out samples where all rollouts are either entirely correct or entirely incorrect. 
By ensuring that every prompt in a batch provides effective gradients, dynamic filtering stabilizes IF-RL training and pushes the upper bound of model performance.
Furthermore, we observed that extended IF-RL training can lead to excessive token usage, which is often unnecessary for fulfilling specific constraints in general chat domains. 
To mitigate this, we apply overlong penalty, which penalizes samples that fail to complete generation within the maximum sequence length with a zero reward.

Unlike Nemotron Cascade \citep{wang2025nemotron}, we position IF-RL as the first stage of our Cascade RL training for two primary reasons:
(i) IF-RL can negatively impact human alignment capabilities (e.g., ArenaHard), while our subsequent generative-reward-model-based RLHF has a negligible impact on instruction following scores. 
By prioritizing instruction adherence first, we can focus on maximizing instruction following performance and then utilize the later stages to recover and refine human preference alignment.
(ii) An early IF-RL stage produces a model with superior instruction-following capabilities, which serves as a strong teacher for subsequent multi-domain on-policy distillation.
Another difference from Nemotron Cascade \citep{wang2025nemotron} is that our IF-RL is trained exclusively in ``thinking mode'' without incorporating a reward model. 
We found that the ``thinking mode'' yields higher accuracy on instruction-following benchmarks (e.g., IFBench \citep{pyatkin2025generalizingverifiableinstructionfollowing}). 
Because subsequent RL stages recover any regressions in human preference alignment introduced during IF-RL, we can focus entirely on maximizing instruction adherence without incurring the computational overhead of an auxiliary reward model.

We use a batch size of 128, sampling 16 responses per prompt with temperature
1.0 and top-p 1.0.
We adopt a learning rate of 3e-6 with AdamW \citep{kingma2014adam}, and set both the
entropy loss coefficient and KL loss coefficient to 0. Our IF-RL with dynamic filtering takes around 180 steps. 
The full set of hyperparameters is provided in Appendix~\ref{appendix:training_hyperparams}.

\subsection{Multi-domain RL}
\label{sec:workbench}
Following IF-RL, we conduct an additional stage of multi-domain RL that covers three capabilities: multi-choice question answering (MCQA) in the STEM domain, agentic tool calling, and structured output for instruction following. The datasets are drawn from the NVIDIA Nano-v3 RL training blend \citep{blakeman2025nemotron}. The data mixture consists of approximately 55\% MCQA, 30\% agentic tool calling using the Workplace Assistant setup \citep{blakeman2025nemotron}, and 15\% structured output.

We group these domains into a single multi-domain RL stage for two main reasons.
First, we do not observe performance degradation across evaluation benchmarks when training on the blended domains. Instead, the model exhibits consistent improvements on benchmarks including MMLU-Pro, $\tau^2$-Bench, and IF-Bench.
Second, the response lengths and verification times of these datasets are similar, which minimizes training inefficiencies caused by waiting for longer generations or slower environment verification.

During training, we use a batch size of 128 and sample 16 responses per prompt with temperature 1.0 and top-p 1.0 (see Appendix~\ref{appendix:training_hyperparams}). 
We adopt a learning rate of $3\times10^{-6}$ with AdamW \citep{kingma2014adam}, and set both the entropy loss coefficient and KL loss coefficient to zero. This multi-domain RL stage runs for approximately 70 training steps.

\subsection{Multi-domain On-Policy Distillation (MOPD)}
\label{sec:mopd}
While well-designed Cascade RL substantially reduces catastrophic forgetting compared with vanilla sequential RL in an arbitrary order, it does not fully eliminate capability drift as the number of training environments increases.
In practice, we observe noticeable fluctuations across different benchmark categories tracked throughout training, and the dominant trade-offs differ by stage. For example, certain RLVR training often reduces model entropy and shortens reasoning traces, thus can negatively impact mathematical reasoning performance, while RLHF-oriented optimization can partially trade off against instruction-following behavior. These observations motivate an additional training stage for re-balancing capabilities within the Cascade RL process.

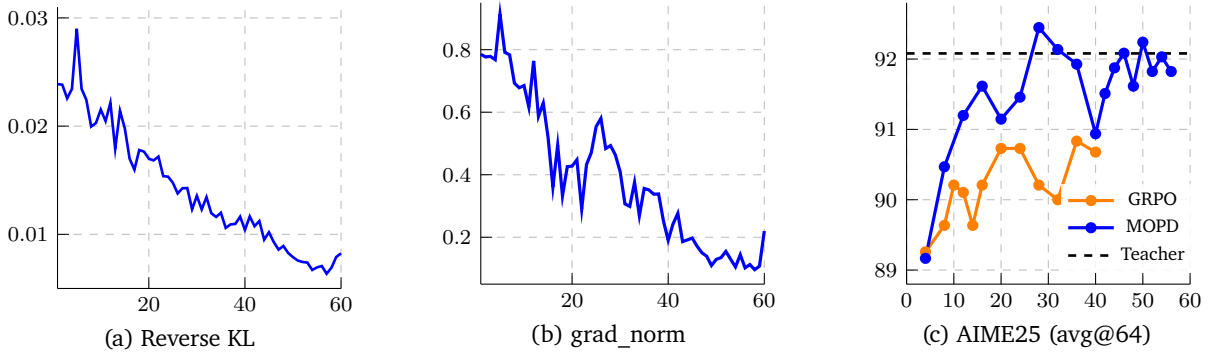
\begin{figure}[t]
\centering

\captionsetup[subfigure]{
    labelformat=simple,
    justification=centering,
    singlelinecheck=true,
    skip=2pt
}
\renewcommand\thesubfigure{(\alph{subfigure})}

\newlength{\panelw}
\setlength{\panelw}{0.32\columnwidth}

\begin{subfigure}[t]{\panelw}
\vspace{0pt}
\centering
\begin{tikzpicture}
\begin{axis}[
    width=\linewidth,
    height=\linewidth,
    ylabel={},
    xmin=1, xmax=60,
    ymax=0.031, ymin=0.005,
    axis line style={black},
    tick style={black},
    xticklabel style={font=\footnotesize},
    xlabel style={font=\normalsize},
    title style={font=\normalsize, yshift=-2pt},
    grid=major,
    grid style={dashed, gray!50},
    enlargelimits=false,
    clip=false,
    axis background/.style={fill=white},
    axis x line*=bottom,
    axis y line*=left,
    ytick={0.03,0.02,0.01},
    scaled y ticks=false,
    yticklabel style={
        font=\footnotesize,
        /pgf/number format/fixed,
        /pgf/number format/precision=2
    },
]
\addplot[blue, line width=1pt, mark=none] coordinates {(1, 0.0238857641816139) (2, 0.0238247364759445) (3, 0.0225688759237527)
(4, 0.0234374683350324) (5, 0.0289991255849599) (6, 0.0234534405171871)
(7, 0.0224261842668056) (8, 0.0199623573571443) (9, 0.0202894937247037)
(10, 0.0215304661542177) (11, 0.0204663947224617) (12, 0.0221571885049343)
(13, 0.0179081112146377) (14, 0.0214127209037542) (15, 0.0197978142998027)
(16, 0.0169971883296966) (17, 0.0159754566848278) (18, 0.0177929531782865)
(19, 0.0176453441381454) (20, 0.0169870350509882) (21, 0.0168328955769538)
(22, 0.0171852614727285) (23, 0.0153766380622982) (24, 0.0153335379436612)
(25, 0.0147675136104226) (26, 0.0137787051498890) (27, 0.0142603088166505)
(28, 0.0142802838236093) (29, 0.0123213157057762) (30, 0.0135544491931796)
(31, 0.0123051255941391) (32, 0.0134348971707121) (33, 0.0119509497508406)
(34, 0.0116091845557093) (35, 0.0119917495176196) (36, 0.0105974890962243)
(37, 0.0109170190595808) (38, 0.0109635563567280) (39, 0.0116366967558860)
(40, 0.0104075409471988) (41, 0.0116556892171502) (42, 0.0107546709477901)
(43, 0.0112280035391450) (44, 0.0095076365396380) (45, 0.0101935714483261)
(46, 0.0092856874695157) (47, 0.0085876304656267) (48, 0.0089256372302770)
(49, 0.0082765091195702) (50, 0.0078904861584305) (51, 0.0075785983353853)
(52, 0.0074473314452916) (53, 0.0073945718779042) (54, 0.0067088410258293)
(55, 0.0069459979422390) (56, 0.0070756725035607) (57, 0.0063568777404725)
(58, 0.0069187772460281) (59, 0.0079119149595499) (60, 0.0082423239946365)
};
\end{axis}
\end{tikzpicture}
\caption{Reverse KL}
\label{fig:gap_subfig}
\end{subfigure}
\hfill
\begin{subfigure}[t]{\panelw}
\vspace{0pt}
\centering
\begin{tikzpicture}
\begin{axis}[
    width=\linewidth,
    height=\linewidth,
    ylabel={},
    xmin=1, xmax=60,
    ymin=0.05, ymax=0.95,
    axis line style={black},
    tick style={black},
    xticklabel style={font=\footnotesize},
    yticklabel style={font=\footnotesize},
    xlabel style={font=\normalsize},
    title style={font=\normalsize, yshift=-2pt},
    grid=major,
    grid style={dashed, gray!50},
    enlargelimits=false,
    clip=false,
    axis background/.style={fill=white},
    axis x line*=bottom,
    axis y line*=left,
]
\addplot[blue, line width=1.2pt, mark=none] coordinates {
(1, 0.785221) (2, 0.776754) (3, 0.778572) (4, 0.767991) (5, 0.913067)
(6, 0.791450) (7, 0.783577) (8, 0.692772) (9, 0.678163) (10, 0.685105)
(11, 0.616323) (12, 0.764257) (13, 0.586600) (14, 0.628945) (15, 0.519329)
(16, 0.360479) (17, 0.494690) (18, 0.332221) (19, 0.425811) (20, 0.427618)
(21, 0.447718) (22, 0.290328) (23, 0.432296) (24, 0.470807) (25, 0.553893)
(26, 0.580210) (27, 0.483748) (28, 0.493175) (29, 0.463557) (30, 0.409925)
(31, 0.306549) (32, 0.298681) (33, 0.375703) (34, 0.268753) (35, 0.356206)
(36, 0.351278) (37, 0.337543) (38, 0.338436) (39, 0.252828) (40, 0.190057)
(41, 0.242339) (42, 0.276334) (43, 0.186042) (44, 0.191591) (45, 0.197318)
(46, 0.171713) (47, 0.150536) (48, 0.139477) (49, 0.109530) (50, 0.129255)
(51, 0.134853) (52, 0.154719) (53, 0.129558) (54, 0.104919) (55, 0.144196)
(56, 0.102077) (57, 0.112683) (58, 0.096249) (59, 0.107670) (60, 0.219964)
};
\end{axis}
\end{tikzpicture}
\caption{grad\_norm}
\label{fig:grad_norm_subfig}
\end{subfigure}
\hfill
\begin{subfigure}[t]{\panelw}
\vspace{0pt}
\centering
\begin{tikzpicture}
\begin{axis}[
    width=\linewidth,
    height=\linewidth,
    xmin=0, xmax=60,
    ymin=88.8, ymax=92.8,
    axis line style={black},
    tick style={black},
    xtick={0,10,20,30,40,50,60},
    xticklabel style={font=\footnotesize},
    yticklabel style={font=\footnotesize},
    xlabel style={font=\normalsize},
    ylabel style={font=\normalsize},
    title style={font=\normalsize, yshift=-2pt},
    grid=major,
    grid style={dashed, gray!50},
    enlargelimits=false,
    clip=false,
    axis background/.style={fill=white},
    axis x line*=bottom,
    axis y line*=left,
    legend style={
        draw=none,
        fill=white,
        fill opacity=0.85,
        text opacity=1,
        font=\scriptsize,
        at={(1.02,0.04)},
        anchor=south east,
        inner xsep=2pt,
        inner ysep=1pt
    },
]
\addplot[
    orange,
    line width=1.2pt,
    mark=*,
    mark size=1.5pt
] coordinates {
(4,89.259259) (8,89.635417) (10,90.208333)
(12,90.104167) (14,89.635417) (16,90.208333) 
(20,90.729167) (24,90.729167) (28,90.208333)
(32,90.000000) (36,90.833333) (40,90.677083)
};
\addlegendentry{GRPO}

\addplot[
    blue,
    line width=1.2pt,
    mark=*,
    mark size=1.5pt
] coordinates {
(4,89.166667) (8,90.468750) (12,91.197917) (16,91.614583) (20,91.145833)
(24,91.458333) (28,92.447917) (32,92.135417) (36,91.927083) (40,90.937500)
(42,91.510417) (44,91.875000) (46,92.083333) (48,91.614583) (50,92.239583)
(52,91.822917) (54,92.031250) (56,91.822917)
};
\addlegendentry{MOPD}

\addplot[
    black,
    dashed,
    line width=1pt
] coordinates {(0,92.08) (60,92.08)};
\addlegendentry{Teacher}

\end{axis}
\end{tikzpicture}
\caption{AIME25 (avg@64)}
\label{fig:aime25_subfig}
\end{subfigure}
\captionsetup{justification=centering} 
\caption{Training dynamics and downstream evaluation.}
\label{fig:three_panel_training}
\end{figure}

We therefore adopt \textbf{multi-domain on-policy distillation (MOPD)}~\citep{xiao2026mimo, yang2025qwen3,agarwal2024policy,Gu2024MiniLLM, lu2025onpolicydistillation, zeng2026glm} as a complementary post-training stage. In our setting, MOPD is particularly attractive for three reasons. First, teacher checkpoints can be selected directly from the Cascade RL pipeline by choosing the strongest validation checkpoint for each benchmark category, which makes it easy to assemble a capability-diverse teacher pool without introducing external model families. Second, because these teachers are derived from the same SFT initialization, they share the same tokenizer and vocabulary as the student, reducing distribution shift and avoiding additional alignment issues. Third, MOPD provides a dense token-level training advantage, which is especially useful compared with sparse outcome rewards, and in Figure~\ref{fig:aime25_subfig} we show its training-efficiency benefits compared with GRPO.

\paragraph{MOPD objective.}
Let $\pi^{\textit{inf}}$ denote the student policy used for response generation in the inference engine, and let $\pi^{\textit{train}}$ denote the student policy optimized by the training engine. For each prompt $x$, we sample a response $y=(y_1,\ldots,y_T)\sim \pi^{\textit{inf}}(\cdot\mid x)$. We then select a domain teacher $\pi^{\textit{domain}_i}$ for that training example, where $\textit{domain}_i$ indicates the capability domain associated with the chosen teacher. Writing $s_t=(x,y_{<t})$ for the decoding state at step $t$, we define the token-level distillation advantage using reverse-KL as
\begin{equation}
a_t^{\textit{MOPD}}
=
\log \pi^{\textit{domain}_i}(y_t \mid s_t)
-
\log \pi^{\textit{train}}(y_t \mid s_t).
\label{eq:mopd_adv}
\end{equation}
Intuitively, this term is positive when the domain teacher assigns a higher probability to the sampled token than the current training policy, and therefore serves as a dense token-level distillation advantage that converges toward 0 during training. The log-probability difference is computed only on the student-sampled token rather than over the full vocabulary.

Because responses are sampled under $\pi^{\textit{inf}}$ but optimized under $\pi^{\textit{train}}$, we apply truncated importance weighting to account for train--inference mismatch:
\begin{equation}
r_t
=
\frac{\pi^{\textit{train}}(y_t \mid s_t)}{\pi^{\textit{inf}}(y_t \mid s_t)},
\qquad
w_t
=
\mathrm{sg}[r_t]\,
\mathbf{1}\!\left[\epsilon_{\mathrm{low}} \le r_t \le \epsilon_{\mathrm{high}}\right],
\label{eq:mopd_weight}
\end{equation}
where $\mathrm{sg}[\cdot]$ denotes stop-gradient. We then optimize the surrogate objective
\begin{equation}
\mathcal{L}_{\textit{MOPD}}
=
-\mathbb{E}_{x \sim \mathcal{D},\, y \sim \pi^{\textit{inf}}(\cdot \mid x)}
\left[
\frac{1}{|\mathcal{V}(y)|}
\sum_{t \in \mathcal{V}(y)}
w_t \,
\mathrm{sg}\!\left[a_t^{\textit{MOPD}}\right]
\log \pi^{\textit{train}}(y_t \mid s_t)
\right],
\end{equation}
where $\mathcal{V}(y)$ is the set of valid response tokens retained by the token mask.

\paragraph{Hyperparameters.}
Unless otherwise specified, we use a rollout size of 4 and 128 prompts per update, giving an effective batch size of 512 responses. In later experiments, we find that using 512 prompts with rollout size 1 yields slightly more stable optimization while producing similar final results. We use a learning rate of $2\times10^{-6}$ with linear warm-up over the first 30 optimization steps, starting from $2\times10^{-7}$. Training typically converges within 40-50 optimization steps (Fig.~\ref{fig:gap_subfig}). We find the warm-up stage important for stability: gradient norms are substantially larger at the beginning of training and decrease rapidly after the warm-up phase (Fig.~\ref{fig:grad_norm_subfig}). For truncated importance weighting, we set $\epsilon_{\mathrm{low}}=0.5$ and $\epsilon_{\mathrm{high}}=2.0$. In the main experiments, we use three domain teachers corresponding to math, RLHF, multi-domain. 
The math teacher is the initial SFT checkpoint, which already exhibits strong mathematical reasoning capabilities thanks to the meticulously curated SFT dataset. 
The RLHF teacher is a checkpoint optimized through RLHF from the initial SFT checkpoint.
The multi-domain teacher is selected from the checkpoints after previous IF-RL + Multi-domain RL stages. 
We sample prompts accordingly from the RL training data pools (RLHF, IF-RL, and Multi-domain), as well as from AceReason-Math for math~\citep{chen2025acereason}.

\begin{table}[t]
\centering
 \captionsetup{justification=centering}
\caption{Comparison of MOPD and RLHF at matched evaluation checkpoints on ArenaHard V2.0.}
\label{tab:mopd_arenahard}
\renewcommand{\arraystretch}{0.8}
\begin{tabular}{lccc}
\toprule
& & \multicolumn{2}{c}{ArenaHard v2} \\
\cmidrule(lr){3-4}
Method & Steps & Hard Prompt & Creative Writing \\
\midrule
Initial & 0   & 71.5 & 40.6 \\
\midrule
\multirow{2}{*}{RLHF} & 100 & 81.7 & 68.6 \\
                      & 160 & 80.7 & 71.2 \\
\midrule
MOPD                  & 52  & 85.5 & 71.0 \\
\bottomrule
\end{tabular}
\end{table}

\paragraph{Training efficiency advantage.}
MOPD provides a dense token-level distillation advantage, whereas GRPO relies on a sparse sequence-level outcome reward that is shared across all generated tokens. This makes MOPD substantially more sample- and step-efficient in practice. Starting from the same initial checkpoint, MOPD consistently reaches stronger performance in fewer optimization steps. On AIME25 (Figure~\ref{fig:aime25_subfig}), under math-only training, GRPO improves from 89.9 to 91.0 after 25 steps, while MOPD reaches 92.0 within 30 steps and recovers teacher-level performance. A similar trend appears on ArenaHard v2 (Table~\ref{tab:mopd_arenahard}). After 52 steps, MOPD improves Hard Prompt from 71.5 to 85.5 and Creative Writing from 40.6 to 71.0. In contrast, RLHF training requires 160 steps to reach 80.7 on Hard Prompt and 71.2 on Creative Writing. These results show that the dense token-level advantage in on-policy distillation lead to much faster training convergence.

\subsection{Reinforcement Learning from Human Feedback (RLHF)}
\label{sec:rlhf}
Building on multi-domain on-policy distillation, our RLHF recipe focuses on human preferece learning.
This process further enhances creative writing and non-verifiable problem-solving in coding and mathematics, as measured by ArenaHard v2~\citep{li2024crowdsourced}, while maintaining performance across other domains without degradation.

\subsubsection{Dataset}
We adopt the RLHF training dataset from NVIDIA Nano-v3 \citep{blakeman2025nemotron}, which comprises HelpSteer3 \citep{wang2025helpsteer3preferenc}, a commercially-friendly subset of the arena-human-preference-140k dataset \citep{chiang2024chatbot}, and a synthetic safety blend \citep{blakeman2025nemotron}. 
Following the NVIDIA Nano-v3 \citep{blakeman2025nemotron}, we utilize Qwen3-235B-A22B-Thinking-2507~\citep{yang2025qwen3} as our generative reward model (GenRM), trained via the HelpSteer3 framework \citep{wang2025helpsteer3preferenceopenhumanannotatedpreference}. 
Given a conversation history, a user request, and two candidate responses, the GenRM first reasons through the strengths and weaknesses of each response before producing individual helpfulness scores and a final comparative ranking.

\subsubsection{Training recipe}
Following a training recipe similar to NVIDIA Nano-v3 \citep{blakeman2025nemotron}, we conduct RLHF using the GenRM.
To ensure the training signals are of high quality, we adopt pair-wise comparisons for all pairs of rollouts per prompt. 
We aggregate the reward scores in the same way as NVIDIA Nano-v3 RLHF training, and apply the same length-normalized reward adjustment and quality-gated conciseness bonus \citep{blakeman2025nemotron}.
These mechanisms encourage shorter responses without sacrificing quality, effectively mitigating the rapid growth of inference token usage. 

Different from Nemotron Cascade \citep{wang2025nemotron}, we train RLHF exclusively in the thinking mode. While incorporating both thinking and non-thinking modes can improve training convergence and yield slight gains on evaluation benchmarks, we observe a significant degradation in instruction-following performance. The resulting drop is substantial enough that the gains obtained in the earlier RLVR stage cannot be fully recovered.

\subsubsection{Hyper-parameters}
We use a batch size of 128, generating 16 rollout per prompt with a temperature of 1.0 and a top-p value of 1.0. We use a maximum response length of 16K during RLHF without applying overlong filtering. 
We adopt a learning rate of 3e-6 with AdamW \citep{kingma2014adam}.
We set the entropy loss coefficient  to 0 and the KL loss coefficient to 0.03 to keep the model capabilities on other domains. The training takes around 30 steps. 

\subsection{Long-context RL}
\label{sec:lcrl}
Following RLHF, we conduct a stage of long-context RL to further enhance the model’s long-context understanding and reasoning capabilities. We use the NVIDIA Nano-v3 RL data blend \citep{blakeman2025nemotron}, but restrict this phase to long-context datasets only. In our experiments, incorporating other domains during long-context RL negatively affects performance on unrelated benchmarks, motivating this domain-specific training setup.

We adopt the Nemo-Gym RL environment \citep{nemo-gym} and use Qwen3-235B-A22B-Instruct-2507 as an LLM judge to evaluate model rollouts for question answering tasks. During training, input sequences are limited to 32K tokens, and the maximum sequence length is set to 49K tokens without applying overlength filtering.

We train with a batch size of 128, generating 16 rollouts per prompt with temperature 1.0 and top-p 1.0. Optimization is performed using AdamW \citep{kingma2014adam} with a learning rate of $3\times10^{-6}$, while both the entropy and KL loss coefficients are set to zero. Training runs for approximately 30 steps, as we observe a rapid increase in generated tokens beyond that point.

\subsection{Code RL}
\label{sec:coderl}

\subsubsection{Data Curation}

We construct our Code RL training set from the Nemotron-Cascade coding corpus~\citep{wang2025nemotron}, which contains coding prompts sourced from modern competitive programming platforms such as AtCoder, Codeforces, and AIZU with robust test cases for reward verification. To improve training efficiency and strengthen deep reasoning, we aggressively filter out prompts that GPT-OSS-120B solves correctly in all 8 of 8 rollouts, yielding a compact final set of only 3.5K samples. We find that high-difficulty prompts paired with strong test cases are critical for further boosting model performance.

\subsubsection{Training Details}

We conduct Code RL using a batch size of 128 and a learning rate of $3\times 10^{-6}$ with the AdamW optimizer. Compared to Nemotron-Cascade, we increase the maximum response length during RL to 118K tokens and the number of rollouts per sample to 16, enabling the policy to better capture sparse reward signals on extremely difficult problems that require long reasoning traces. We adopt the strict binary reward function to avoid potential reward hacking and keep the whole training to be fully on-policy for stability. To support the resulting verification throughput of $128 \times 16=2,048$ code executions per RL step, we deploy an asynchronous reward verification server that completes each batch in 427.2 seconds across 384 CPU cores.

\subsection{Software Engineering Reinforcement Learning (SWE RL)}
\label{sec:swerl}

\subsubsection{Agentless RL}
\paragraph{Training Details and Hyperparameters.}
To enhance the models' code repair capability, we adopt the same data source as \citet{wang2025nemotron} for agentless code repair reinforcement learning (RL) training. Since most instances do not provide executable Docker environments, we employ GPTOOS-120B as a reward model to evaluate the quality of code repairs generated by our models.
Following \citet{wang2025nemotron}, for each instance we construct prompts using both the golden localization and the top-$5$ retrieved localizations, and filter out relatively easy samples. 
We perform agentless SWE RL with a batch size of $128 \times 16 = 2{,}048$ (128 prompts with 16 rollouts per prompt), a maximum sequence length of 98,304, and a learning rate of $3\times10^{-6}$ using the AdamW optimizer. 
We sample responses with temperature 1.0 and top-p 1.0.
During training, we mask the loss for prompts for which none of the rollouts receives a reward greater than 0.5. We observe that these difficult prompts degrade the stability and effectiveness of agentless SWE RL training. Our agentless RL training typically converges within 40–50 steps.

\paragraph{Can Agentless RL Training Helps Agentic Tasks?}
Table~\ref{tab:swe_agentless_rl} shows that agentless RL training not only improves model performance within the agentless framework but also enhances the models’ ability to solve SWE tasks in agentic settings. 
Note that for Agentless Mini evaluation, we employ a code embedding model, NV-Embed-Code \citep{sohrabizadehnemotron}, to retrieve 5 candidate files whose code contents are semantically similar to the problem context. 
This result suggests that improving models’ code repair capability alone can generalize across different scaffolds, consistent with the observations from \citet{yang2026kimidev}.

\begin{table}[!t]
\centering
\small
\renewcommand{\arraystretch}{1.15}
\captionsetup{justification=centering}
\caption{Effectiveness of Agentless RL on SWE-bench Verified.}
\label{tab:swe_agentless_rl}
\begin{tabular}{ccccc}
\toprule
\multirow{2}{*}{Scaffold} 
  & \multicolumn{2}{c}{Agentless Mini} 
  & \multicolumn{2}{c}{OpenHands} \\
\cmidrule(lr){2-3} \cmidrule(lr){4-5}
 & {\footnotesize avg@4} & {\footnotesize pass@4}
 & {\footnotesize avg@4} & {\footnotesize pass@4} \\
\midrule
Init. & 41.9\% & 55.2\% & 49.8\% & 64.2\% \\
\midrule
after Agentless RL & 44.3\% & 57.4\% & 50.8\% & 65.0\% \\
\bottomrule
\end{tabular}
\end{table}

\subsubsection{Execution-based RL for Agentic SWE Scaffold}
Modern software engineering agents rely on scaffolding frameworks that coordinate repository interaction, tool calling, code editing, and test execution. Training agents to operate effectively within these environments requires optimizing not only individual model outputs but the entire problem-solving trajectory. To address this, we apply Reinforcement Learning from Verifiable Rewards (RLVR) directly within agentic SWE scaffolds, enabling end-to-end optimization of the full agent workflow. Our training environments integrate established OpenHands frameworks~\citep{wang2025openhands}, which provide structured tool usage, repository interaction, and iterative patch generation.

We train agents using execution-based reinforcement learning in fully executable software environments, where each episode corresponds to resolving a software issue instance from benchmarks such as SWE-bench. The agent operates inside an instrumented repository that exposes tools for file inspection, search, code editing, and test execution. Candidate patches generated by the agent are executed within the environment, which returns verifiable signals from compilation results and unit test outcomes, enabling automatic reward computation without human annotation. Through the OpenHands scaffolding framework, the agent iteratively localizes defects, proposes patches, and validates them through test execution. Environment feedback—including compilation errors, failing tests, or successful test passes—provides deterministic rewards that directly reflect functional correctness.

Specifically, we conduct execution-based agentic reinforcement learning with a batch size of 1024, corresponding to 16 prompts with 64 rollouts per prompt. The maximum context length is set to 256k tokens, and the agent is allowed up to 200 interaction turns, providing a larger reasoning token budget during agentic coding problem solving. Training data is drawn from SWE-Gym~\citep{training-software} and R2E-Subset~\citep{jain2025r2e}. We generate 16 rollouts per instance using our intermediate model and evaluate them using the verification pipeline. Instances for which all rollouts pass verification (100\% accuracy), indicating overly simple problems, are removed from the dataset. For instances where none of the rollouts pass verification (0\% accuracy), indicating extremely difficult problems, we randomly discard 90\% of such cases to reduce their proportion in the training data.


\section{International Mathematical Olympiad (IMO)}
\label{sec:imo}

\subsection{IMO 2025}
In Table~\ref{tab:main_IMO_IOI_ICPC_2025}, we evaluate Nemotron-Cascade-2-30B-A3B on the IMO 2025 problem set using a self-improving test-time scaling framework~\citep{shao2025deepseekmath}, in which the model iteratively generates candidate solutions, verifies them, and refines them based on its own feedback. Remarkably, despite its relatively modest 30B-A3B scale, the model successfully solves the first five problems. We provide the full model solutions in Appendix~\ref{appendix:imo2025-model-solutions}, together with comments from the human expert. These results are particularly encouraging, as they suggest that strong olympiad-level mathematical reasoning can emerge from a comparatively compact model when paired with effective inference-time scaling. There remain several promising directions for improvement: expert review indicates that some proofs are longer than necessary, include superfluous intermediate steps or definitions, occasionally expose traces of intermediate reasoning, and sometimes contain minor typographical issues. For Problem 2, the model adopts an analytic solution strategy, similar to OpenAI’s approach, rather than a more geometric approach such as that used by Gemini Deep Think (IMO Gold).

\subsection{IMO-ProofBench}

\begin{table}[h]
\centering
\small
\caption{IMO-ProofBench~\citep{luong2025towards} reports scores split into the Basic (30 problems) and Advanced (30 problems) subtasks, as well as Overall (60 problems). Expert-evaluated results are taken from the IMO-ProofBench leaderboard (accessed on 2026/3/9).}
\begin{tabular}{lccc}
\toprule
\textbf{Model} & \multicolumn{3}{c}{\textbf{IMO-ProofBench}} \\
\cmidrule(lr){2-4}
 & \textit{Basic (30)} & \textit{Advanced (30)} & \textit{Overall (60)} \\
\midrule
Aletheia~\citep{feng2026towards}                      & - & 91.9 & - \\
Gemini 3 Deep Think~\citep{deepthinkteam2026gemini3deepthink}  & - & 76.7 & - \\
Gemini Deep Think (IMO Gold)~\citep{googledeepmind2025imogold}  & 89.0 & 65.7 & 76.7 \\
DeepSeek-Math-V2-671B-A37B~\citep{shao2025deepseekmath}              & 99.0 & 61.9 & 80.2 \\
DeepSeek-Math-V2-671B-A37B (\textit{our reproduced score})~$^\dag$ & 99.5 & 57.7 & 78.6 \\
\rowcolor{gray!10} \textbf{Nemotron-Cascade-2-30B-A3B}$^\dag$ & 92.5 & 53.4 & 72.9 \\
GPT-5.2-Thinking (high)~\citep{openai2025gpt5} & - & 35.7 & - \\
Gemini 3 Pro~\citep{google2025gemini3} & - & 30.0 & - \\
GPT-5 Pro~\citep{openai2025gpt5} & - & 28.6 & - \\
\bottomrule
\end{tabular}
\begin{flushleft}
\small
\setlength{\hangindent}{1em} 
\vspace{0.2em}
$\dag$ Use DeepSeek-V3.2-Speciale as the judge model with LLM ProofAutoGrader prompt~\citep{luong2025towards}.\\
\end{flushleft}
\label{tab:imo-proofbench}
\end{table}

As shown in Table~\ref{tab:imo-proofbench}, Nemotron-Cascade-2-30B-A3B achieves $72.9$ on IMO-ProofBench with generate-verify-refine test-time scaling, placing it within 8 points of DeepSeek-Math-V2-671B-A37B despite using 10$\times$ fewer active parameters. It reaches 90+ on Basic split and surpass the QED-Nano-4B ($54.0$)~\citep{qednano2026} by 18 points, though the latter is not directly comparable due to judge model. Re-evaluating the provided DeepSeek-Math-V2 proofs under our LLM-judge setup yields a score within 4 points of the reported human rating, suggesting that our protocol does not substantially overestimate performance (more details in Appendix~\ref{appendix:proof_benchmarks}). In Figure~\ref{fig:imo_proofbench_tts}, we show that increasing test-time compute improves Nemotron-Cascade-2-30B-A3B on IMO-ProofBench (Advanced), raising the score from 40.7 at round 1 to 53.4 at round 5 and narrowing the gap to DeepSeek-Math-V2 under the same grader.

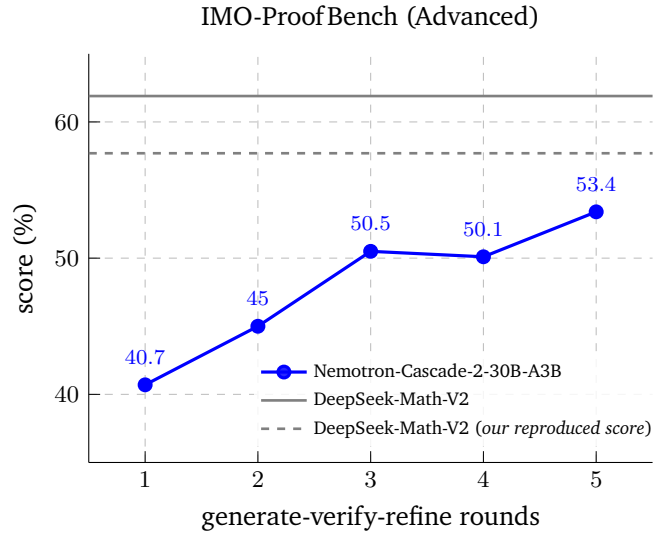
\begin{figure}[!ht]
\centering
\begin{tikzpicture}
\begin{axis}[
    width=9cm,
    height=7cm,
    title={IMO-ProofBench (Advanced)},
    xlabel={generate-verify-refine rounds},
    ylabel={score (\%)},
    xmin=0.5, xmax=5.5,
    ymin=35, ymax=65,
    xtick={1,2,3,4,5},
    axis line style={black},
    tick style={black},
    xticklabel style={font=\small},
    yticklabel style={font=\small},
    xlabel style={font=\normalsize},
    ylabel style={font=\normalsize},
    title style={font=\normalsize},
    grid=major,
    grid style={dashed, gray!50},
    enlargelimits=false,
    clip=false,
    axis background/.style={fill=white},
    axis x line*=bottom,
    axis y line*=left,
    legend style={
        draw=none,
        fill=white,
        fill opacity=0.85,
        text opacity=1,
        font=\scriptsize,
        at={(1.02,0.04)},
        anchor=south east,
        inner xsep=2pt,
        inner ysep=1pt,
        cells={anchor=west}
    },
]

\addplot[
    blue,
    line width=1.2pt,
    mark=*,
    mark size=2.2pt,
    point meta=y,
    nodes near coords,
    every node near coord/.append style={
        font=\footnotesize,
        yshift=4pt
    },
    nodes near coords style={
        /pgf/number format/fixed,
        /pgf/number format/precision=1
    }
] coordinates {
(1, 40.7)
(2, 45)
(3, 50.5)
(4, 50.1)
(5, 53.4)
};
\addlegendentry{Nemotron-Cascade-2-30B-A3B}

\addplot[
    gray,
    line width=1pt
] coordinates {(0.5,61.9) (5.5,61.9)};
\addlegendentry{DeepSeek-Math-V2}

\addplot[
    gray,
    dashed,
    line width=1pt
] coordinates {(0.5,57.7) (5.5,57.7)};
\addlegendentry{DeepSeek-Math-V2 (\textit{our reproduced score})}

\end{axis}
\end{tikzpicture}
\caption{IMO-ProofBench (Advanced) score graded by LLM ProofAutoGrader (DeepSeek-V3.2-Speciale).}
\label{fig:imo_proofbench_tts}
\end{figure}

\section{Competitive Coding}
\label{sec:ioi}

\begin{table}[!htbp]
\centering
\caption{Competitive programming results on comprehensive benchmarks, evaluated against a significantly expanded set of proprietary and open-source baseline models.}
\begin{adjustbox}{width={1.0\textwidth}}
\setlength{\tabcolsep}{10pt}
\renewcommand{\arraystretch}{1.2}
\begin{tabular}{l|c|cccccc|cc}

\arrayrulecolor{black}\toprule
\multirow{3}{*}{\bf Models} & \multicolumn{1}{c|}{\bf LiveCodeBench} & \multicolumn{6}{c|}{\bf LiveCodeBench Pro}               & \multicolumn{2}{c}{\bf Codeforces}  \\
                                  & \bf  v6              & \multicolumn{3}{c}{\bf 25Q1} & \multicolumn{3}{c|}{\bf 25Q2} & \multicolumn{2}{c}{2501~-~2507 } \\
                            & 2408~-~2505     & {\bf \color[HTML]{808080} Easy}        & {\bf \color[HTML]{808080} Med}  & {\bf \color[HTML]{808080} Hard}     & {\bf \color[HTML]{808080} Easy}         & {\bf \color[HTML]{808080} Med}   & {\bf \color[HTML]{808080} Hard}    & \bf ELO         & \bf  Percentile        \\ \arrayrulecolor{black} \midrule

GPT-5.2 (high)  & - & 96.6 & 75.0 & 5.9 & 91.8 & 59.6 & 23.1 & 2590 & 99.9 \\
Gemini-3 Pro  & 90.7 & 94.4 & 70.0 & 5.9 & 94.8 & 45.6 & 7.7 & 2440  & 99.8 \\
GPT-o4-mini (high)                                          & 80.2                &     85.4                  &           51.7     & 0.0       &          84.5              &              29.8        & 0.0
&        2266                     &   99.5                           \\

\arrayrulecolor{gray} \midrule
DeepSeek-v3.2-Speciale  & 88.7 & 89.7 & 48.1 & 0.0 & 88.5 & 43.1 & 0.0  & 2353 & 99.7 \\
GPT-OSS-120B (high)  & 87.0 & 88.8 & 41.9 & 0.7 & 88.5 & 31.1 & 0.0 & 2320 & 99.6 \\
Kimi-K2.5-1T-thinking & 85.0 & 88.5 & 45.6 & 0.0 & 90.2 & 37.9 & 0.0 & 2333 & 99.7 \\
Qwen-3.5-397B-A17B & 83.6 & 89.3 & 44.4 & 0.0 & 88.1 & 31.4 & 0.0 & 2350 & 99.7 \\
Qwen-3.5-122B-A10B & 78.9 & 87.6 & 35.6 & 0.0 & 84.3 & 24.2 & 0.0 & 2233 & 99.4 \\
Qwen-3.5-35B-A3B & 74.6 & 84.6 & 25.6 & 0.0 & 81.1 & 17.8 & 0.0 & 2181 & 99.1 \\
Nemotron-3-Super-120B-A12B & 78.7 & 83.0 & 31.0 & 0.0 & 81.7 & 23.2 & 0.0 & 2212 & 99.4 \\
Qwen3-235B-A22B-Thinking-2507         &   78.7             &      75.8                    &        18.8                  &   0.0 &    77.6                   &           17.5         & 0.0
&              2119                &        98.6                      \\

Nemotron-Cascade-14B                                       &  74.6                         &                 71.6         &     16.3           & 0.0           &    68.9                      &       10.5 & 0.0
&            2004                 &             97.9            \\ 
Qwen3-Next-80B-A3B-Thinking      &  73.2           &      68.5                    &          16.3        & 0.0        &       69.1                  &           7.5       & 0.0  
&                 1894            &           96.8                   \\
Nemotron-3-Nano-30B-A3B & 68.3 & 60.3 & 6.0 & 0.0 & 54.5 & 3.5 & 0.0 & 1681 & 93.1 \\
\arrayrulecolor{black}

\midrule
\rowcolor{gray!20} Nemotron-Cascade-2-30B-A3B                                          & \bf 87.2                         &                 \bf 88.1        &  \bf   39.2       & \bf 0.7              &    \bf 87.0                     &    \bf   27.6 & 0.0
&        \bf     2320                 &  \bf                99.6            \\ 
\rowcolor{gray!20} Nemotron-Cascade-2-30B-A3B (TIR)                                          & \bf 88.4                        &               \bf   91.0       &  \bf 45.2       & \bf 2.2              &   \bf  89.3                  &    \bf 36.8 &   0.0 & \bf   2345                &  \bf 99.7            \\ 
\bottomrule
\end{tabular}
\label{tab:competitive_programming_results}
\end{adjustbox}
\end{table}

\subsection{IOI 2025 and ICPC World Finals 2025}
For IOI 2025, we adapt the IOI Test-Time Scaling pipeline from Nemotron-Cascade~\citep{wang2025nemotron}, which can be viewed as a multi-round \textit{generate-select-submit} framework that exploits the model's reasoning ability under IOI's official rules. Each subtask is allotted at most 50 rounds. Within each round, we prompt our model to generate 40 candidate solutions, aggregated with (1) submission history with official judge verdicts from previous rounds, and (2) shared insights from high scored or fully solved subtasks within the same main task. The complete chat template is provided in Appendix~\ref{appendix:cp_template}. Using this approach, we achieved full score on Problem 3 and 4, achieving a gold-medal score of 439.28 within at most $40 \times 50 = 2000$ model generations, while the score of 507.66 is achievable within 5000 generations. Notably, on Problem 2 which requires designing and optimizing a heuristic algorithm, our pipeline reached over 86 points in just 5 rounds (at most 200 model generations), demonstrating the effectiveness of self-refinement and cross-subtask insights.

For ICPC World Finals 2025, we generate up to 1000 solutions per problem and submit them for official evaluation after initial filtering. We successfully solved 10 out of 12 problems, achieving the \#4 Gold medal placement, with 8 problems (except Problems A and I) solved within only 100 submissions.

\subsection{Competitive Coding Benchmark Results}

We evaluate our Nemotron-Cascade-2-30B-A3B model 
on various competitive coding benchmarks, including LiveCodeBench v6~\citep{jain2024livecodebench}, and LiveCodeBench Pro~\citep{zheng2025livecodebench}'s 25Q1 and 25Q2 splits. We also estimate Codeforces ELO score through simulated participation on \textbf{40} Div.1/Div.2 Codeforces Rounds held from 2501 to 2507. We report our avg@8 results under 128K-token thinking budget, the sampling temperature of 1.0 and the \texttt{top\_p} of 0.95. For Tool-Integrated Reasoning (TIR) results, we allow our model to call a stateful Python executor for up to 100 calls. For baseline model evaluation, we follow their recommended inference configurations, ensuring a thinking budget of at least 128K tokens to at most 256K tokens. More evaluation details can be found in Appendix~\ref{appendix:benchmarks} and Appendix~\ref{appendix:elo_rating_on_codeforces}.

As shown in Table~\ref{tab:competitive_programming_results}, Nemotron-Cascade-2-30B-A3B achieves magnificent Pass@1 accuracy and ELO rating, even compared with frontier open-source models with over 100B total params, such as Nemotron-3-Super-120B-A12B, GPT-OSS-120B, and Qwen-3.5-122B-A10B. With Tool-Integrated Reasoning (TIR), our model's performance can be further boosted especially on hard problems, and match the strongest open-source models with more than 300B total parameters, such as Kimi-K2.5-1T-Thinking, Qwen-3.5-397B-A17B, and DeepSeek-v3.2-Speciale, which either lack TIR support for deep reasoning or perform poorly with Python TIR. Notably, Nemotron-Cascade-2-30B-A3B achieves above 0\% on the LiveCodeBench Pro hard split within 8 attempts, demonstrating strong reasoning ability on problems that are extremely difficult even for humans.

\section{Acknowledgments}
\label{appendix:contributors}

We would like to extend our gratitude to the NVIDIA Nemo team for the valuable discussion and collaboration
on building reasoning models. We especially wish to thank Boris Ginsburg, Oleksii Kuchaiev, Igor Gitman, Olivier Delalleau, Zhilin Wang, Olivier Delalleau, Tugrul Konuk, Wei Du, Somshubra Majumdar, Wasi Uddin Ahmad, Siddhartha Jain, Jiaqi Zeng, Yi Dong, Alexander Bukharin, Vahid Noroozi, Khushi Bhardwaj, Sugam Dipak Devare, Jian Zhang, and Jonathan Cohen.

We thank Ying Lin for helpful discussions and useful input in building the knowledge-intensive SFT dataset.
We also thank Atefeh Sohrabizadeh, Jialin Song, and Jonathan Raiman for valuable discussions on SWE-bench.

\clearpage

\paragraph{\LARGE Appendix}
\vspace{0.3cm}

\appendix

\section{Benchmarks and Evaluation Setups}
\label{appendix:benchmarks}

\subsection{Math}
\subsubsection{Non-proof Math}
\label{appendix:math_benchmarks}
For non-proof math reasoning tasks, we include 
\begin{itemize}[leftmargin=2em]
    \item
    \textbf{AIME 2025}~\citep{aime2025} consists of 30 problems from American Invitational Mathematics Examination at 2025.
    \item
    \textbf{AIME 2026}~\citep{aime2026} consists of 30 problems from American Invitational Mathematics Examination at 2026.
    \item
    \textbf{HMMT Feb 2025}~\citep{hmmt_feb_2025} consists of 30 problems from Harvard-MIT Mathematics Tournament 2025 February math competition.
    \item
    \textbf{IMO-AnswerBench}~\citep{luong2025towards} consists of 400 problems with verifiable answers carefully chosen from past Olympiad competitions
    and then altered by experts to avoid memorization.
\end{itemize}

For Nemotron-Cascade-2-30B-A3 evaluated on AIME 2025, AIME 2026 and HMMT 2025 Feb, we set the thinking budget (maximum response length) to 131K tokens, the sampling temperature to 1.0, the top-p value to 1.0. For the with-tool setting, we enable tool use by appending a system-prompt postfix, allowing the model to call a stateful Python executor for up to 100 tool calls with a maximum response length of 131K tokens.
For IMO-AnswerBench, we set to 256K tokens because we found the questions are significantly more difficult. We use and report the LLM-Judge score using GPT-OSS-120B~\citep{agarwal2025gpt} as the judge and the AnswerAutoGrader prompt~\citep{luong2025towards} for answer correctness on IMO-AnswerBench as the short answers are complicated for rule-based verifier to compute. Following \citet{liu2024acemath,liu2025acereason}, we report avg@64 for AIME/HMMT and avg@16 for IMO-AnswerBench.

For baseline models, we use official numbers from their reports or evaluate them with the recommended settings if the official numbers are unavailable.

\subsubsection{Math Proof}
\label{appendix:proof_benchmarks}
For math proof tasks, we include

\begin{itemize}[leftmargin=2em]
    \item
    \textbf{IMO 2025}~\citep{imo2025} consists of 6 problems from IMO 2025.
    \item
    \textbf{IMO-ProofBench}~\citep{luong2025towards} is designed to evaluate the
    ability of AI models to construct comprehensive
    and valid mathematical arguments. This benchmark consists of 60 proof-based problems, curated to mirror the kinds of problems found in
    the IMO.
\end{itemize}

For Nemotron-Cascade-2-30B-A3, we apply test-time scaling following the DeepSeek-Math-V2 generate-verify-refine pipeline, using the same instructions. We implement this pipeline with NeMo-Skills~\citep{nemoskills2025}. We use the default hyperparameters from DeepSeek-Math-V2: 128 proof generations, 64 verifications per proof, selection of the top 32 proofs for refinement, and 8 verification analyses paired with each proof, prioritizing the lowest-rated analyses. We then generate 4 refined proofs and continue for up to 8 rounds, or until the average proof score reaches the threshold of 0.99999. We set the maximum generation length to 256K tokens, with temperature 1.0 and top-p 0.95. 

For IMO-ProofBench Basic and 11 problems from the Advanced split (i.e., Problems 1, 4, 7, 13, 14, 17, 19, 22, 25, 26, and 28), we reduce the compute budget to 32 proof generations, 16 verifications, top 8 proofs, and 2 rounds to save compute. For IMO-ProofBench evaluation, we use DeepSeek-V3.2-Speciale to make sure the results are reproducible later and run 64 grading attempts with the ProofAutoGrader prompt~\citep{luong2025towards}. We found that reporting mean score yields 73.8 for DeepSeek-Math-V2 on the Advanced split, which is substantially more generous than the human rating of 61.9. We therefore adopt a simple aggregation rule based on analysis: \texttt{if any judge assigns a score of 0, the final score is set to 0; otherwise, return the mean score}. Under this rule, DeepSeek-Math-V2 obtains 57.7, which is much closer to the human rating and reduces the discrepancy from 11.9 points to 4.2 points.

\subsection{Code Reasoning}
For code generation tasks, we include
\begin{itemize}[leftmargin=2em]
    \item 
    \textbf{LiveCodeBench}~\citep{jain2024livecodebench} contains diverse algorithm coding problems with unit tests, collected from AtCoder, LeetCode platforms. We evaluate models competitive coding capability on LiveCodeBench v6~(2024/08-2025/05, \textbf{454} problems in total). We report pass@1 accuracy in \emph{thinking} mode, averaged over 8 generations (avg@8).
    \item 
    \textbf{LiveCodeBench Pro}~\citep{zheng2025livecodebench} contains daily-updated challenging competitive coding problems with strong unit tests, collected mainly from top-tier coding contests. We report pass@1 accuracy on Easy/Med difficulty splits in \emph{thinking} mode, averaged over 8 generations (avg@8) on two recently released subsets: 2025Q1 (2025/01-2025/04, \textbf{166} problems in total) and 2025Q2 (2025/04-2025/07, \textbf{167} problems in total).
    \item 
    \textbf{IOI and ICPC World Finals} represent the most challenging and prestigious annual algorithmic coding competitions, gathering the world's top human contestants. The IOI awards gold medals to approximately the top 8.3\% (one-twelfth) of participants, while the ICPC World Finals (ICPCWF) limits gold medals to only the top 4 teams globally.
    \item
    \textbf{SciCode}~\citep{tian2024scicode} serves as a challenging benchmark to evaluate model's ability on solving realistic scientific research tasks from STEM domains. It contains \textbf{338} subproblems from \textbf{80} main tasks.
\end{itemize}

For Nemotron-Cascade-2-30B-A3B evaluated on LiveCodeBench v6 and LiveCodeBench Pro, we use a 128K-token thinking budget, a sampling temperature of 1.0, a top-p of 0.95. For the with-tool setting, we enable tool use by appending a system-prompt postfix, allowing the model to call a stateful Python executor for up to 100 tool calls with a maximum response length of 131K tokens. We evaluate baseline models with their recommended inference configurations, ensuring a thinking budget of at least 128K tokens.

\subsection{Knowledge and STEM}
For knowledge reasoning tasks, we include:
\begin{itemize}[leftmargin=2em]
    %
    \item 
    \textbf{MMLU-Redux}~\citep{gema2024mmlu} is a benchmark consisting of a subset of 3,000 manually re-annotated questions across 30 MMLU subjects~\citep{hendrycks2020measuring}, which eliminates the original annotation errors.
    We evaluate the models in \emph{thinking} mode and, due to the large test set size, report exact match~(EM) accuracy based on a single generation per question.
    \item 
    \textbf{MMLU-Pro}~\citep{wang2024mmlu} is an enhanced version of the original MMLU benchmark that mitigates model saturation by expanding to over 12,000 graduate-level questions and increasing answer choices from four to ten. We report EM accuracy in \emph{thinking} mode using one generation per question.
    \item 
    \textbf{GPQA-Diamond}~\citep{rein2024gpqa} is a benchmark for assessing an LLM’s scientific reasoning capability. It consists of the highest quality 198 GPQA questions covering graduate-level physics, biology, and chemistry. We report pass@1 accuracy in \emph{thinking} mode, averaged over 8 generations per question (avg@8) to reduce variance.
    \item 
    \textbf{HLE}~\citep{phan2025humanity} is a frontier academic reasoning benchmark spanning a broad range of expert-level subjects. We evaluate on its text-only split, which contains 2,158 examples.
\end{itemize}
For Nemotron-Cascade-2-30B-A3B evaluated on MMLU-Redux, MMLU-Pro, GPQA-Diamond and HLE in \emph{thinking} mode, we use a temperature of 1.0, a top-p value of 0.95, and a 128K-token thinking budget (maximum response length). 
For HLE, we use the default system prompt and append ``Please place your final answer inside \texttt{\textbackslash boxed\{\}}'' to each question, and use GPT-OSS-120B as the LLM judge for answer extraction and correctness verification with the prompt in Appendix~\ref{appendix:hle_judge_prompt}. Compared with the official HLE response format, which requests an explanation, an answer, and a confidence score, this boxed-answer prompt improves the accuracy by 6--7 points, primarily on the math subset, by better aligning with the answer format used in our math SFT data.

\subsection{Alignment and Instruction-Following}

For alignment tasks, we include:
\begin{itemize}[leftmargin=2em]
    %
    \item 
    \textbf{ArenaHard 2.0}~\citep{li2024crowdsourced} is a human-preference alignment benchmark featuring 750 diverse and rigorous real-user prompts. 
    The dataset is specifically structured with 500 prompts targeting open-ended software engineering problems and complex mathematical questions, while the remaining 250 focus on creative writing.
    It uses an automatic LLM-as-Judge approach to estimate human preferences relative to a baseline model, enabling fully automated, low-cost, and fast evaluation without human intervention.  
    In our experiments, we report results without style control to allow for straightforward comparison with the officially reported numbers of other models. We evaluate the models in \emph{thinking} mode, and use GPT-4.1 as the automated judge. 
    \item \textbf{IFBench}~\citep{pyatkin2025generalizingverifiableinstructionfollowing} extends IFEval~\citep{zhou2023instructionfollowingevaluationlargelanguage} by introducing 58 new, diverse, and challenging verifiable out-of-domain instruction constraints. It provides a separate constraint list to ensure no overlap between training and test constraints, enabling evaluation of an LLM’s generalization ability. The test set contains 294 prompts. We report pass@1 accuracy in \emph{thinking} mode, averaged over 8 generations (avg@8).
    \item 
    \textbf{Scale AI Multi-Challenge}~\citep{deshpande2025multichallenge} is a benchmark designed to evaluate LLMs in multi-turn conversations with human users. It consists of four challenge categories: Instruction Retention, Inference Memory, Reliable Versioned Editing, and Self-Coherence. These tasks require models to simultaneously perform accurate instruction following, effective context management, and in-context reasoning. The test set contains 273 conversations in total. We report pass@1 accuracy in thinking mode, averaged across 10 generations (avg@10).
\end{itemize}
For Nemotron-Cascade models evaluated on IFEval in \emph{non-thinking} mode, on IFBench and ArenaHard in \emph{thinking} mode, we use a temperature of 0.6, a top-p value of 0.95, and a maximum response length of 32K tokens. 
For baseline models, we use officially reported results whenever available; if such results are absent, we evaluate them using their recommended inference configuration or the same settings as ours.

\subsection{Long Context and Context Learning}

For long context and context learning tasks, we include:
\begin{itemize}[leftmargin=2em]
    \item
    \textbf{AA-LCR}~\citep{artificialanalysis2025lcr} consists of 100 challenging text-based questions that require reasoning over multiple long, real-world documents, including company reports, government consultations, legal documents, and academic papers. Each sample contains a document set averaging approximately 100k tokens. The questions are designed such that answers cannot be directly retrieved from the documents and instead require reasoning across multiple sources of information. We report pass@1 accuracy in thinking mode, averaged over 16 generations (avg@16).
    
    \item 
    \textbf{LongBench v2}~\citep{bai2025longbench} contains 503 challenging multiple-choice questions with context lengths ranging from 8k to 2M words. The benchmark spans six task categories: single-document QA, multi-document QA, long in-context learning, long dialogue history understanding, code repository understanding, and long structured data understanding. The questions are designed to be difficult; even human experts equipped with document search tools may require substantial time to answer them correctly. We evaluate models in thinking mode and report pass@1 accuracy averaged over four generations (avg@4).
    
    \item 
    \textbf{NIAH@1M (Ruler Subset)} refers to the needle-in-a-haystack (NIAH) tasks from the RULER benchmark~\citep{hsieh2024ruler}. The NIAH test~\citep{niah_2023} assesses an LLM's long-context ability to retrieve a specific piece of information (the ``needle'') embedded within long distractor text (the ``haystack''). 
    The RULER benchmark defines four variants of this task: Single NIAH, Multi-keys NIAH, Multi-values NIAH, and Multi-queries NIAH. Following~\citet{blakeman2025nemotron}, we evaluate 100 instances from each category using a 1M-token context setting.
    Models are evaluated in reasoning-off mode, and we report pass@1 accuracy from a single generation (avg@1).
    
    \item 
    \textbf{CL-Bench}~\citep{dou2026cl} evaluates an LLM’s ability to learn from provided context and apply the acquired knowledge to solve tasks, a process referred to as context learning. The benchmark contains 1,899 test samples spanning 500 complex contexts and 31,607 verification rubrics, all developed by experienced domain experts. The knowledge required to complete these tasks largely falls outside what existing models typically learn during pre-training, requiring models to learn directly from the provided context. Models are evaluated in thinking mode, and we report pass@1 accuracy from a single generation (avg@1).
    
\end{itemize}

\subsection{Agentic Tasks}
For agentic tasks, we include:
\begin{itemize}[leftmargin=2em]
    \item 
    \textbf{BFCL v4}~\citep{patil2025bfcl} offers a comprehensive agentic evaluation framework for LLMs, covering tasks such as web search, memory reading and writing, and function invocation across multiple programming languages. We follow the official BFCL V4 evaluation protocol and report scores across a combination of Agentic, multi-turn, live, and non-live categories. Models are evaluated in thinking mode, and we report pass@1 accuracy based on a single generation (avg@1).
    \item
    \textbf{SWE-bench Verified}~\citep{openai2024swe_verified} is a subset of the original test set from SWE-bench~\citep{jimenez2023swe}, consisting of 500 samples verified to be non-problematic by human annotators. We evaluate models in \emph{non-thinking} mode and report pass@1 accuracy, averaged over 4 generations per prompt~(avg@4).
    \item 
    \textbf{$\tau^2$-Bench}~\citep{barres2025tau2} evaluates multi-turn customer-service agents in environments with explicit policies, tool use, and shared world-state updates. We evaluate on the three official subsets: airline (50 examples), retail (114 examples), and telecom (114 examples). To keep the standard error within 1.5, we report avg@16 on airline and avg@8 on both retail and telecom.
    \item 
    \textbf{Terminal Bench 2.0}\citep{merrill2026terminalbenchbenchmarkingagentshard} is adopted for evaluating agents in terminal-based environments, which comprises of 89 human-validated tasks across specialized fields such as scientific computing, machine learning, and system administration. Moving beyond simple code generation, this benchmark focuses on end-to-end workflows, requiring agents to demonstrate proficiency in holistic operations like model training, system configuration, and software debugging rather than just producing isolated functions. We evaluate the model using the default Terminus-2 scaffolding. We report avg@5 task success rate.
\end{itemize}

For SWE-bench Verified, we use the OpenHands scaffold~\citep{wang2025openhands} as the agentic coding evaluation framework. We adopt a full interaction retention policy for agent trajectories, preserving the complete history of tool calls, observations, and model outputs across turns. This includes prior file views, search results, executed commands, and intermediate patches, enabling the model to maintain state and reason effectively over long-horizon debugging processes. We set the maximum context length to 256K tokens and allow up to 200 turns, consistent with our execution-based agentic SWE-RL training configuration. Notably, this evaluation setup closely mirrors our training environment, as both rely on execution-based feedback and multi-turn interaction within the same tool-augmented scaffold. This alignment reduces train–test mismatch and enables the model to more effectively transfer learned behaviors, such as iterative debugging, hypothesis refinement, and tool-driven reasoning, to the evaluation setting.

For $\tau^2$-Bench evaluation, we adopt a \emph{latest-turn thought retention} policy for managing reasoning traces in multi-turn interactions: we retain the model's reasoning content after the most recent user turn, while discarding reasoning content from earlier turns. The official $\tau^2$-Bench evaluation code follows a \emph{no thought carry-over} policy, which removes all prior reasoning content; in our experiments, this evaluation setup consistently reduces scores by 3--5 points relative to latest-turn thought retention. We attribute this gap to train--test mismatch, since our SFT data for $\tau^2$-style interactions is constructed with the same latest-turn thought retention policy, which is also the thought-state management strategy used in Nemotron-3-Nano-v3 and DeepSeek-V3.2. For the telecom subset, we additionally modify the system prompt to emphasize the dual-control setting by repeating the instruction ``Make sure you guide the user through the steps, do not perform user-side actions yourself.'' three times. We also tested a \emph{full thought retention} policy, which preserves reasoning content from all previous turns and more closely matches RL training, but found it gives similar accuracy to latest-turn thought retention while incurring substantially longer contexts. We therefore report our final $\tau^2$-Bench results using latest-turn thought retention.

\subsection{Multilingual}

For multilingual tasks, we include:

\begin{itemize}[leftmargin=2em]
    \item
    \textbf{MMLU-ProX}~\citep{xuan2025mmluprox} expands the challenging MMLU-Pro benchmark to include 29 languages. Following~\citet{blakeman2025nemotron}, six languages are selected for evaluation: English (en), German (de), Spanish (es), French (fr), Italian (it), and Japanese (ja). The model is evaluated in thinking mode, and we report pass@1 accuracy from a single generation (avg@1).

    \item 
    \textbf{WMT24++}~\citep{deutsch2025wmt24pp} extends the WMT24 machine translation benchmark to cover 55 languages. Following~\citet{blakeman2025nemotron}, we evaluate on five translation pairs: English to German (en → de), English to Spanish (en → es), English to French (en → fr), English to Italian (en → it), and English to Japanese (en → ja). We use XCOMET-XXL~\citep{guerreiro2024xcomet} as the evaluation metric to assess the translation quality. Our model is evaluated in thinking mode, and we report pass@1 accuracy based on a single generation (avg@1).
    
\end{itemize}

\section{Training Hyperparameters}
\label{appendix:training_hyperparams}
We list the training hyperparameters for the Nemotron-Cascade-2-30B-A3B during all stages in Table~\ref{tab:hyperparams_sft}, \ref{tab:hyperparams_coderl}, \ref{tab:hyperparams_sft_execution_swerl}.

%

\begin{table}[!htbp]
\centering
{\footnotesize
\caption{Training hyperparameters for Nemotron-Cascade-2-30B-A3B in SFT.}
\vspace{2pt}
\setlength{\tabcolsep}{20pt} 
\renewcommand{\arraystretch}{1.5} 
\begin{tabular}{lc}
\toprule
Hyperparameters                   &                   \\  \midrule
Global batch size                  & $64$        \\
Packed sequence length                  & $256$K                        
\\
Max learning rate                  & $5\times 10^{-5}$       \\
Min learning rate                  & $5\times 10^{-6}$     \\
Learning rate warmup steps         & 200                                   \\
Scheduler                          & cosine                         \\
Max Steps                          & 40,000                        \\
Optimizer                          & AdamW                          \\
Optimizer config                   & $\beta_1=0.9$, $\beta_2=0.98$  \\
Weight decay                       & $0.1$                          \\
\# of training steps               & 33,000 \\ 
\bottomrule
\end{tabular}
\label{tab:hyperparams_sft}
}
\end{table}

%

\begin{table}[!hbp]
\centering
\caption{Training hyperparameters of Nemotron-Cascade-2-30B-A3B in Cascade RL (IF-RL, Multi-domain RL, MOPD).  }
\vspace{2pt}
{\small
\setlength{\tabcolsep}{20pt} 
\renewcommand{\arraystretch}{1.5} 
\begin{tabular}{l|ccc}
\toprule
Hyper-parameters & IF-RL & Multi-domain RL  & MOPD \\ 
\midrule
Max response length & 49K & 49K & 98K  \\
Batch size & 128  & 128 & 128  \\
\# Rollout size & 16 & 16 & 4 \\
Learning rate & $3\times10^{-6}$  & $3\times10^{-6}$ & $3\times10^{-6}$  \\
Steps & 180 & 70 & 52 \\
         & AdamW  & Adam    & AdamW              \\
Optimizer  & $\beta_1=0.9$ & $\beta_1=0.9$  & $\beta_1=0.9$ \\
& $\beta_2=0.95$ & $\beta_2=0.95$ & $\beta_2=0.95$\\
Temperature & 1.0 & 1.0 & 1.0    \\
Top-p & 1.0 & 1.0 & 1.0 \\
Overlong filtering & False & True & False  \\
\bottomrule
\end{tabular}
}
\label{tab:hyperparams_1}
\end{table}

\begin{table}[!hbp]
\centering
\caption{Training hyperparameters of Nemotron-Cascade-2-30B-A3B in Cascade RL (RLHF, Long-context RL, Code RL).  }
\vspace{2pt}
{\small
\setlength{\tabcolsep}{20pt} 
\renewcommand{\arraystretch}{1.5} 
\begin{tabular}{l|ccc}
\toprule
Hyper-parameters & RLHF & Long-context RL  & Code RL  \\ 
\midrule
Max response length & 16K & 49K & 118K  \\
Batch size & 128  & 128 & 128  \\
\# Rollout size & 16 & 16 & 16 \\
Learning rate & $3\times10^{-6}$  & $3\times10^{-6}$ & $3\times10^{-6}$  \\
Steps & 25 & 30 & 22 \\
         & AdamW  & Adam    & AdamW              \\
Optimizer  & $\beta_1=0.9$ & $\beta_1=0.9$  & $\beta_1=0.9$ \\
& $\beta_2=0.95$ & $\beta_2=0.95$ & $\beta_2=0.95$\\
Temperature & 1.0 & 1.0 & 1.0    \\
Top-p & 1.0 & 1.0 & 0.95  \\
Overlong filtering & True & True & True  \\
\bottomrule
\end{tabular}
}
\label{tab:hyperparams_coderl}
\end{table}


\begin{table}[!htbp]
\centering
{\footnotesize
\caption{Training hyperparameters of Nemotron-Cascade-2-30B-A3B model in execution-based agentic SWE-RL.}
\vspace{2pt}
\setlength{\tabcolsep}{20pt} 
\renewcommand{\arraystretch}{1.5} 
\begin{tabular}{lc}
\toprule
Hyperparameters                   &                          \\  \midrule
\# prompts per step                  & $16$                        \\
\# rollout                  & $64$                        \\
Temperature                  & $0.8$                        \\
Max sequence length                  & $256k$                        \\
Max turn                  & $200$                        \\
Max learning rate                  & $3\times 10^{-6}$        \\
Min learning rate                  & $0$    \\
Learning rate warmup steps         & 10                                   \\
\bottomrule
\end{tabular}
\label{tab:hyperparams_sft_execution_swerl}
}
\end{table}


\section{Prompt Templates}

\subsection{Prompt Templates for Test-Time Scaling on IOI 2025}
\label{appendix:cp_template}

\begin{promptbox}
Write Python code to solve the problem. Please place the solution code in the following format:\texttt{\\```python\\ \# Your solution code here\\```}
\\
\texttt{\{problem\_statement\}}

Below you are provided the accepted correct solutions but with different input constraints. You may use them as a reference for your insights.
\\ 
=======================
\\ 
\#\# Different Constraints (for reference only):
\\ 
\texttt{\{subtask\_constraints\}}
\\ 
\#\#\# Accepted Code:
\\ 
\texttt{[CODE]}
\\
=======================
\\
\#\# Different Constraints (for reference only):
\\ 
...
\\ 
=======================
\\ 
From here, you are also given your submission history containing **incorrect** code and their corresponding official judgement verdicts as reference -- Official judgement verdicts and problem statement/conditions are 100\% reliable. You should make improvements from them if they could help:
\\
=======================
\\
\#\#\# Incorrect Code
\\ 
\texttt{[CODE]}
\\
Judgement Verdict: \texttt{[VERDICT]}, Score: \texttt{[SCORE]}
\\
=======================
\\
\#\#\# Incorrect Code
\\
... 
\\ 
=======================
\end{promptbox}

\subsection{HLE Judge Prompt}
\label{appendix:hle_judge_prompt}

\begin{promptbox}
\small
Judge whether the following [response] to [question] is correct or not based on the precise and unambiguous [correct\_answer] below.\\
\\
\texttt{[question]: \{question\}}\\
\\
\texttt{[response]: \{response\}}\\
\\
Your judgement must be in the format and criteria specified below:\\
\\
extracted\_final\_answer: The final exact answer extracted from the [response]. Put the extracted answer as 'None' if there is no exact, final answer to extract from the response.\\
\\
\texttt{[correct\_answer]: \{correct\_answer\}}\\
\\
reasoning: Explain why the extracted\_final\_answer is correct or incorrect based on [correct\_answer], focusing only on if there are meaningful differences between [correct\_answer] and the extracted\_final\_answer. Do not comment on any background to the problem, do not attempt to solve the problem, do not argue for any answer different than [correct\_answer], focus only on whether the answers match.\\
\\
correct: Answer 'yes' if extracted\_final\_answer matches the [correct\_answer] given above, or is within a small margin of error for numerical problems. Answer 'no' otherwise, i.e. if there if there is any inconsistency, ambiguity, non-equivalency, or if the extracted answer is incorrect.\\
\\
confidence: The extracted confidence score between 0|\%| and 100|\%| from [response]. Put 100 if there is no confidence score available.
\end{promptbox}

\section{ELO Rating Analysis}
\label{appendix:elo_rating_on_codeforces}

We perform ELO rating analysis on our Nemotron-Cascade-2-30B-A3B model based on \textbf{40} recent Div.1 and Div.2 Codeforces contests held between 2501–2507. Problems and evaluations are provided by LiveCodeBench Pro~\citep{zheng2025livecodebench}. We adopt similar rating estimation approach as in~\cite{wang2025nemotron}, by allowing model with up to $N=8$ submissions to each contest problems, estimating model performance and relative ranking to human contestants with expected penalty consideration. We generate the model’s responses using a temperature of 1.0, \texttt{top-p} of 0.95, and a maximum token budget of 128K. The performance details of our Nemotron-Cascade-2-30B-A3B model (\textbf{with} and \textbf{without} python-tool use) can be found in Table~\ref{tab:elo_rating} and Table~\ref{tab:elo_rating_tir}, respectively.

We observed our model's strong code reasoning ability on solving really tough problems and achieving high ranking even on some Div. 1 rounds (Round 999, 1012, 1015, 1021 etc.), while maintaining stable performance on solving easy-medium level problems. However, the models still has weakness on dealing with problems that requiring constructive algorithms, interactive manner, and hypothesis-driven ideas.


\newcolumntype{P}[1]{>{\centering\arraybackslash}m{#1}}

\begin{table}[!htbp]
\centering
\caption{Nemotron-Cascade-2-30B-A3B performance details on 40 Div.1 and Div.2 Codeforces Rounds ranging from 2501 to 2507 \textbf{without} python-tool use. We attempt each problem with $N=8$ times in total. For regular codeforces rounds, we present the score after considering expected penalties for each problem. For ICPC style rounds, we mark passed/failed problems as \textcolor{ForestGreen}{+} and \textcolor{BrickRed}{-} correspondingly. We compute the estimated rank to human contestants and the corresponding Elo score as shown in rightmost two columns.}
\scalebox{0.47}{
\begin{tabular}{c| *{11}{P{1.1cm}} |c c c c}
\toprule
\bfseries Contest Name & \multicolumn{11}{c|}{\bfseries Contest Problems} &
\bfseries Score & \bfseries Penalty & \bfseries Est. Rank & \bfseries ELO \\ \midrule
\multirow{2}{*}{Hello 2025} & \bf A  & \bf B  & \bf C  & \bf D  & \bf E1  & \bf E2  & \bf F  & \bf G  & \bf H  &  &  &  \multirow{2}{*}{10779.46} & \multirow{2}{*}{-} & \multirow{2}{*}{13/16703} & \multirow{2}{*}{3449} \\
&  \color{ForestGreen} 500.00 &  \color{ForestGreen} 1000.00 &  \color{ForestGreen} 1493.75 &  \color{ForestGreen} 2235.71 & \color{BrickRed} 0.0 &  \color{ForestGreen} 1900.00 & \color{BrickRed} 0.0 &  \color{ForestGreen} 3650.00 & \color{BrickRed} 0.0  &  &  &  \\ \hline
\multirow{2}{*}{Codeforces Round 996 (Div. 2)} & \bf A  & \bf B  & \bf C  & \bf D  & \bf E  & \bf F  &  &  &  &  &  &  \multirow{2}{*}{5793.75} & \multirow{2}{*}{-} & \multirow{2}{*}{2/21232} & \multirow{2}{*}{2198} \\
&  \color{ForestGreen} 500.00 &  \color{ForestGreen} 993.75 &  \color{ForestGreen} 1475.00 & \color{BrickRed} 0.0 &  \color{ForestGreen} 2825.00 & \color{BrickRed} 0.0  &  &  &  &  &  &  \\ \hline
\multirow{2}{*}{Codeforces Round 997 (Div. 2)} & \bf A  & \bf B  & \bf C  & \bf D  & \bf E  & \bf F1  & \bf F2  &  &  &  &  &  \multirow{2}{*}{9378.75} & \multirow{2}{*}{-} & \multirow{2}{*}{1/18823} & \multirow{2}{*}{2198} \\
&  \color{ForestGreen} 493.75 &  \color{ForestGreen} 1250.00 &  \color{ForestGreen} 1475.00 & \color{BrickRed} 0.0 &  \color{ForestGreen} 2225.00 &  \color{ForestGreen} 2710.00 &  \color{ForestGreen} 1225.00  &  &  &  &  &  \\ \hline
\multirow{2}{*}{IAEPC Preliminary Contest (Codeforces Round 999, Div. 1 + Div. 2)} & \bf A  & \bf B  & \bf C  & \bf D  & \bf E  & \bf F1  & \bf F2  & \bf G  & \bf H1  & \bf H2  & \bf I  &  \multirow{2}{*}{9278.75} & \multirow{2}{*}{-} & \multirow{2}{*}{43/12647} & \multirow{2}{*}{3076} \\
&  \color{ForestGreen} 500.00 &  \color{ForestGreen} 1000.00 &  \color{ForestGreen} 1500.00 &  \color{ForestGreen} 1493.75 &  \color{ForestGreen} 1960.00 & \color{BrickRed} 0.0 & \color{BrickRed} 0.0 & \color{BrickRed} 0.0 &  \color{ForestGreen} 2825.00 & \color{BrickRed} 0.0 & \color{BrickRed} 0.0  &  \\ \hline
\multirow{2}{*}{Codeforces Round 1000 (Div. 2)} & \bf A  & \bf B  & \bf C  & \bf D  & \bf E  & \bf F1  & \bf F2  &  &  &  &  &  \multirow{2}{*}{10935.71} & \multirow{2}{*}{-} & \multirow{2}{*}{1/17169} & \multirow{2}{*}{2200} \\
&  \color{ForestGreen} 500.00 &  \color{ForestGreen} 985.71 &  \color{ForestGreen} 1500.00 &  \color{ForestGreen} 2250.00 &  \color{ForestGreen} 2687.50 &  \color{ForestGreen} 1687.50 &  \color{ForestGreen} 1325.00  &  &  &  &  &  \\ \hline
\multirow{2}{*}{Ethflow Round 1 (Codeforces Round 1001, Div. 1 + Div. 2)} & \bf A  & \bf B  & \bf C  & \bf D  & \bf E1  & \bf E2  & \bf F  & \bf G  & \bf H  &  &  &  \multirow{2}{*}{2493.75} & \multirow{2}{*}{-} & \multirow{2}{*}{1727/16234} & \multirow{2}{*}{1898} \\
&  \color{ForestGreen} 500.00 &  \color{ForestGreen} 993.75 &  \color{ForestGreen} 1000.00 & \color{BrickRed} 0.0 & \color{BrickRed} 0.0 & \color{BrickRed} 0.0 & \color{BrickRed} 0.0 & \color{BrickRed} 0.0 & \color{BrickRed} 0.0  &  &  &  \\ \hline
\multirow{2}{*}{Codeforces Round 1002 (Div. 2)} & \bf A  & \bf B  & \bf C  & \bf D  & \bf E1  & \bf E2  &  &  &  &  &  &  \multirow{2}{*}{3300.00} & \multirow{2}{*}{-} & \multirow{2}{*}{1102/19443} & \multirow{2}{*}{1882} \\
&  \color{ForestGreen} 500.00 &  \color{ForestGreen} 975.00 & \color{BrickRed} 0.0 &  \color{ForestGreen} 1825.00 & \color{BrickRed} 0.0 & \color{BrickRed} 0.0  &  &  &  &  &  &  \\ \hline
\multirow{2}{*}{Codeforces Round 1004 (Div. 1)} & \bf A  & \bf B  & \bf C  & \bf D1  & \bf D2  & \bf E  & \bf F  &  &  &  &  &  \multirow{2}{*}{2681.25} & \multirow{2}{*}{-} & \multirow{2}{*}{145/1030} & \multirow{2}{*}{2666} \\
& \color{BrickRed} 0.0 &  \color{ForestGreen} 687.50 &  \color{ForestGreen} 1250.00 &  \color{ForestGreen} 743.75 & \color{BrickRed} 0.0 & \color{BrickRed} 0.0 & \color{BrickRed} 0.0  &  &  &  &  &  \\ \hline
\multirow{2}{*}{Codeforces Round 1004 (Div. 2)} & \bf A  & \bf B  & \bf C  & \bf D  & \bf E  & \bf F  & \bf G  &  &  &  &  &  \multirow{2}{*}{5397.50} & \multirow{2}{*}{-} & \multirow{2}{*}{8/16749} & \multirow{2}{*}{2098} \\
&  \color{ForestGreen} 500.00 &  \color{ForestGreen} 960.00 & \color{BrickRed} 0.0 & \color{BrickRed} 0.0 &  \color{ForestGreen} 1687.50 &  \color{ForestGreen} 2250.00 & \color{BrickRed} 0.0  &  &  &  &  &  \\ \hline
\multirow{2}{*}{Codeforces Round 1005 (Div. 2)} & \bf A  & \bf B  & \bf C  & \bf D  & \bf E  & \bf F  &  &  &  &  &  &  \multirow{2}{*}{9198.21} & \multirow{2}{*}{-} & \multirow{2}{*}{1/17621} & \multirow{2}{*}{2260} \\
&  \color{ForestGreen} 493.75 &  \color{ForestGreen} 1000.00 &  \color{ForestGreen} 1243.75 &  \color{ForestGreen} 1735.71 &  \color{ForestGreen} 2075.00 &  \color{ForestGreen} 2650.00  &  &  &  &  &  &  \\ \hline
\multirow{2}{*}{Educational Codeforces Round 174 (Rated for Div. 2)} & \bf A  & \bf B  & \bf C  & \bf D  & \bf E  & \bf F  &  &  &  &  &  &  \multirow{2}{*}{4} & \multirow{2}{*}{2.86} & \multirow{2}{*}{156/16701} & \multirow{2}{*}{2242} \\
& \color{ForestGreen} + & \color{ForestGreen} + & \color{ForestGreen} + & \color{ForestGreen} + & \color{BrickRed} - & \color{BrickRed} -  &  &  &  &  &  &  \\ \hline
\multirow{2}{*}{Educational Codeforces Round 175 (Rated for Div. 2)} & \bf A  & \bf B  & \bf C  & \bf D  & \bf E  & \bf F  &  &  &  &  &  &  \multirow{2}{*}{4} & \multirow{2}{*}{0.00} & \multirow{2}{*}{234/16060} & \multirow{2}{*}{2195} \\
& \color{ForestGreen} + & \color{ForestGreen} + & \color{ForestGreen} + & \color{ForestGreen} + & \color{BrickRed} - & \color{BrickRed} -  &  &  &  &  &  &  \\ \hline
\multirow{2}{*}{Codeforces Round 1007 (Div. 2)} & \bf A  & \bf B  & \bf C  & \bf D1  & \bf D2  & \bf E  & \bf F  &  &  &  &  &  \multirow{2}{*}{8429.46} & \multirow{2}{*}{-} & \multirow{2}{*}{1/16254} & \multirow{2}{*}{2198} \\
&  \color{ForestGreen} 500.00 &  \color{ForestGreen} 1000.00 &  \color{ForestGreen} 1485.71 &  \color{ForestGreen} 1743.75 &  \color{ForestGreen} 1225.00 &  \color{ForestGreen} 2475.00 & \color{BrickRed} 0.0  &  &  &  &  &  \\ \hline
\multirow{2}{*}{Codeforces Round 1008 (Div. 1)} & \bf A  & \bf B  & \bf C  & \bf D  & \bf E  & \bf F  & \bf G  &  &  &  &  &  \multirow{2}{*}{2000.00} & \multirow{2}{*}{-} & \multirow{2}{*}{355/909} & \multirow{2}{*}{2312} \\
&  \color{ForestGreen} 500.00 & \color{BrickRed} 0.0 &  \color{ForestGreen} 1500.00 & \color{BrickRed} 0.0 & \color{BrickRed} 0.0 & \color{BrickRed} 0.0 & \color{BrickRed} 0.0  &  &  &  &  &  \\ \hline
\multirow{2}{*}{Codeforces Round 1008 (Div. 2)} & \bf A  & \bf B  & \bf C  & \bf D  & \bf E  & \bf F  & \bf G  &  &  &  &  &  \multirow{2}{*}{6825.00} & \multirow{2}{*}{-} & \multirow{2}{*}{9/14641} & \multirow{2}{*}{2008} \\
&  \color{ForestGreen} 500.00 &  \color{ForestGreen} 750.00 &  \color{ForestGreen} 1250.00 &  \color{ForestGreen} 1575.00 & \color{BrickRed} 0.0 &  \color{ForestGreen} 2750.00 & \color{BrickRed} 0.0  &  &  &  &  &  \\ \hline
\multirow{2}{*}{Educational Codeforces Round 176 (Rated for Div. 2)} & \bf A  & \bf B  & \bf C  & \bf D  & \bf E  & \bf F  &  &  &  &  &  &  \multirow{2}{*}{5} & \multirow{2}{*}{10.86} & \multirow{2}{*}{2/18159} & \multirow{2}{*}{2198} \\
& \color{ForestGreen} + & \color{ForestGreen} + & \color{ForestGreen} + & \color{ForestGreen} + & \color{ForestGreen} + & \color{BrickRed} -  &  &  &  &  &  &  \\ \hline
\multirow{2}{*}{Codeforces Round 1011 (Div. 2)} & \bf A  & \bf B  & \bf C  & \bf D  & \bf E  & \bf F1  & \bf F2  &  &  &  &  &  \multirow{2}{*}{10137.50} & \multirow{2}{*}{-} & \multirow{2}{*}{1/15906} & \multirow{2}{*}{2200} \\
&  \color{ForestGreen} 500.00 &  \color{ForestGreen} 1250.00 &  \color{ForestGreen} 1250.00 &  \color{ForestGreen} 1743.75 &  \color{ForestGreen} 2500.00 &  \color{ForestGreen} 1993.75 &  \color{ForestGreen} 900.00  &  &  &  &  &  \\ \hline
\multirow{2}{*}{Codeforces Round 1012 (Div. 1)} & \bf A  & \bf B1  & \bf B2  & \bf C1  & \bf C2  & \bf D  & \bf E  &  &  &  &  &  \multirow{2}{*}{2985.00} & \multirow{2}{*}{-} & \multirow{2}{*}{24/653} & \multirow{2}{*}{3057} \\
&  \color{ForestGreen} 710.00 &  \color{ForestGreen} 975.00 &  \color{ForestGreen} 325.00 &  \color{ForestGreen} 975.00 & \color{BrickRed} 0.0 & \color{BrickRed} 0.0 & \color{BrickRed} 0.0  &  &  &  &  &  \\ \hline
\multirow{2}{*}{Codeforces Round 1012 (Div. 2)} & \bf A  & \bf B  & \bf C  & \bf D  & \bf E1  & \bf E2  & \bf F1  & \bf F2  &  &  &  &  \multirow{2}{*}{9945.00} & \multirow{2}{*}{-} & \multirow{2}{*}{1/8536} & \multirow{2}{*}{2007} \\
&  \color{ForestGreen} 500.00 &  \color{ForestGreen} 960.00 &  \color{ForestGreen} 1750.00 &  \color{ForestGreen} 1960.00 &  \color{ForestGreen} 1975.00 &  \color{ForestGreen} 825.00 &  \color{ForestGreen} 1975.00 & \color{BrickRed} 0.0  &  &  &  &  \\ \hline
\multirow{2}{*}{Codeforces Round 1014 (Div. 2)} & \bf A  & \bf B  & \bf C  & \bf D  & \bf E  & \bf F  &  &  &  &  &  &  \multirow{2}{*}{6500.00} & \multirow{2}{*}{-} & \multirow{2}{*}{2/15842} & \multirow{2}{*}{2213} \\
&  \color{ForestGreen} 500.00 &  \color{ForestGreen} 750.00 &  \color{ForestGreen} 1250.00 &  \color{ForestGreen} 1750.00 &  \color{ForestGreen} 2250.00 & \color{BrickRed} 0.0  &  &  &  &  &  &  \\ \hline
\multirow{2}{*}{Teza Round 1 (Codeforces Round 1015, Div. 1 + Div. 2)} & \bf A  & \bf B  & \bf C  & \bf D  & \bf E  & \bf F  & \bf G1  & \bf G2  & \bf H  &  &  &  \multirow{2}{*}{12521.43} & \multirow{2}{*}{-} & \multirow{2}{*}{4/11206} & \multirow{2}{*}{3830} \\
&  \color{ForestGreen} 750.00 &  \color{ForestGreen} 1000.00 &  \color{ForestGreen} 1500.00 &  \color{ForestGreen} 1735.71 &  \color{ForestGreen} 2235.71 &  \color{ForestGreen} 2825.00 &  \color{ForestGreen} 2475.00 & \color{BrickRed} 0.0 & \color{BrickRed} 0.0  &  &  &  \\ \hline
\multirow{2}{*}{Neowise Labs Contest 1 (Codeforces Round 1018, Div. 1 + Div. 2)} & \bf A  & \bf B  & \bf C  & \bf D  & \bf E  & \bf F  & \bf G  & \bf H  &  &  &  &  \multirow{2}{*}{4400.00} & \multirow{2}{*}{-} & \multirow{2}{*}{493/12771} & \multirow{2}{*}{2312} \\
&  \color{ForestGreen} 500.00 &  \color{ForestGreen} 750.00 &  \color{ForestGreen} 1500.00 &  \color{ForestGreen} 1650.00 & \color{BrickRed} 0.0 & \color{BrickRed} 0.0 & \color{BrickRed} 0.0 & \color{BrickRed} 0.0  &  &  &  &  \\ \hline
\multirow{2}{*}{Codeforces Round 1019 (Div. 2)} & \bf A  & \bf B  & \bf C  & \bf D  & \bf E  & \bf F  &  &  &  &  &  &  \multirow{2}{*}{4825.00} & \multirow{2}{*}{-} & \multirow{2}{*}{47/14465} & \multirow{2}{*}{2202} \\
&  \color{ForestGreen} 500.00 &  \color{ForestGreen} 1000.00 &  \color{ForestGreen} 1500.00 &  \color{ForestGreen} 1825.00 & \color{BrickRed} 0.0 & \color{BrickRed} 0.0  &  &  &  &  &  &  \\ \hline
\multirow{2}{*}{Codeforces Round 1021 (Div. 1)} & \bf A  & \bf B  & \bf C  & \bf D  & \bf E  & \bf F  &  &  &  &  &  &  \multirow{2}{*}{3218.75} & \multirow{2}{*}{-} & \multirow{2}{*}{75/651} & \multirow{2}{*}{2760} \\
&  \color{ForestGreen} 493.75 &  \color{ForestGreen} 900.00 & \color{BrickRed} 0.0 &  \color{ForestGreen} 1825.00 & \color{BrickRed} 0.0 & \color{BrickRed} 0.0  &  &  &  &  &  &  \\ \hline
\multirow{2}{*}{Codeforces Round 1021 (Div. 2)} & \bf A  & \bf B  & \bf C  & \bf D  & \bf E  & \bf F  &  &  &  &  &  &  \multirow{2}{*}{8468.75} & \multirow{2}{*}{-} & \multirow{2}{*}{1/5824} & \multirow{2}{*}{2019} \\
&  \color{ForestGreen} 500.00 &  \color{ForestGreen} 1250.00 &  \color{ForestGreen} 1493.75 &  \color{ForestGreen} 2150.00 & \color{BrickRed} 0.0 &  \color{ForestGreen} 3075.00  &  &  &  &  &  &  \\ \hline
\multirow{2}{*}{Educational Codeforces Round 178 (Rated for Div. 2)} & \bf A  & \bf B  & \bf C  & \bf D  & \bf E  & \bf F  & \bf G  &  &  &  &  &  \multirow{2}{*}{6} & \multirow{2}{*}{12.50} & \multirow{2}{*}{4/11706} & \multirow{2}{*}{2215} \\
& \color{ForestGreen} + & \color{ForestGreen} + & \color{ForestGreen} + & \color{ForestGreen} + & \color{ForestGreen} + & \color{ForestGreen} + & \color{BrickRed} -  &  &  &  &  &  \\ \hline
\multirow{2}{*}{Codeforces Round 1022 (Div. 2)} & \bf A  & \bf B  & \bf C  & \bf D  & \bf E  & \bf F  &  &  &  &  &  &  \multirow{2}{*}{3087.50} & \multirow{2}{*}{-} & \multirow{2}{*}{308/11127} & \multirow{2}{*}{2132} \\
&  \color{ForestGreen} 500.00 &  \color{ForestGreen} 1187.50 &  \color{ForestGreen} 1400.00 & \color{BrickRed} 0.0 & \color{BrickRed} 0.0 & \color{BrickRed} 0.0  &  &  &  &  &  &  \\ \hline
\multirow{2}{*}{Codeforces Round 1023 (Div. 2)} & \bf A  & \bf B  & \bf C  & \bf D  & \bf E  & \bf F1  & \bf F2  &  &  &  &  &  \multirow{2}{*}{6506.25} & \multirow{2}{*}{-} & \multirow{2}{*}{6/11636} & \multirow{2}{*}{2209} \\
&  \color{ForestGreen} 250.00 &  \color{ForestGreen} 750.00 &  \color{ForestGreen} 1493.75 &  \color{ForestGreen} 1937.50 & \color{BrickRed} 0.0 &  \color{ForestGreen} 2075.00 & \color{BrickRed} 0.0  &  &  &  &  &  \\ \hline
\multirow{2}{*}{Codeforces Round 1024 (Div. 1)} & \bf A  & \bf B  & \bf C  & \bf D  & \bf E  & \bf F  &  &  &  &  &  &  \multirow{2}{*}{1729.46} & \multirow{2}{*}{-} & \multirow{2}{*}{477/857} & \multirow{2}{*}{2149} \\
&  \color{ForestGreen} 485.71 &  \color{ForestGreen} 1243.75 & \color{BrickRed} 0.0 & \color{BrickRed} 0.0 & \color{BrickRed} 0.0 & \color{BrickRed} 0.0  &  &  &  &  &  &  \\ \hline
\multirow{2}{*}{Codeforces Round 1024 (Div. 2)} & \bf A  & \bf B  & \bf C  & \bf D  & \bf E  & \bf F  &  &  &  &  &  &  \multirow{2}{*}{3479.46} & \multirow{2}{*}{-} & \multirow{2}{*}{34/11201} & \multirow{2}{*}{1998} \\
&  \color{ForestGreen} 250.00 &  \color{ForestGreen} 500.00 &  \color{ForestGreen} 985.71 &  \color{ForestGreen} 1743.75 & \color{BrickRed} 0.0 & \color{BrickRed} 0.0  &  &  &  &  &  &  \\ \hline
\multirow{2}{*}{Codeforces Round 1025 (Div. 2)} & \bf A  & \bf B  & \bf C1  & \bf C2  & \bf C3  & \bf D  & \bf E  & \bf F  &  &  &  &  \multirow{2}{*}{7985.71} & \multirow{2}{*}{-} & \multirow{2}{*}{1/15945} & \multirow{2}{*}{2197} \\
&  \color{ForestGreen} 500.00 &  \color{ForestGreen} 985.71 &  \color{ForestGreen} 1243.75 &  \color{ForestGreen} 575.00 &  \color{ForestGreen} 500.00 &  \color{ForestGreen} 1687.50 &  \color{ForestGreen} 2493.75 & \color{BrickRed} 0.0  &  &  &  &  \\ \hline
\multirow{2}{*}{Codeforces Round 1026 (Div. 2)} & \bf A  & \bf B  & \bf C  & \bf D  & \bf E  & \bf F  &  &  &  &  &  &  \multirow{2}{*}{9897.50} & \multirow{2}{*}{-} & \multirow{2}{*}{1/17668} & \multirow{2}{*}{2198} \\
&  \color{ForestGreen} 500.00 &  \color{ForestGreen} 750.00 &  \color{ForestGreen} 1500.00 &  \color{ForestGreen} 1960.00 &  \color{ForestGreen} 2250.00 &  \color{ForestGreen} 2937.50  &  &  &  &  &  &  \\ \hline
\multirow{2}{*}{Codeforces Round 1028 (Div. 1)} & \bf A  & \bf B  & \bf C  & \bf D  & \bf E  & \bf F1  & \bf F2  &  &  &  &  &  \multirow{2}{*}{2710.00} & \multirow{2}{*}{-} & \multirow{2}{*}{75/956} & \multirow{2}{*}{2865} \\
&  \color{ForestGreen} 500.00 & \color{BrickRed} 0.0 & \color{BrickRed} 0.0 &  \color{ForestGreen} 2210.00 & \color{BrickRed} 0.0 & \color{BrickRed} 0.0 & \color{BrickRed} 0.0  &  &  &  &  &  \\ \hline
\multirow{2}{*}{Codeforces Round 1028 (Div. 2)} & \bf A  & \bf B  & \bf C  & \bf D  & \bf E  & \bf F  &  &  &  &  &  &  \multirow{2}{*}{5453.75} & \multirow{2}{*}{-} & \multirow{2}{*}{4/18314} & \multirow{2}{*}{2018} \\
&  \color{ForestGreen} 493.75 &  \color{ForestGreen} 750.00 &  \color{ForestGreen} 1250.00 & \color{BrickRed} 0.0 & \color{BrickRed} 0.0 &  \color{ForestGreen} 2960.00  &  &  &  &  &  &  \\ \hline
\multirow{2}{*}{Educational Codeforces Round 179 (Rated for Div. 2)} & \bf A  & \bf B  & \bf C  & \bf D  & \bf E  & \bf F  & \bf G  &  &  &  &  &  \multirow{2}{*}{5} & \multirow{2}{*}{60.00} & \multirow{2}{*}{94/12301} & \multirow{2}{*}{2231} \\
& \color{ForestGreen} + & \color{BrickRed} - & \color{ForestGreen} + & \color{ForestGreen} + & \color{ForestGreen} + & \color{ForestGreen} + & \color{BrickRed} -  &  &  &  &  &  \\ \hline
\multirow{2}{*}{Codeforces Round 1030 (Div. 2)} & \bf A  & \bf B  & \bf C  & \bf D1  & \bf D2  & \bf E  & \bf F  &  &  &  &  &  \multirow{2}{*}{7003.75} & \multirow{2}{*}{-} & \multirow{2}{*}{2/18335} & \multirow{2}{*}{2205} \\
&  \color{ForestGreen} 500.00 &  \color{ForestGreen} 975.00 &  \color{ForestGreen} 1000.00 &  \color{ForestGreen} 1243.75 &  \color{ForestGreen} 960.00 &  \color{ForestGreen} 2325.00 & \color{BrickRed} 0.0  &  &  &  &  &  \\ \hline
\multirow{2}{*}{Codeforces Round 1031 (Div. 2)} & \bf A  & \bf B  & \bf C  & \bf D  & \bf E  & \bf F  &  &  &  &  &  &  \multirow{2}{*}{4060.71} & \multirow{2}{*}{-} & \multirow{2}{*}{20/11032} & \multirow{2}{*}{2216} \\
&  \color{ForestGreen} 500.00 &  \color{ForestGreen} 735.71 & \color{BrickRed} 0.0 & \color{BrickRed} 0.0 & \color{BrickRed} 0.0 &  \color{ForestGreen} 2825.00  &  &  &  &  &  &  \\ \hline
\multirow{2}{*}{Codeforces Round 1033 (Div. 2) and CodeNite 2025} & \bf A  & \bf B  & \bf C  & \bf D  & \bf E  & \bf F  & \bf G  &  &  &  &  &  \multirow{2}{*}{9623.21} & \multirow{2}{*}{-} & \multirow{2}{*}{1/12948} & \multirow{2}{*}{2216} \\
&  \color{ForestGreen} 493.75 &  \color{ForestGreen} 750.00 &  \color{ForestGreen} 1250.00 &  \color{ForestGreen} 1735.71 &  \color{ForestGreen} 2493.75 &  \color{ForestGreen} 2900.00 & \color{BrickRed} 0.0  &  &  &  &  &  \\ \hline
\multirow{2}{*}{Educational Codeforces Round 180 (Rated for Div. 2)} & \bf A  & \bf B  & \bf C  & \bf D  & \bf E  & \bf F  &  &  &  &  &  &  \multirow{2}{*}{5} & \multirow{2}{*}{33.75} & \multirow{2}{*}{8/17128} & \multirow{2}{*}{2253} \\
& \color{ForestGreen} + & \color{ForestGreen} + & \color{ForestGreen} + & \color{ForestGreen} + & \color{ForestGreen} + & \color{BrickRed} -  &  &  &  &  &  &  \\ \hline
\multirow{2}{*}{Codeforces Round 1035 (Div. 2)} & \bf A  & \bf B  & \bf C  & \bf D  & \bf E  & \bf F  &  &  &  &  &  &  \multirow{2}{*}{2985.71} & \multirow{2}{*}{-} & \multirow{2}{*}{587/15624} & \multirow{2}{*}{2008} \\
&  \color{ForestGreen} 500.00 &  \color{ForestGreen} 1000.00 &  \color{ForestGreen} 1485.71 & \color{BrickRed} 0.0 & \color{BrickRed} 0.0 & \color{BrickRed} 0.0  &  &  &  &  &  &  \\
\bottomrule
\end{tabular}}
\label{tab:elo_rating}
\end{table}

\begin{table}[!htbp]
\centering
\caption{Nemotron-Cascade-2-30B-A3B performance details on 40 Div.1 and Div.2 Codeforces Rounds ranging from 2501 to 2507 \textbf{with} python-tool use. We attempt each problem with $N=8$ times in total. For regular codeforces rounds, we present the score after considering expected penalties for each problem. For ICPC style rounds, we mark passed/failed problems as \textcolor{ForestGreen}{+} and \textcolor{BrickRed}{-} correspondingly. We compute the estimated rank to human contestants and the corresponding Elo score as shown in rightmost two columns.}
\scalebox{0.47}{
\begin{tabular}{c| *{11}{P{1.1cm}} |c c c c}
\toprule
\bfseries Contest Name & \multicolumn{11}{c|}{\bfseries Contest Problems} &
\bfseries Score & \bfseries Penalty & \bfseries Est. Rank & \bfseries ELO \\ \midrule
\multirow{2}{*}{Hello 2025} & \bf A  & \bf B  & \bf C  & \bf D  & \bf E1  & \bf E2  & \bf F  & \bf G  & \bf H  &  &  &  \multirow{2}{*}{11712.50} & \multirow{2}{*}{-} & \multirow{2}{*}{11/16703} & \multirow{2}{*}{3497} \\
&  \color{ForestGreen} 500.00 &  \color{ForestGreen} 1000.00 &  \color{ForestGreen} 1500.00 &  \color{ForestGreen} 2225.00 &  \color{ForestGreen} 937.50 &  \color{ForestGreen} 1900.00 & \color{BrickRed} 0.0 &  \color{ForestGreen} 3650.00 & \color{BrickRed} 0.0  &  &  &  \\ \hline
\multirow{2}{*}{Codeforces Round 996 (Div. 2)} & \bf A  & \bf B  & \bf C  & \bf D  & \bf E  & \bf F  &  &  &  &  &  &  \multirow{2}{*}{5025.00} & \multirow{2}{*}{-} & \multirow{2}{*}{2/21232} & \multirow{2}{*}{2198} \\
&  \color{ForestGreen} 500.00 &  \color{ForestGreen} 975.00 &  \color{ForestGreen} 1475.00 &  \color{ForestGreen} 2075.00 & \color{BrickRed} 0.0 & \color{BrickRed} 0.0  &  &  &  &  &  &  \\ \hline
\multirow{2}{*}{Codeforces Round 997 (Div. 2)} & \bf A  & \bf B  & \bf C  & \bf D  & \bf E  & \bf F1  & \bf F2  &  &  &  &  &  \multirow{2}{*}{11187.50} & \multirow{2}{*}{-} & \multirow{2}{*}{1/18823} & \multirow{2}{*}{2198} \\
&  \color{ForestGreen} 493.75 &  \color{ForestGreen} 1250.00 &  \color{ForestGreen} 1500.00 &  \color{ForestGreen} 1900.00 &  \color{ForestGreen} 2075.00 &  \color{ForestGreen} 2743.75 &  \color{ForestGreen} 1225.00  &  &  &  &  &  \\ \hline
\multirow{2}{*}{IAEPC Preliminary Contest (Codeforces Round 999, Div. 1 + Div. 2)} & \bf A  & \bf B  & \bf C  & \bf D  & \bf E  & \bf F1  & \bf F2  & \bf G  & \bf H1  & \bf H2  & \bf I  &  \multirow{2}{*}{9416.96} & \multirow{2}{*}{-} & \multirow{2}{*}{40/12647} & \multirow{2}{*}{3097} \\
&  \color{ForestGreen} 500.00 &  \color{ForestGreen} 1000.00 &  \color{ForestGreen} 1493.75 &  \color{ForestGreen} 1485.71 &  \color{ForestGreen} 2000.00 & \color{BrickRed} 0.0 & \color{BrickRed} 0.0 & \color{BrickRed} 0.0 &  \color{ForestGreen} 2937.50 & \color{BrickRed} 0.0 & \color{BrickRed} 0.0  &  \\ \hline
\multirow{2}{*}{Codeforces Round 1000 (Div. 2)} & \bf A  & \bf B  & \bf C  & \bf D  & \bf E  & \bf F1  & \bf F2  &  &  &  &  &  \multirow{2}{*}{10981.25} & \multirow{2}{*}{-} & \multirow{2}{*}{1/17169} & \multirow{2}{*}{2200} \\
&  \color{ForestGreen} 500.00 &  \color{ForestGreen} 1000.00 &  \color{ForestGreen} 1500.00 &  \color{ForestGreen} 2243.75 &  \color{ForestGreen} 2725.00 &  \color{ForestGreen} 1687.50 &  \color{ForestGreen} 1325.00  &  &  &  &  &  \\ \hline
\multirow{2}{*}{Ethflow Round 1 (Codeforces Round 1001, Div. 1 + Div. 2)} & \bf A  & \bf B  & \bf C  & \bf D  & \bf E1  & \bf E2  & \bf F  & \bf G  & \bf H  &  &  &  \multirow{2}{*}{2493.75} & \multirow{2}{*}{-} & \multirow{2}{*}{1727/16234} & \multirow{2}{*}{1898} \\
&  \color{ForestGreen} 500.00 &  \color{ForestGreen} 993.75 &  \color{ForestGreen} 1000.00 & \color{BrickRed} 0.0 & \color{BrickRed} 0.0 & \color{BrickRed} 0.0 & \color{BrickRed} 0.0 & \color{BrickRed} 0.0 & \color{BrickRed} 0.0  &  &  &  \\ \hline
\multirow{2}{*}{Codeforces Round 1002 (Div. 2)} & \bf A  & \bf B  & \bf C  & \bf D  & \bf E1  & \bf E2  &  &  &  &  &  &  \multirow{2}{*}{3300.00} & \multirow{2}{*}{-} & \multirow{2}{*}{1102/19443} & \multirow{2}{*}{1882} \\
&  \color{ForestGreen} 500.00 &  \color{ForestGreen} 975.00 & \color{BrickRed} 0.0 &  \color{ForestGreen} 1825.00 & \color{BrickRed} 0.0 & \color{BrickRed} 0.0  &  &  &  &  &  &  \\ \hline
\multirow{2}{*}{Codeforces Round 1004 (Div. 1)} & \bf A  & \bf B  & \bf C  & \bf D1  & \bf D2  & \bf E  & \bf F  &  &  &  &  &  \multirow{2}{*}{2743.75} & \multirow{2}{*}{-} & \multirow{2}{*}{122/1030} & \multirow{2}{*}{2721} \\
& \color{BrickRed} 0.0 &  \color{ForestGreen} 743.75 &  \color{ForestGreen} 1250.00 &  \color{ForestGreen} 750.00 & \color{BrickRed} 0.0 & \color{BrickRed} 0.0 & \color{BrickRed} 0.0  &  &  &  &  &  \\ \hline
\multirow{2}{*}{Codeforces Round 1004 (Div. 2)} & \bf A  & \bf B  & \bf C  & \bf D  & \bf E  & \bf F  & \bf G  &  &  &  &  &  \multirow{2}{*}{5487.50} & \multirow{2}{*}{-} & \multirow{2}{*}{6/16749} & \multirow{2}{*}{2098} \\
&  \color{ForestGreen} 500.00 &  \color{ForestGreen} 993.75 & \color{BrickRed} 0.0 & \color{BrickRed} 0.0 &  \color{ForestGreen} 1743.75 &  \color{ForestGreen} 2250.00 & \color{BrickRed} 0.0  &  &  &  &  &  \\ \hline
\multirow{2}{*}{Codeforces Round 1005 (Div. 2)} & \bf A  & \bf B  & \bf C  & \bf D  & \bf E  & \bf F  &  &  &  &  &  &  \multirow{2}{*}{9230.71} & \multirow{2}{*}{-} & \multirow{2}{*}{1/17621} & \multirow{2}{*}{2260} \\
&  \color{ForestGreen} 500.00 &  \color{ForestGreen} 985.71 &  \color{ForestGreen} 1250.00 &  \color{ForestGreen} 1710.00 &  \color{ForestGreen} 2210.00 &  \color{ForestGreen} 2575.00  &  &  &  &  &  &  \\ \hline
\multirow{2}{*}{Educational Codeforces Round 174 (Rated for Div. 2)} & \bf A  & \bf B  & \bf C  & \bf D  & \bf E  & \bf F  &  &  &  &  &  &  \multirow{2}{*}{4} & \multirow{2}{*}{2.50} & \multirow{2}{*}{156/16701} & \multirow{2}{*}{2242} \\
& \color{ForestGreen} + & \color{ForestGreen} + & \color{ForestGreen} + & \color{ForestGreen} + & \color{BrickRed} - & \color{BrickRed} -  &  &  &  &  &  &  \\ \hline
\multirow{2}{*}{Educational Codeforces Round 175 (Rated for Div. 2)} & \bf A  & \bf B  & \bf C  & \bf D  & \bf E  & \bf F  &  &  &  &  &  &  \multirow{2}{*}{5} & \multirow{2}{*}{5.00} & \multirow{2}{*}{3/16060} & \multirow{2}{*}{2198} \\
& \color{ForestGreen} + & \color{ForestGreen} + & \color{ForestGreen} + & \color{ForestGreen} + & \color{ForestGreen} + & \color{BrickRed} -  &  &  &  &  &  &  \\ \hline
\multirow{2}{*}{Codeforces Round 1007 (Div. 2)} & \bf A  & \bf B  & \bf C  & \bf D1  & \bf D2  & \bf E  & \bf F  &  &  &  &  &  \multirow{2}{*}{8443.75} & \multirow{2}{*}{-} & \multirow{2}{*}{1/16254} & \multirow{2}{*}{2198} \\
&  \color{ForestGreen} 500.00 &  \color{ForestGreen} 1000.00 &  \color{ForestGreen} 1500.00 &  \color{ForestGreen} 1725.00 &  \color{ForestGreen} 1225.00 &  \color{ForestGreen} 2493.75 & \color{BrickRed} 0.0  &  &  &  &  &  \\ \hline
\multirow{2}{*}{Codeforces Round 1008 (Div. 1)} & \bf A  & \bf B  & \bf C  & \bf D  & \bf E  & \bf F  & \bf G  &  &  &  &  &  \multirow{2}{*}{2000.00} & \multirow{2}{*}{-} & \multirow{2}{*}{355/909} & \multirow{2}{*}{2312} \\
&  \color{ForestGreen} 500.00 & \color{BrickRed} 0.0 &  \color{ForestGreen} 1500.00 & \color{BrickRed} 0.0 & \color{BrickRed} 0.0 & \color{BrickRed} 0.0 & \color{BrickRed} 0.0  &  &  &  &  &  \\ \hline
\multirow{2}{*}{Codeforces Round 1008 (Div. 2)} & \bf A  & \bf B  & \bf C  & \bf D  & \bf E  & \bf F  & \bf G  &  &  &  &  &  \multirow{2}{*}{6975.00} & \multirow{2}{*}{-} & \multirow{2}{*}{5/14641} & \multirow{2}{*}{2008} \\
&  \color{ForestGreen} 500.00 &  \color{ForestGreen} 750.00 &  \color{ForestGreen} 1250.00 &  \color{ForestGreen} 1725.00 & \color{BrickRed} 0.0 &  \color{ForestGreen} 2750.00 & \color{BrickRed} 0.0  &  &  &  &  &  \\ \hline
\multirow{2}{*}{Educational Codeforces Round 176 (Rated for Div. 2)} & \bf A  & \bf B  & \bf C  & \bf D  & \bf E  & \bf F  &  &  &  &  &  &  \multirow{2}{*}{5} & \multirow{2}{*}{2.50} & \multirow{2}{*}{2/18159} & \multirow{2}{*}{2198} \\
& \color{ForestGreen} + & \color{ForestGreen} + & \color{ForestGreen} + & \color{ForestGreen} + & \color{ForestGreen} + & \color{BrickRed} -  &  &  &  &  &  &  \\ \hline
\multirow{2}{*}{Codeforces Round 1011 (Div. 2)} & \bf A  & \bf B  & \bf C  & \bf D  & \bf E  & \bf F1  & \bf F2  &  &  &  &  &  \multirow{2}{*}{10137.50} & \multirow{2}{*}{-} & \multirow{2}{*}{1/15906} & \multirow{2}{*}{2200} \\
&  \color{ForestGreen} 500.00 &  \color{ForestGreen} 1250.00 &  \color{ForestGreen} 1250.00 &  \color{ForestGreen} 1743.75 &  \color{ForestGreen} 2493.75 &  \color{ForestGreen} 2000.00 &  \color{ForestGreen} 900.00  &  &  &  &  &  \\ \hline
\multirow{2}{*}{Codeforces Round 1012 (Div. 1)} & \bf A  & \bf B1  & \bf B2  & \bf C1  & \bf C2  & \bf D  & \bf E  &  &  &  &  &  \multirow{2}{*}{2693.75} & \multirow{2}{*}{-} & \multirow{2}{*}{66/653} & \multirow{2}{*}{2745} \\
&  \color{ForestGreen} 725.00 &  \color{ForestGreen} 975.00 & \color{BrickRed} 0.0 &  \color{ForestGreen} 993.75 & \color{BrickRed} 0.0 & \color{BrickRed} 0.0 & \color{BrickRed} 0.0  &  &  &  &  &  \\ \hline
\multirow{2}{*}{Codeforces Round 1012 (Div. 2)} & \bf A  & \bf B  & \bf C  & \bf D  & \bf E1  & \bf E2  & \bf F1  & \bf F2  &  &  &  &  \multirow{2}{*}{9193.75} & \multirow{2}{*}{-} & \multirow{2}{*}{1/8536} & \multirow{2}{*}{2007} \\
&  \color{ForestGreen} 500.00 &  \color{ForestGreen} 1000.00 &  \color{ForestGreen} 1750.00 &  \color{ForestGreen} 1975.00 &  \color{ForestGreen} 1975.00 & \color{BrickRed} 0.0 &  \color{ForestGreen} 1993.75 & \color{BrickRed} 0.0  &  &  &  &  \\ \hline
\multirow{2}{*}{Codeforces Round 1014 (Div. 2)} & \bf A  & \bf B  & \bf C  & \bf D  & \bf E  & \bf F  &  &  &  &  &  &  \multirow{2}{*}{6500.00} & \multirow{2}{*}{-} & \multirow{2}{*}{2/15842} & \multirow{2}{*}{2213} \\
&  \color{ForestGreen} 500.00 &  \color{ForestGreen} 750.00 &  \color{ForestGreen} 1250.00 &  \color{ForestGreen} 1750.00 &  \color{ForestGreen} 2250.00 & \color{BrickRed} 0.0  &  &  &  &  &  &  \\ \hline
\multirow{2}{*}{Teza Round 1 (Codeforces Round 1015, Div. 1 + Div. 2)} & \bf A  & \bf B  & \bf C  & \bf D  & \bf E  & \bf F  & \bf G1  & \bf G2  & \bf H  &  &  &  \multirow{2}{*}{9723.21} & \multirow{2}{*}{-} & \multirow{2}{*}{55/11206} & \multirow{2}{*}{3008} \\
&  \color{ForestGreen} 750.00 &  \color{ForestGreen} 1000.00 &  \color{ForestGreen} 1500.00 &  \color{ForestGreen} 1743.75 &  \color{ForestGreen} 2243.75 & \color{BrickRed} 0.0 &  \color{ForestGreen} 2485.71 & \color{BrickRed} 0.0 & \color{BrickRed} 0.0  &  &  &  \\ \hline
\multirow{2}{*}{Neowise Labs Contest 1 (Codeforces Round 1018, Div. 1 + Div. 2)} & \bf A  & \bf B  & \bf C  & \bf D  & \bf E  & \bf F  & \bf G  & \bf H  &  &  &  &  \multirow{2}{*}{6397.50} & \multirow{2}{*}{-} & \multirow{2}{*}{70/12771} & \multirow{2}{*}{2933} \\
&  \color{ForestGreen} 500.00 &  \color{ForestGreen} 750.00 &  \color{ForestGreen} 1500.00 &  \color{ForestGreen} 1687.50 &  \color{ForestGreen} 1960.00 & \color{BrickRed} 0.0 & \color{BrickRed} 0.0 & \color{BrickRed} 0.0  &  &  &  &  \\ \hline
\multirow{2}{*}{Codeforces Round 1019 (Div. 2)} & \bf A  & \bf B  & \bf C  & \bf D  & \bf E  & \bf F  &  &  &  &  &  &  \multirow{2}{*}{7725.00} & \multirow{2}{*}{-} & \multirow{2}{*}{2/14465} & \multirow{2}{*}{2202} \\
&  \color{ForestGreen} 500.00 &  \color{ForestGreen} 1000.00 &  \color{ForestGreen} 1500.00 &  \color{ForestGreen} 1825.00 & \color{BrickRed} 0.0 &  \color{ForestGreen} 2900.00  &  &  &  &  &  &  \\ \hline
\multirow{2}{*}{Codeforces Round 1021 (Div. 1)} & \bf A  & \bf B  & \bf C  & \bf D  & \bf E  & \bf F  &  &  &  &  &  &  \multirow{2}{*}{4899.46} & \multirow{2}{*}{-} & \multirow{2}{*}{21/651} & \multirow{2}{*}{3143} \\
&  \color{ForestGreen} 493.75 &  \color{ForestGreen} 985.71 &  \color{ForestGreen} 1460.00 &  \color{ForestGreen} 1960.00 & \color{BrickRed} 0.0 & \color{BrickRed} 0.0  &  &  &  &  &  &  \\ \hline
\multirow{2}{*}{Codeforces Round 1021 (Div. 2)} & \bf A  & \bf B  & \bf C  & \bf D  & \bf E  & \bf F  &  &  &  &  &  &  \multirow{2}{*}{11399.46} & \multirow{2}{*}{-} & \multirow{2}{*}{1/5824} & \multirow{2}{*}{2019} \\
&  \color{ForestGreen} 500.00 &  \color{ForestGreen} 1250.00 &  \color{ForestGreen} 1493.75 &  \color{ForestGreen} 2235.71 &  \color{ForestGreen} 2710.00 &  \color{ForestGreen} 3210.00  &  &  &  &  &  &  \\ \hline
\multirow{2}{*}{Educational Codeforces Round 178 (Rated for Div. 2)} & \bf A  & \bf B  & \bf C  & \bf D  & \bf E  & \bf F  & \bf G  &  &  &  &  &  \multirow{2}{*}{6} & \multirow{2}{*}{12.11} & \multirow{2}{*}{4/11706} & \multirow{2}{*}{2215} \\
& \color{ForestGreen} + & \color{ForestGreen} + & \color{ForestGreen} + & \color{ForestGreen} + & \color{ForestGreen} + & \color{ForestGreen} + & \color{BrickRed} -  &  &  &  &  &  \\ \hline
\multirow{2}{*}{Codeforces Round 1022 (Div. 2)} & \bf A  & \bf B  & \bf C  & \bf D  & \bf E  & \bf F  &  &  &  &  &  &  \multirow{2}{*}{3235.71} & \multirow{2}{*}{-} & \multirow{2}{*}{300/11127} & \multirow{2}{*}{2137} \\
&  \color{ForestGreen} 500.00 &  \color{ForestGreen} 1250.00 &  \color{ForestGreen} 1485.71 & \color{BrickRed} 0.0 & \color{BrickRed} 0.0 & \color{BrickRed} 0.0  &  &  &  &  &  &  \\ \hline
\multirow{2}{*}{Codeforces Round 1023 (Div. 2)} & \bf A  & \bf B  & \bf C  & \bf D  & \bf E  & \bf F1  & \bf F2  &  &  &  &  &  \multirow{2}{*}{6537.50} & \multirow{2}{*}{-} & \multirow{2}{*}{6/11636} & \multirow{2}{*}{2209} \\
&  \color{ForestGreen} 250.00 &  \color{ForestGreen} 743.75 &  \color{ForestGreen} 1493.75 &  \color{ForestGreen} 1900.00 & \color{BrickRed} 0.0 &  \color{ForestGreen} 2150.00 & \color{BrickRed} 0.0  &  &  &  &  &  \\ \hline
\multirow{2}{*}{Codeforces Round 1024 (Div. 1)} & \bf A  & \bf B  & \bf C  & \bf D  & \bf E  & \bf F  &  &  &  &  &  &  \multirow{2}{*}{4075.00} & \multirow{2}{*}{-} & \multirow{2}{*}{156/857} & \multirow{2}{*}{2590} \\
&  \color{ForestGreen} 500.00 &  \color{ForestGreen} 1250.00 & \color{BrickRed} 0.0 &  \color{ForestGreen} 2325.00 & \color{BrickRed} 0.0 & \color{BrickRed} 0.0  &  &  &  &  &  &  \\ \hline
\multirow{2}{*}{Codeforces Round 1024 (Div. 2)} & \bf A  & \bf B  & \bf C  & \bf D  & \bf E  & \bf F  &  &  &  &  &  &  \multirow{2}{*}{6325.00} & \multirow{2}{*}{-} & \multirow{2}{*}{2/11201} & \multirow{2}{*}{1998} \\
&  \color{ForestGreen} 250.00 &  \color{ForestGreen} 500.00 &  \color{ForestGreen} 1000.00 &  \color{ForestGreen} 1750.00 & \color{BrickRed} 0.0 &  \color{ForestGreen} 2825.00  &  &  &  &  &  &  \\ \hline
\multirow{2}{*}{Codeforces Round 1025 (Div. 2)} & \bf A  & \bf B  & \bf C1  & \bf C2  & \bf C3  & \bf D  & \bf E  & \bf F  &  &  &  &  \multirow{2}{*}{8087.14} & \multirow{2}{*}{-} & \multirow{2}{*}{1/15945} & \multirow{2}{*}{2197} \\
&  \color{ForestGreen} 500.00 &  \color{ForestGreen} 985.71 &  \color{ForestGreen} 1235.71 &  \color{ForestGreen} 710.00 &  \color{ForestGreen} 460.00 &  \color{ForestGreen} 1710.00 &  \color{ForestGreen} 2485.71 & \color{BrickRed} 0.0  &  &  &  &  \\ \hline
\multirow{2}{*}{Codeforces Round 1026 (Div. 2)} & \bf A  & \bf B  & \bf C  & \bf D  & \bf E  & \bf F  &  &  &  &  &  &  \multirow{2}{*}{7931.25} & \multirow{2}{*}{-} & \multirow{2}{*}{8/17668} & \multirow{2}{*}{2198} \\
&  \color{ForestGreen} 500.00 &  \color{ForestGreen} 750.00 &  \color{ForestGreen} 1500.00 & \color{BrickRed} 0.0 &  \color{ForestGreen} 2243.75 &  \color{ForestGreen} 2937.50  &  &  &  &  &  &  \\ \hline
\multirow{2}{*}{Codeforces Round 1028 (Div. 1)} & \bf A  & \bf B  & \bf C  & \bf D  & \bf E  & \bf F1  & \bf F2  &  &  &  &  &  \multirow{2}{*}{3960.71} & \multirow{2}{*}{-} & \multirow{2}{*}{36/956} & \multirow{2}{*}{3093} \\
&  \color{ForestGreen} 500.00 &  \color{ForestGreen} 1225.00 & \color{BrickRed} 0.0 &  \color{ForestGreen} 2235.71 & \color{BrickRed} 0.0 & \color{BrickRed} 0.0 & \color{BrickRed} 0.0  &  &  &  &  &  \\ \hline
\multirow{2}{*}{Codeforces Round 1028 (Div. 2)} & \bf A  & \bf B  & \bf C  & \bf D  & \bf E  & \bf F  &  &  &  &  &  &  \multirow{2}{*}{7454.46} & \multirow{2}{*}{-} & \multirow{2}{*}{1/18314} & \multirow{2}{*}{2018} \\
&  \color{ForestGreen} 493.75 &  \color{ForestGreen} 750.00 &  \color{ForestGreen} 1250.00 &  \color{ForestGreen} 1975.00 & \color{BrickRed} 0.0 &  \color{ForestGreen} 2985.71  &  &  &  &  &  &  \\ \hline
\multirow{2}{*}{Educational Codeforces Round 179 (Rated for Div. 2)} & \bf A  & \bf B  & \bf C  & \bf D  & \bf E  & \bf F  & \bf G  &  &  &  &  &  \multirow{2}{*}{6} & \multirow{2}{*}{40.11} & \multirow{2}{*}{3/12301} & \multirow{2}{*}{2231} \\
& \color{ForestGreen} + & \color{ForestGreen} + & \color{ForestGreen} + & \color{ForestGreen} + & \color{ForestGreen} + & \color{ForestGreen} + & \color{BrickRed} -  &  &  &  &  &  \\ \hline
\multirow{2}{*}{Codeforces Round 1030 (Div. 2)} & \bf A  & \bf B  & \bf C  & \bf D1  & \bf D2  & \bf E  & \bf F  &  &  &  &  &  \multirow{2}{*}{7160.71} & \multirow{2}{*}{-} & \multirow{2}{*}{2/18335} & \multirow{2}{*}{2205} \\
&  \color{ForestGreen} 500.00 &  \color{ForestGreen} 993.75 &  \color{ForestGreen} 1000.00 &  \color{ForestGreen} 1243.75 &  \color{ForestGreen} 937.50 &  \color{ForestGreen} 2485.71 & \color{BrickRed} 0.0  &  &  &  &  &  \\ \hline
\multirow{2}{*}{Codeforces Round 1031 (Div. 2)} & \bf A  & \bf B  & \bf C  & \bf D  & \bf E  & \bf F  &  &  &  &  &  &  \multirow{2}{*}{4143.75} & \multirow{2}{*}{-} & \multirow{2}{*}{20/11032} & \multirow{2}{*}{2216} \\
&  \color{ForestGreen} 500.00 &  \color{ForestGreen} 743.75 & \color{BrickRed} 0.0 & \color{BrickRed} 0.0 & \color{BrickRed} 0.0 &  \color{ForestGreen} 2900.00  &  &  &  &  &  &  \\ \hline
\multirow{2}{*}{Codeforces Round 1033 (Div. 2) and CodeNite 2025} & \bf A  & \bf B  & \bf C  & \bf D  & \bf E  & \bf F  & \bf G  &  &  &  &  &  \multirow{2}{*}{9650.00} & \multirow{2}{*}{-} & \multirow{2}{*}{1/12948} & \multirow{2}{*}{2216} \\
&  \color{ForestGreen} 500.00 &  \color{ForestGreen} 750.00 &  \color{ForestGreen} 1250.00 &  \color{ForestGreen} 1750.00 &  \color{ForestGreen} 2500.00 &  \color{ForestGreen} 2900.00 & \color{BrickRed} 0.0  &  &  &  &  &  \\ \hline
\multirow{2}{*}{Educational Codeforces Round 180 (Rated for Div. 2)} & \bf A  & \bf B  & \bf C  & \bf D  & \bf E  & \bf F  &  &  &  &  &  &  \multirow{2}{*}{5} & \multirow{2}{*}{10.86} & \multirow{2}{*}{8/17128} & \multirow{2}{*}{2253} \\
& \color{ForestGreen} + & \color{ForestGreen} + & \color{ForestGreen} + & \color{ForestGreen} + & \color{ForestGreen} + & \color{BrickRed} -  &  &  &  &  &  &  \\ \hline
\bottomrule
\end{tabular}}
\label{tab:elo_rating_tir}
\end{table}

\clearpage


\begingroup
\definecolor{problemtitle}{HTML}{6A617F}
\definecolor{problemfill}{HTML}{E9E4EE}

\providecommand{\tightlist}{%
  \setlength{\itemsep}{0.35em}%
  \setlength{\parskip}{0pt}%
}

\newtcolorbox{problemstatementbox}[1]{%
  breakable,
  colback=problemfill,
  colframe=black!85,
  colbacktitle=problemtitle,
  coltitle=white,
  fonttitle=\bfseries,
  title=#1,
  boxrule=0.8pt,
  arc=6pt,
  left=10pt,
  right=10pt,
  top=10pt,
  bottom=8pt,
  toptitle=6pt,
  bottomtitle=6pt,
  before skip=0.5em,
  after skip=1.1em,
}

\newenvironment{compactproof}{%
  \begingroup
  \small
  \setlength{\parskip}{0.22em}
  \setlength{\abovedisplayskip}{0.45em}
  \setlength{\belowdisplayskip}{0.45em}
  \setlength{\abovedisplayshortskip}{0.3em}
  \setlength{\belowdisplayshortskip}{0.3em}
  \setlist[itemize]{leftmargin=1.5em, topsep=0.2em, itemsep=0.1em, parsep=0pt}
  \setlist[enumerate]{leftmargin=1.7em, topsep=0.2em, itemsep=0.1em, parsep=0pt}
  \linespread{0.97}\selectfont
}{%
  \par
  \endgroup
}

\def\imoappendixmode{1}

\definecolor{problemtitle}{HTML}{6A617F}
\definecolor{problemfill}{HTML}{E9E4EE}
\definecolor{commentfill}{HTML}{F7F4FA}
\definecolor{commentrule}{HTML}{8A7E9D}

\providecommand{\tightlist}{%
  \setlength{\itemsep}{0.35em}%
  \setlength{\parskip}{0pt}%
}
\providecommand{\texorpdfstring}[2]{#1}
\providecommand{\phantomsection}{}
\setlist[itemize]{leftmargin=1.8em, topsep=0.4em}
\setlist[enumerate]{leftmargin=2em, topsep=0.4em}
\allowdisplaybreaks
\setcounter{secnumdepth}{1}
\setcounter{tocdepth}{1}
\setlength{\headheight}{14pt}
\clubpenalty=10000
\widowpenalty=10000
\emergencystretch=2em

\renewtcolorbox{problemstatementbox}[1]{%
  breakable,
  enhanced,
  colback=problemfill,
  colframe=black!85,
  colbacktitle=problemtitle,
  coltitle=white,
  fonttitle=\bfseries,
  title=#1,
  boxrule=0.8pt,
  arc=6pt,
  left=10pt,
  right=10pt,
  top=10pt,
  bottom=8pt,
  toptitle=6pt,
  bottomtitle=6pt,
  before skip=0.5em,
  after skip=1.1em,
}

\renewenvironment{compactproof}{%
  \begingroup
  \small
  \setlength{\parskip}{0.22em}
  \setlength{\abovedisplayskip}{0.45em}
  \setlength{\belowdisplayskip}{0.45em}
  \setlength{\abovedisplayshortskip}{0.3em}
  \setlength{\belowdisplayshortskip}{0.3em}
  \setlist[itemize]{leftmargin=1.5em, topsep=0.2em, itemsep=0.1em, parsep=0pt}
  \setlist[enumerate]{leftmargin=1.7em, topsep=0.2em, itemsep=0.1em, parsep=0pt}
  \linespread{0.97}\selectfont
}{%
  \par
  \endgroup
}

\newtcolorbox{expertcommentbox}{%
  breakable,
  enhanced,
  colback=commentfill,
  colframe=commentrule!75!black,
  boxrule=0.45pt,
  borderline west={2.4pt}{0pt}{commentrule},
  arc=4pt,
  left=10pt,
  right=10pt,
  top=7pt,
  bottom=7pt,
  before skip=0.8em,
  after skip=1.15em,
}

\newenvironment{humanexpertcomment}{%
  \Needspace{6\baselineskip}
  \begin{expertcommentbox}
  {\color{problemtitle}\bfseries\small \textsc{Human Expert Comment}}\par
  \vspace{0.35em}
  \begingroup
  \small
  \setlength{\parskip}{0.35em}
  \setlist[itemize]{leftmargin=1.4em, topsep=0.2em, itemsep=0.1em, parsep=0pt}
  \setlist[enumerate]{leftmargin=1.6em, topsep=0.2em, itemsep=0.1em, parsep=0pt}
}{%
  \par
  \endgroup
  \end{expertcommentbox}
}

\definecolor{annotfill}{HTML}{FFF8E7}
\definecolor{annotrule}{HTML}{C4960C}
\definecolor{annotlabel}{HTML}{8B6914}
\definecolor{annothighlight}{HTML}{FFEEA3}

\newtcolorbox{annotstyle}{
  enhanced jigsaw, frame hidden,
  colback=annotfill,
  borderline west={1.5pt}{0pt}{annotrule},
  boxrule=0pt, arc=0pt,
  left=4pt, right=3pt, top=2pt, bottom=2pt,
  width=\marginparwidth,
  nobeforeafter,
}

\newcommand{\annotbox}[1]{%
  \marginpar{%
    \begin{annotstyle}
      \raggedright\scriptsize\setlength{\parskip}{2pt}%
      #1%
    \end{annotstyle}%
  }%
}

\newcommand{\annot}[3]{%
  \colorbox{annothighlight}{#2}%
  {\textsuperscript{\scriptsize\color{annotlabel}\textsf{#1}}}%
  \annotbox{#3}%
}

\newcommand{\annotmark}[2]{%
  {\textsuperscript{\scriptsize\color{annotlabel}\textsf{#1}}}%
  \annotbox{#2}%
}

\newcommand{\annotnote}[3]{%
  {\color{annotlabel}\bfseries\sffamily #1}~{\color{annotlabel}\textbar}~{\itshape #2}\par\smallskip #3%
}

\newcommand{\proofremark}[3]{}
\newcommand{\aref}[1]{}

\newtcolorbox{llmcomment}{
  breakable,
  enhanced,
  frame hidden,
  colback=commentfill,
  boxrule=0pt,
  borderline west={2.4pt}{0pt}{commentrule},
  arc=4pt,
  left=10pt,
  right=10pt,
  top=7pt,
  bottom=7pt,
  before skip=0.8em,
  after skip=1.15em,
  title={\textsc{LLM Judge Comment}},
  fonttitle=\bfseries\small,
  coltitle=problemtitle,
  colbacktitle=commentfill,
  coltext=black,
  fontupper=\small
}

\pagestyle{fancy}
\fancyhf{}
\fancyhead[L]{\textsc{IMO2025}}
\fancyhead[R]{\textsc{IMO 2025 Model Proofs}}
\fancyfoot[R]{\thepage\ of \pageref{LastPage}}




\newgeometry{left=2.0cm,right=6.5cm,top=3cm,bottom=3cm,marginparwidth=5.3cm,marginparsep=0.5cm} 
\section{IMO 2025 Model Solutions}
\label{appendix:imo2025-model-solutions}
\markboth{Problem 1}{Problem 1}
\begin{problemstatementbox}{Problem 1}
        A line in the plane is called \emph{sunny} if it is not parallel to any
of the \(x\)-axis, the \(y\)-axis, or the line \(x + y = 0\).

Let \(n \geq 3\) be a given integer. Determine all nonnegative integers
\(k\) such that there exist \(n\) distinct lines in the plane satisfying
both of the following:

\begin{itemize}
    \item for all positive integers $a$ and $b$ with $a + b \leq n + 1$, the point $(a, b)$ lies on at least one of the lines; and
    \item exactly $k$ of the $n$ lines are sunny.
\end{itemize}
        \tcblower
        \textbf{Short Answer}\par
        k = 0, 1, 3 for all n
        \end{problemstatementbox}

        \Needspace{10\baselineskip}
        \textbf{Model Proof}\\
        \begin{compactproof}
        \textbf{Notation.} For an integer \(n\ge 3\) define\\
\[
S_n=\{(a,b)\in\mathbb Z^2\mid a\ge 1,\;b\ge 1,\;a+b\le n+1\}.
\]\\
A line in the plane is called \emph{sunny} if its slope is neither \(0\)
(horizontal), \(\infty\) (vertical), nor \(-1\) (the slope of the line
\(x+y=0\)).

We have to determine all non-negative integers \(k\) for which there
exist \(n\) distinct lines covering all points of \(S_n\) with exactly
\(k\) of them sunny.

\begin{center}\rule{0.5\linewidth}{0.5pt}\end{center}

\textbf{1. Constructions showing that \(k=0,1,3\)
are
attainable}

\textbf{\(k=0\).}\\
Take the \(n\) vertical lines \(x=1,x=2,\dots,x=n\). Every point
\((a,b)\in S_n\) has \(a\le n\), hence lies on the line \(x=a\). These
lines are not sunny. Clearly there are \(n\) distinct lines, so \(k=0\)
is possible for every \(n\ge 3\).

\textbf{\(k=1\).}\\
For \(n\ge 3\) take the \(n-1\) vertical lines \(x=1,\dots,x=n-1\) and
the line\\
\[
L:\;\text{through }(n,1)\text{ and }(n-2,2).
\]\\
The slope of \(L\) is \(\frac{2-1}{(n-2)-n}=-\frac12\); it is not
\(0,\infty,-1\), so \(L\) is sunny.\\
- Points with \(a\le n-1\) are covered by the vertical line \(x=a\).\\
- The only point with \(a=n\) is \((n,1)\), which lies on \(L\).

Thus all points of \(S_n\) are covered. Hence \(k=1\) is attainable for
all \(n\ge 3\).

\textbf{\(k=3\).}

\emph{Case \(n=3\).}\\
\(S_3\) consists of the six points\\
\[
(1,1),(1,2),(1,3),(2,1),(2,2),(3,1).
\]\\
Take the three sunny lines\\
\[
\begin{aligned}
\ell_1&:\; y=x, \\
\ell_2&:\; y=-2x+5, \\
\ell_3&:\; y=-\tfrac12x+\tfrac52.
\end{aligned}
\]\\
Their slopes are \(1,-2,-\tfrac12\) - none is \(0,\infty,-1\), so they
are sunny. One checks directly that each of the six points belongs to at
least one of them. Therefore \(k=3\) is possible for \(n=3\).

\emph{Case \(n\ge 4\).}\\
Take the \(n-3\) vertical lines \(x=1,x=2,\dots,x=n-3\).\\
The remaining points of \(S_n\) are those with \(a\ge n-2\):\\
\[
W=\{(n-2,1),(n-2,2),(n-2,3),(n-1,1),(n-1,2),(n,1)\}.
\]\\
Cover \(W\) by the following three sunny lines:\\
\[
\begin{aligned}
L_a&:\;\text{through }(n-2,1)\text{ and }(n-1,2) &&\text{(slope }1),\\
L_b&:\;\text{through }(n-2,2)\text{ and }(n,1) &&\text{(slope }-\tfrac12),\\
L_c&:\;\text{through }(n-2,3)\text{ and }(n-1,1) &&\text{(slope }-2).
\end{aligned}
\]\\
All slopes are different from \(0,\infty,-1\), hence these lines are
sunny.\\
Together with the \(n-3\) vertical lines we have exactly \(n\) distinct
lines. Every point with \(a\le n-3\) lies on a vertical line; the six
points of \(W\) lie on \(L_a,L_b,L_c\). Thus the whole \(S_n\) is
covered. Consequently \(k=3\) is attainable for all \(n\ge 3\).

\begin{center}\rule{0.5\linewidth}{0.5pt}\end{center}

\textbf{2. Impossibility: no other value of \(k\)
can
occur}

We prove by induction on \(n\) that any covering of \(S_n\) by \(n\)
lines must have \(k\in\{0,1,3\}\).

\paragraph*{2.1. The boundary set \(X\) and a
lemma}

Define the three \emph{special} lines\\
\[
S_1:\; x=1,\qquad S_2:\; y=1,\qquad S_3:\; x+y=n+1.
\]\\
These are the sides of the right triangle that contains \(S_n\). Let
\(X\) be the set of points of \(S_n\) lying on these three sides; i.e.\\
\[\textcolor{annotlabel}{
X = \{(1,b)\mid 1\le b\le n\}\cup\{(a,1)\mid 1\le a\le n\}\cup\{(a,b)\mid a+b=n+1\}}.
\]\annotmark{A1}{\annotnote{A1}{}{Similar to definition of $S_n$, we need to require $a,b$ are integers.}}\\
There are \textcolor{annotlabel}{\(|X|=3n-3\) distinct points (the three corners are counted twice)}\annotmark{A2}{\annotnote{A2}{}{What means of counted twice? ~Two options: 1.~Each edge have $n$ point, so totally $3n$ in 3 edges, but three corners counted twice so $3n{-}3$ distinct. 2.~Directly say $3n{-}3$ distinct points without mentioning counted twice is also okay.}}

\textbf{Lemma.} A line that is \textbf{not} one of \(S_1,S_2,S_3\)
contains at most two points of \(X\).

\emph{Proof.} \textcolor{annotlabel}{If such a line contained two distinct points from the same
side, it would coincide with that side, contradicting the assumption.
Hence it can contain at most one point from each side.}\annotmark{A3}{\annotnote{A3}{}{This sentence is not related to the proof.}} Moreover, a
straight line can intersect the boundary of a triangle in at most two
points (unless it coincides with a side). Therefore it contains at most
two points of \(X\). \(\square\)

\paragraph*{2.2. A counting argument for
\(n\ge 4\)}

Consider any covering of \(S_n\) by \(n\) lines, and \textcolor{annotlabel}{let \(k\) be the number of sunny lines}\annotmark{A4}{\annotnote{A4}{}{The definition is unnecessary since $k$ is not used in 2.2.}}. For each line \(\ell\) denote by \(f(\ell)\) the
number of points of \(X\) that lie on \(\ell\). Because every point of
\(X\) is covered at least once,\\
\[
\sum_{\ell} f(\ell) \ge |X| = 3n-3. \tag{1}
\]\\
If a line is one of the special sides \(S_1,S_2,S_3\), then it contains
all \(n\) points of that side, so \(f(\ell)=n\). If it is not a special
side (whether sunny or not), the lemma gives \(f(\ell)\le 2\).

Let \(s\) be the number of the three special sides that actually appear
among the \(n\) lines. Then\\
\[
\sum_{\ell} f(\ell) \le s\cdot n + (n-s)\cdot 2 = 2n + s(n-2). \tag{2}
\]\\
Combining (1) and (2) we obtain\\
\[
3n-3 \le 2n + s(n-2) \quad\Longrightarrow\quad n-3 \le s(n-2). \tag{3}
\]\\
For \(n\ge 4\) we have \textcolor{annotlabel}{$n-2>0$}\annotmark{A5}{\annotnote{A5}{}{$n-3>0$}}, so (3) forces \(s\ge 1\). Hence
\textbf{every covering of \(S_n\) with \(n\ge 4\) lines must contain at
least one of the sides \(x=1\), \(y=1\) or \(x+y=n+1\)}.

\paragraph*{2.3. Reducing the problem to
\(n-1\)}

Assume \(n\ge 4\) and that a covering of \(S_n\) with \(n\) lines and
\(k\) sunny lines exists. By the previous paragraph, one of the special
sides is present; we show how to produce a covering of \(S_{n-1}\) with
\(n-1\) lines and the same \(k\).

\emph{Case A: \(x=1\) is present.}\\
Remove the line \(x=1\). The remaining \(n-1\) lines cover all points of
\(S_n\) with \(a\ge 2\). Let\\
\[
S' = \{(a,b)\in S_n \mid a\ge 2\},
\]\\
and define the translation \(\varphi:S'\to S_{n-1}\) by
\textcolor{annotlabel}{$\varphi(a,b)=(a-1,b)$}\annotmark{A6}{\annotnote{A6}{}{$\phi((a,b))$ since input is a pair $(a,b)$ not two parameter $a$ and $b$.}}. This is a bijection. Translate the remaining
\(n-1\) lines by \((-1,0)\): for each line \(\ell\) (with
$\ell\neq (x=1)$), set $\ell' = \{(x-1,y)\mid (x,y)\in\ell\}$.
Translation preserves slopes, so \(\ell'\) is sunny exactly when
\(\ell\) is sunny, and distinct lines remain distinct. Because the
original lines covered \(S'\), their translates cover
\(\varphi(S')=S_{n-1}\). Thus we obtain a covering of \(S_{n-1}\) by
\(n-1\) lines with exactly \(k\) sunny lines.

\emph{Case B: \(y=1\) is present.}\\
Analogously, remove \(y=1\), translate the remaining lines by
\((0,-1)\), and obtain a covering of \(S_{n-1}\) with \(n-1\) lines and
\(k\) sunny lines.

\emph{Case C: \(x+y=n+1\) is present.}\\
Remove this line. The set \(\{(a,b)\in S_n\mid a+b\le n\}\) is exactly
\(S_{n-1}\). The remaining \(n-1\) lines cover \(S_{n-1}\) directly
(they cover all of \(S_n\), and the removed line contains no point of
\(S_{n-1}\)). No translation is needed, and the number of sunny lines
stays \(k\).

In all three cases we have produced a covering of \(S_{n-1}\) by \(n-1\)
lines with the same \(k\).

\paragraph*{2.4. Base case \(n=3\) and
induction}

\textbf{Base case \(n=3\).}\\
We have already exhibited coverings with \(k=0,1,3\). It remains to show
that \(k=2\) is impossible.

The set \(S_3\) is exactly the boundary \(X\) for \(n=3\). Observe that
the only lines containing two points of \(S_3\) and being sunny are\\
\[
\begin{aligned}
\ell_1&:\text{ through }(1,1),(2,2)\;(slope 1),\\
\ell_2&:\text{ through }(1,2),(3,1)\;(slope -1/2),\\
\ell_3&:\text{ through }(1,3),(2,1)\;(slope -2).
\end{aligned}
\]
Any other line containing two points has slope \(0\), \(\infty\), or
\(-1\), hence is non-sunny.

Suppose a covering of \(S_3\) by three lines has exactly \(k=2\) sunny
lines. Then there is exactly one non-sunny line \(N\).

\emph{If \(N\) is one of the special sides} (say \(x=1\)), then \(N\)
covers three points. The remaining three points must be covered by the
two sunny lines. But any line through two of those three points is
either vertical, horizontal, or of slope \(-1\), all non-sunny. Hence
each sunny line can contain at most one of these points, so they cover
at most two points - contradiction.

\emph{If \(N\) is not a special side}, then \(N\) contains at most two
points. Let \(f(\ell)\) be the number of points of \(S_3\) on line
\(\ell\). To cover all six points we need \(\sum f(\ell)\ge 6\). For a
sunny line that is not special, \(f(\ell)\le 2\) (by the lemma). If any
line had \(f(\ell)\le 1\), then the total sum would be at most
\(1+2+2=5<6\). Therefore each line must satisfy \(f(\ell)=2\). In
particular, the two sunny lines each contain exactly two points, and
\(N\) also contains exactly two points. Thus the three lines partition
\(S_3\) into three disjoint collinear pairs.

The only collinear pairs that can be covered by a sunny line are
\(\{\ell_1\},\{\ell_2\},\{\ell_3\}\). Hence the two sunny lines must be
two of \(\{\ell_1,\ell_2,\ell_3\}\). The uncovered pair is exactly the
pair of the remaining \(\ell_m\). For \(N\) to cover that pair, \(N\)
would have to be the line through those two points, which is \(\ell_m\),
a sunny line - contradicting that \(N\) is non-sunny. Therefore \(k=2\)
is impossible. Consequently, for \(n=3\) the only possible \(k\) are
\(0,1,3\).

\textbf{Induction step for \(n\ge 4\).}\\
Assume the statement holds for \(n-1\). Given any covering of \(S_n\) by
\(n\) lines with \(k\) sunny lines, we have shown how to produce a
covering of \(S_{n-1}\) by \(n-1\) lines with the same \(k\). By the
induction hypothesis, \(k\in\{0,1,3\}\). Hence the statement holds for
\(n\) as well.

By induction, for every \(n\ge 3\) any covering of \(S_n\) by \(n\)
lines must have \(k\in\{0,1,3\}\). Together with the constructions in
\S{}1, we conclude that the set of all non-negative integers \(k\) for
which such a configuration exists is exactly

\[
\boxed{\{0,\;1,\;3\}}.
\]

\begin{center}\rule{0.5\linewidth}{0.5pt}\end{center}
\end{compactproof}

\begin{humanexpertcomment}
\[\boxed{7/7}\]

\end{humanexpertcomment}

\clearpage
\phantomsection

\restoregeometry
\markboth{Problem 2}{Problem 2}
\begin{problemstatementbox}{Problem 2}
        Let \(\Omega\) and \(\Gamma\) be circles with centres \(M\) and \(N\),
respectively, such that the radius of \(\Omega\) is less than the radius
of \(\Gamma\). Suppose \(\Omega\) and \(\Gamma\) intersect at two
distinct points \(A\) and \(B\). Line \(MN\) intersects \(\Omega\) at
\(C\) and \(\Gamma\) at \(D\), so that \(C, M, N, D\) lie on \(MN\) in
that order. Let \(P\) be the circumcentre of triangle \(ACD\). Line
\(AP\) meets \(\Omega\) again at \(E \ne A\) and meets \(\Gamma\) again
at \(F \ne A\). Let \(H\) be the orthocentre of triangle \(PMN\).

Prove that the line through \(H\) parallel to \(AP\) is tangent to the
circumcircle of triangle \(BEF\).
        \tcblower
        \textbf{Short Answer}\par
        \emph{No short answer provided.}
        \end{problemstatementbox}

        \Needspace{10\baselineskip}
        \subsection*{Model Proof}
        \begin{compactproof}
        We set up a coordinate system with \(M=(0,0)\) and \(N=(d,0)\),
\(d>0\).\\
Circles: \(\Omega: x^2+y^2=r^2\) (radius \(r\)),
\(\Gamma: (x-d)^2+y^2=R^2\) (radius \(R\)), \(0<r<R\).\\
They intersect in two points; let \(A=(x_A,y_A)\) with \(y_A>0\),
\(B=(x_A,-y_A)\). Solving the circle equations gives

\[
x_A=\frac{d^2+r^2-R^2}{2d},\qquad y_A^2=\frac{\big((R+r)^2-d^2\big)\big(d^2-(R-r)^2\big)}{4d^2}.
\]

Define

\[
U=R+r,\quad V=R-r,\quad K=d^2-V^2,\quad L=U^2-d^2.
\]

Then \(4y_A^2 = \dfrac{LK}{d^2}\). (Note that \(d\) satisfies \(V<d<U\)
because the circles intersect in two distinct points.)

The line \(MN\) (the \(x\)-axis) meets \(\Omega\) at \(C=(-r,0)\) and
\(\Gamma\) at \(D=(d+R,0)\). The order on \(MN\) is \(C,M,N,D\).

\begin{center}\rule{0.5\linewidth}{0.5pt}\end{center}

\subsubsection*{1. Circumcenter \(P\) of
\(\triangle ACD\)}

Since \(C\) and \(D\) lie on the \(x\)-axis, the perpendicular bisector
of \(CD\) is the vertical line \(x = h\) where

\[
h=\frac{-r+(d+R)}{2}=\frac{d+R-r}{2}=\frac{d+V}{2}.
\]

Thus \(P=(h,p)\) for some \(p\). From \(PA=PC\) we obtain

\[
(h-x_A)^2+(p-y_A)^2=(h+r)^2+p^2.
\]

Solving for \(p\) yields

\[
p = -\frac{K\,T}{4d\,y_A},\qquad\text{where } T = d+R+r = d+U.
\]

The vector \(\mathbf{v}=P-A\) is

\[
\mathbf{v} = (v_x,v_y)=\left(\frac{VT}{2d},\; -\frac{KUT}{4d^2y_A}\right).
\]

\begin{center}\rule{0.5\linewidth}{0.5pt}\end{center}

\subsubsection*{2. Second intersections \(E,F\) and
midpoint \(S\) of
\(EF\)}

The line \(AP\) consists of points \(A+t\mathbf{v}\). Substituting into
\(\Omega\) (using \(A\cdot A=r^2\)) gives

\[
2t\,A\cdot\mathbf{v}+t^2|\mathbf{v}|^2=0,
\]

so the non-zero root is \(t_E = -2A\cdot\mathbf{v}/|\mathbf{v}|^2\).
Compute

\[
A\cdot\mathbf{v} = -\frac{rT}{2}\quad\Longrightarrow\quad t_E = \frac{rT}{|\mathbf{v}|^2}.
\]

Similarly, for \(\Gamma\) we use \((A-N)\cdot\mathbf{v} = -RT/2\) and
obtain

\[
t_F = \frac{RT}{|\mathbf{v}|^2}.
\]

Hence \(E=A+t_E\mathbf{v},\; F=A+t_F\mathbf{v}\). The midpoint \(S\) of
\(EF\) is \(S=A+t_S\mathbf{v}\) with

\[
t_S = \frac{t_E+t_F}{2} = \frac{(r+R)T}{2|\mathbf{v}|^2}= \frac{UT}{2|\mathbf{v}|^2}.
\]

\begin{center}\rule{0.5\linewidth}{0.5pt}\end{center}

\subsubsection*{3. Unit vectors along and perpendicular
to
\(AP\)}

Let \(\mathbf{e}=\mathbf{v}/|\mathbf{v}|\) and
\(\mathbf{n}=(-v_y,v_x)/|\mathbf{v}|\). Then \(\mathbf{e}\) is the
direction of \(AP\), \(\mathbf{n}\) is perpendicular to \(AP\), and
\(EF\) is parallel to \(\mathbf{e}\). The half-length of \(EF\) is

\[
\ell = \frac{|t_F-t_E|\,|\mathbf{v}|}{2} = \frac{VT}{2|\mathbf{v}|},\qquad \ell^2 = \frac{V^2T^2}{4|\mathbf{v}|^2}.
\]

Thus \(E = S-\ell\mathbf{e},\; F = S+\ell\mathbf{e}\).

\begin{center}\rule{0.5\linewidth}{0.5pt}\end{center}

\subsubsection*{4. Orthocenter \(H\) of
\(\triangle PMN\)}

Points: \(M=(0,0),\; N=(d,0),\; P=(h,p)\).\\
The altitude from \(P\) is the vertical line \(x=h\).\\
The altitude from \(M\) is perpendicular to \(PN\); slope of \(PN\) is
\(-p/(d-h)\), so the altitude from \(M\) has equation
\(y = \frac{d-h}{p}x\). Intersection gives

\[
H = \left(h,\; \frac{h(d-h)}{p}\right).
\]

Now \(h(d-h) = \frac{(d+V)(d-V)}{4} = \frac{d^2-V^2}{4} = \frac{K}{4}\).
Using \(p = -\frac{KT}{4dy_A}\), we obtain

\[
H = \left(h,\; \frac{K/4}{-KT/(4dy_A)}\right) = \left(h,\; -\frac{dy_A}{T}\right).
\]

\begin{center}\rule{0.5\linewidth}{0.5pt}\end{center}

\subsubsection*{5. Computation of
\(\beta = (S-B)\cdot\mathbf{n}\) and
\(\gamma = (S-H)\cdot\mathbf{n}\)}

Because \(S = A + t_S\mathbf{v}\) and \(\mathbf{n}\perp\mathbf{v}\), we
have

\[
(S-B)\cdot\mathbf{n} = (A-B)\cdot\mathbf{n},\qquad (S-H)\cdot\mathbf{n} = (A-H)\cdot\mathbf{n}.
\]

Thus

\[
\beta = (A-B)\cdot\mathbf{n},\qquad \gamma = (A-H)\cdot\mathbf{n}.
\]

Now \(A-B = (0,2y_A)\) and \(\mathbf{n}=(-v_y,v_x)/|\mathbf{v}|\), so

\[
\beta = \frac{2y_A v_x}{|\mathbf{v}|} = \frac{2y_A}{|\mathbf{v}|}\cdot\frac{VT}{2d} = \frac{y_A V T}{d\,|\mathbf{v}|}.
\]

To compute \(\gamma\), we use
\(\mathbf{w}\cdot\mathbf{n} = -\frac{\mathbf{w}\times\mathbf{v}}{|\mathbf{v}|}\)
(for any \(\mathbf{w}\)). Hence

\[
\gamma = -\frac{(A-H)\times\mathbf{v}}{|\mathbf{v}|}.
\]

We compute the cross product. Coordinates:

\[
x_A-h = -\frac{VT}{2d},\qquad y_A - H_y = y_A + \frac{dy_A}{T} = y_A\frac{T+d}{T},
\] \[
v_x = \frac{VT}{2d},\qquad v_y = -\frac{KUT}{4d^2y_A}.
\]

Then

\[
(A-H)\times\mathbf{v} = (x_A-h)v_y - (y_A-H_y)v_x
= \left(-\frac{VT}{2d}\right)\!\left(-\frac{KUT}{4d^2y_A}\right) - \left(y_A\frac{T+d}{T}\right)\!\left(\frac{VT}{2d}\right)
= \frac{VKUT^2}{8d^3y_A} - \frac{Vy_A(T+d)}{2d}.
\]

Factor \(\frac{V}{2d}\):

\[
(A-H)\times\mathbf{v} = \frac{V}{2d}\left( \frac{KUT^2}{4d^2y_A} - y_A(T+d) \right).
\]

Let \(B = \frac{KUT^2}{4d^2y_A} - y_A(T+d)\). Multiply by \(2y_A\):

\[
2y_AB = \frac{KUT^2}{2d^2} - 2y_A^2(T+d).
\]

Using \(4y_A^2 = \frac{LK}{d^2}\) we have \(y_A^2 = \frac{LK}{4d^2}\),
so

\[
2y_A^2(T+d) = \frac{LK(T+d)}{2d^2}.
\]

Thus

\[
2y_AB = \frac{K}{2d^2}\bigl( UT^2 - L(T+d) \bigr).
\]

Now \(T = d+U\) and \(L = U^2-d^2\). Compute

\[
UT^2 - L(T+d) = U(d+U)^2 - (U^2-d^2)(2d+U) = 2d^2T.
\]

Hence \(2y_AB = \frac{K}{2d^2}\cdot 2d^2T = KT\), so
\(B = \frac{KT}{2y_A}\). Consequently,

\[
(A-H)\times\mathbf{v} = \frac{V}{2d}\cdot\frac{KT}{2y_A} = \frac{VKT}{4dy_A}.
\]

Therefore

\[
\gamma = -\frac{VKT}{4dy_A|\mathbf{v}|}.
\]

Comparing with \(\beta = \dfrac{y_AVT}{d|\mathbf{v}|}\), we obtain

\[
\gamma = -\frac{K}{4y_A^2}\,\beta.
\]

Moreover, using \(4y_A^2 = \dfrac{LK}{d^2}\) we can write

\[
\frac{K}{4y_A^2} = \frac{d^2}{L}\quad\Longrightarrow\quad \gamma = -\frac{d^2}{L}\,\beta.
\]

\begin{center}\rule{0.5\linewidth}{0.5pt}\end{center}

\subsubsection*{6. Expression for
\(|S-B|^2\)}

Write \(A-B = \alpha\mathbf{e} + \beta\mathbf{n}\), where
\(\alpha = (A-B)\cdot\mathbf{e} = \dfrac{(A-B)\cdot\mathbf{v}}{|\mathbf{v}|}\).
Compute

\[
(A-B)\cdot\mathbf{v} = 2y_A v_y = 2y_A\!\left(-\frac{KUT}{4d^2y_A}\right) = -\frac{KUT}{2d^2},
\] so

\[
\alpha = -\frac{KUT}{2d^2|\mathbf{v}|}.
\]

Since \(S = A + t_S\mathbf{v}\) and
\(\mathbf{v}=|\mathbf{v}|\mathbf{e}\),

\[
S-B = (A-B) + t_S\mathbf{v} = (\alpha\mathbf{e}+\beta\mathbf{n}) + t_S|\mathbf{v}|\mathbf{e} = (\alpha + t_S|\mathbf{v}|)\mathbf{e} + \beta\mathbf{n}.
\]

Now \(t_S|\mathbf{v}| = \dfrac{UT}{2|\mathbf{v}|}\). Thus

\[
\alpha + t_S|\mathbf{v}| = -\frac{KUT}{2d^2|\mathbf{v}|} + \frac{UT}{2|\mathbf{v}|} = \frac{UT}{2|\mathbf{v}|}\left(1-\frac{K}{d^2}\right) = \frac{UT}{2|\mathbf{v}|}\cdot\frac{V^2}{d^2} = \frac{UV^2T}{2d^2|\mathbf{v}|}.
\]

Hence

\[
|S-B|^2 = (\alpha + t_S|\mathbf{v}|)^2 + \beta^2 = \frac{U^2V^4T^2}{4d^4|\mathbf{v}|^2} + \frac{y_A^2V^2T^2}{d^2|\mathbf{v}|^2}.
\]

Substitute \(y_A^2 = \dfrac{LK}{4d^2}\) to obtain

\[
|S-B|^2 = \frac{V^2T^2}{|\mathbf{v}|^2}\left(\frac{U^2V^2}{4d^4} + \frac{LK}{4d^4}\right) = \frac{V^2T^2}{4d^4|\mathbf{v}|^2}\bigl(U^2V^2 + LK\bigr).
\]

\begin{center}\rule{0.5\linewidth}{0.5pt}\end{center}

\subsubsection*{7. Circumcenter \(O\) of
\(\triangle BEF\) and the tangency
condition}

The perpendicular bisector of \(EF\) is the line through \(S\) parallel
to \(\mathbf{n}\), so we can write \(O = S + \lambda\mathbf{n}\) for
some real \(\lambda\). The circumradius \(\rho\) satisfies
\(\rho^2 = |O-E|^2 = \ell^2 + \lambda^2\). The condition
\(|O-B|^2 = \rho^2\) gives

\[
|(S-B)+\lambda\mathbf{n}|^2 = \ell^2 + \lambda^2.
\]

Expanding, and using \((S-B)=\alpha' \mathbf{e} + \beta\mathbf{n}\) with
\(\alpha' = \alpha + t_S|\mathbf{v}|\) (so that
\(|S-B|^2 = \alpha'^2 + \beta^2\)), we get

\[
\alpha'^2 + (\beta+\lambda)^2 = \ell^2 + \lambda^2
\;\Longrightarrow\;
|S-B|^2 + 2\lambda\beta = \ell^2.
\tag{1}
\]

Thus \(2\lambda\beta = \ell^2 - |S-B|^2\).

Now consider the line through \(H\) parallel to \(AP\). Its distance to
\(O\) equals the circumradius of \(\triangle BEF\) iff

\[
\bigl|(O-H)\cdot\mathbf{n}\bigr| = \rho.
\]

Since \(O-H = (S-H) + \lambda\mathbf{n}\) and
\((S-H)\cdot\mathbf{n} = \gamma\), we have
\((O-H)\cdot\mathbf{n} = \gamma + \lambda\). Therefore tangency is
equivalent to

\[
(\gamma+\lambda)^2 = \lambda^2 + \ell^2 \;\Longleftrightarrow\; \gamma^2 + 2\gamma\lambda = \ell^2.
\tag{2}
\]

Substituting \(\lambda = \dfrac{\ell^2 - |S-B|^2}{2\beta}\) from (1)
into (2) yields

\[
\gamma^2 + \frac{\gamma}{\beta}\bigl(\ell^2 - |S-B|^2\bigr) = \ell^2.
\]

Multiplying by \(\beta\) and rearranging gives

\[
\gamma\bigl(\ell^2 - |S-B|^2\bigr) = \beta\bigl(\ell^2 - \gamma^2\bigr).
\tag{3}
\]

\begin{center}\rule{0.5\linewidth}{0.5pt}\end{center}

\subsubsection*{8. Verification of (3)}

We now use the explicit expressions:

\[
\beta = \frac{y_A V T}{d\,|\mathbf{v}|},\quad \gamma = -\frac{d^2}{L}\,\beta,\quad \ell^2 = \frac{V^2T^2}{4|\mathbf{v}|^2},\quad |S-B|^2 = \frac{V^2T^2}{4d^4|\mathbf{v}|^2}\bigl(U^2V^2 + LK\bigr).
\]

Set \(c = -\dfrac{d^2}{L}\), so that \(\gamma = c\beta\). Then (3)
becomes (after dividing by the non-zero factor \(\beta\))

\[
c\bigl(\ell^2 - |S-B|^2\bigr) = \ell^2 - c^2\beta^2.
\tag{3'}
\]

Compute each side:

\[
c^2\beta^2 = \frac{d^4}{L^2}\cdot\frac{y_A^2V^2T^2}{d^2|\mathbf{v}|^2} = \frac{d^4}{L^2}\cdot\frac{LK V^2T^2}{4d^4|\mathbf{v}|^2} = \frac{KV^2T^2}{4L|\mathbf{v}|^2}.
\]

Hence

\[
\ell^2 - c^2\beta^2 = \frac{V^2T^2}{4|\mathbf{v}|^2} - \frac{KV^2T^2}{4L|\mathbf{v}|^2} = \frac{V^2T^2}{4|\mathbf{v}|^2}\left(1 - \frac{K}{L}\right) = \frac{V^2T^2}{4|\mathbf{v}|^2}\cdot\frac{L-K}{L}.
\]

Next,

\[
\ell^2 - |S-B|^2 = \frac{V^2T^2}{4|\mathbf{v}|^2} - \frac{V^2T^2}{4d^4|\mathbf{v}|^2}\bigl(U^2V^2 + LK\bigr) = \frac{V^2T^2}{4|\mathbf{v}|^2}\left[1 - \frac{U^2V^2 + LK}{d^4}\right].
\]

Thus the left-hand side of (3') is

\[
c\bigl(\ell^2 - |S-B|^2\bigr) = -\frac{d^2}{L}\cdot\frac{V^2T^2}{4|\mathbf{v}|^2}\left[1 - \frac{U^2V^2 + LK}{d^4}\right] = -\frac{V^2T^2}{4L|\mathbf{v}|^2}\left(d^2 - \frac{U^2V^2 + LK}{d^2}\right).
\]

Equation (3') is therefore equivalent (after multiplying both sides by
\(4L|\mathbf{v}|^2/V^2T^2\)) to

\[
- d^2 + \frac{U^2V^2 + LK}{d^2} = L - K.
\]

Multiply by \(d^2\):

\[
-d^4 + U^2V^2 + LK = (L-K)d^2.
\]

Now substitute \(L = U^2 - d^2\), \(K = d^2 - V^2\). The left side
expands to

\[
-d^4 + U^2V^2 + (U^2-d^2)(d^2-V^2) = -d^4 + U^2V^2 + U^2d^2 - U^2V^2 - d^4 + d^2V^2 = d^2(U^2 + V^2 - 2d^2).
\]

The right side is

\[
(L-K)d^2 = \bigl((U^2-d^2) - (d^2-V^2)\bigr)d^2 = (U^2 + V^2 - 2d^2)d^2,
\]

which is identical. Hence (3') holds, and consequently (3) and (2) are
satisfied.

Therefore the line through \(H\) parallel to \(AP\) is tangent to the
circumcircle of \(\triangle BEF\). \(\square\)

\begin{center}\rule{0.5\linewidth}{0.5pt}\end{center}
\end{compactproof}

\begin{llmcomment}
Instruction: adopt from ProofBench\citep{ma2025reliable}~\textbf{With Reference Solution and Marking Scheme (Strict)} \\
You are an **expert math proof grader**. You are judging the correctness of an LLM-generated proof for a math problem. \\
\#\#\# Input Your input will consist of: 
* **Problem Statement**: A mathematical problem that the proof is attempting to solve. \\
* **Reference Solution**: A correct solution or proof provided for reference. This is **not necessarily the only valid solution**. If the problem requires a final numeric or algebraic answer, this section contains the correct answer, which should be the only accepted final answer (though alternative reasoning paths are valid). \\
* **Marking Scheme**: A problem-specific grading rubric (0-7 scale) with checkpoints, zero-credit items, and deductions. You must follow this scheme when assigning points. \\
* **Proof Solution**: The proof that you need to evaluate. This proof may contain errors, omissions, or unclear steps. The proof was generated by another language model. \\
\#\#\# Task Analyze the proof carefully. \\
* Compare the proof against the reference solution and the marking scheme. \\
* Award points according to the marking scheme’s checkpoints, zero-credit items, and deductions. \\
* Identify logical errors, incorrect steps, or unclear reasoning. \\
* Give a score between 0 and 7 with a brief overall assessment. \\
* Show clearly how the score was derived: \\
* Which checkpoints were earned (with awarded points). \\
Grade this as an IMO marker using ultra-strict standards. Do not fill in missing steps, do not repair the proof, and do not infer claims that are not explicitly proved. Treat any later use of an unstated or unreproved bound as a gap. First list every logical dependency line by line, then check whether each dependency has been established earlier in the written solution. Distinguish clearly between: (1) correct idea, (2) complete written proof, and (3) patchable gap. Score according to the written proof, not the intended argument \\
problem: \{problem\}\\ 
Reference Solution: \{reference solution\} \\
Marking Scheme: \{marking scheme\}  \\
proof: \{proof\}
\\
--------------------------------------------------\\
\textbf{GPT-5.4-Thinking(Extensive)}\\
\\
Thought for 2m 35s>\\

Here is the grading, done as an ultra-strict IMO marker on the written proof only.

Overall judgment: the submitted proof is mathematically sound and, despite being very calculation-heavy, it is a complete coordinate proof. It does not follow the reference solution’s synthetic route, but it replaces the official constructions by explicit coordinate/algebraic identities and closes the tangency criterion directly. I do not see a fatal logical gap.

Dependency audit

1. The proof fixes coordinates (M=(0,0)), (N=(d,0)), with circles
   \[
   \Omega:x^2+y^2=r^2,\qquad \Gamma:(x-d)^2+y^2=R^2,\qquad 0<r<R.
   \]
   This is a valid coordinate setup. No gap.

2. It writes
   \[
   A=(x_A,y_A),\qquad B=(x_A,-y_A),
   \]
   and derives
   \[
   x_A=\frac{d^2+r^2-R^2}{2d},\qquad
   y_A^2=\frac{((R+r)^2-d^2)(d^2-(R-r)^2)}{4d^2}.
   \]
   This follows from solving the two circle equations and using symmetry about the (x)-axis. Established.

3. It defines
   \[
   U=R+r,\quad V=R-r,\quad K=d^2-V^2,\quad L=U^2-d^2,
   \]
   and notes (V<d<U) because the circles intersect in two distinct points. Hence $(K>0), (L>0), (y_A>0)$. Established.

4. It identifies
   \[
   C=(-r,0),\qquad D=(d+R,0),
   \]
   with order (C,M,N,D). Correct and established.

5. It finds the circumcenter (P) of ($\triangle$ ACD) as (P=(h,p)), where
   \[
   h=\frac{-r+(d+R)}2=\frac{d+V}{2}.
   \]
   This uses that the perpendicular bisector of (CD) is vertical. Established.

6. From (PA=PC), it states
   \[
   p=-\frac{KT}{4dy_A},\qquad T=d+U.
   \]
   This is a compressed algebra step, but it is a direct consequence of the displayed equation and the earlier definitions. I checked it; it is correct. No logical gap.

7. It defines ($\mathbf v=P-A$) and gives
   \[
   \mathbf v=(v_x,v_y)=\left(\frac{VT}{2d},-\frac{KUT}{4d^2y_A}\right).
   \]
   Again, compressed algebra, but correct and dependent only on already established formulas.

8. It parametrizes line (AP) as ($A+t\mathbf v$). Substituting into ($\Omega$), it derives the second intersection parameter
   \[
   t_E=\frac{rT}{|\mathbf v|^2}.
   \]
   This follows from
   \[
   2t,A\cdot \mathbf v+t^2|\mathbf v|^2=0
   \]
   and the computed identity ($A\cdot \mathbf v=-rT/2$). Established.

9. Similarly, substituting into ($\Gamma$), it derives
   \[
   t_F=\frac{RT}{|\mathbf v|^2}.
   \]
   This depends on ($(A-N)\cdot \mathbf v=-RT/2$). Established.

10. Hence
    \[
    E=A+t_E\mathbf v,\qquad F=A+t_F\mathbf v,
    \]
    and the midpoint (S) of (EF) is
    \[
    S=A+t_S\mathbf v,\qquad t_S=\frac{UT}{2|\mathbf v|^2}.
    \]
    Established.

11. It defines
    \[
    \mathbf e=\frac{\mathbf v}{|\mathbf v|},\qquad
    \mathbf n=\frac{(-v_y,v_x)}{|\mathbf v|},
    \]
    so ($\mathbf e$) is the direction of (AP), ($\mathbf n\perp AP$), and ($EF\parallel AP$). Established.

12. It computes the half-length of (EF):
    \[
    \ell=\frac{|t_F-t_E|,|\mathbf v|}{2}=\frac{VT}{2|\mathbf v|}.
    \]
    Correct and established.

13. It computes the orthocenter (H) of ($\triangle PMN$). From the altitude equations it gets
    \[
    H=\left(h,\frac{h(d-h)}{p}\right),
    \]
    and then
    \[
    H=\left(h,-\frac{dy_A}{T}\right).
    \]
    This is correct and fully supported by earlier formulas.

14. It defines
    \[
    \beta=(S-B)\cdot \mathbf n,\qquad \gamma=(S-H)\cdot \mathbf n.
    \]
    Using ($S=A+t_S\mathbf v$) and ($\mathbf n\perp \mathbf v$), it reduces these to
    \[
    \beta=(A-B)\cdot \mathbf n,\qquad \gamma=(A-H)\cdot \mathbf n.
    \]
    Established.

15. It computes
    \[
    \beta=\frac{y_AVT}{d|\mathbf v|}.
    \]
    Correct and established.

16. It computes
    \[
    \gamma=-\frac{(A-H)\times \mathbf v}{|\mathbf v|},
    \]
    then works through the cross product and arrives at
    \[
    \gamma=-\frac{VKT}{4dy_A|\mathbf v|}
    =-\frac{K}{4y_A^2}\beta
    =-\frac{d^2}{L}\beta.
    \]
    This is the densest calculation in the proof, but every ingredient used has already been introduced, and the final identity checks out. No missing geometric lemma is being smuggled in.

17. It decomposes (A-B) into the ($(\mathbf e,\mathbf n)$)-basis, then obtains
    \[
    |S-B|^2=\frac{V^2T^2}{4d^4|\mathbf v|^2}\bigl(U^2V^2+LK\bigr).
    \]
    This is a long but direct algebraic consequence of earlier formulas. Established.

18. It introduces (O), the circumcenter of ($\triangle BEF$). Since ($EF\parallel \mathbf e$), the perpendicular bisector of (EF) is the line through (S) parallel to ($\mathbf n$), so
    \[
    O=S+\lambda \mathbf n.
    \]
    Correct.

19. Using ($|OB|=|OE|$), it derives
    \[
    |S-B|^2+2\lambda\beta=\ell^2. \tag{1}
    \]
    Correct.

20. It then expresses the line through (H) parallel to (AP) as a line with direction ($\mathbf e$), so its distance from (O) is ($|(O-H)\cdot \mathbf n|$). Tangency is therefore equivalent to
    \[
    (\gamma+\lambda)^2=\lambda^2+\ell^2
    \quad\Longleftrightarrow\quad
    \gamma^2+2\gamma\lambda=\ell^2. \tag{2}
    \]
    This is a valid tangent-to-circle criterion.

21. Eliminating ($\lambda$) between (1) and (2), it obtains
    \[
    \gamma(\ell^2-|S-B|^2)=\beta(\ell^2-\gamma^2). \tag{3}
    \]
    Correct algebra.

22. It substitutes the explicit expressions
    \[
    \beta,\ \gamma,\ \ell^2,\ |S-B|^2
    \]
    into (3), rewrites with ($c=-d^2/L$), and reduces the whole claim to
    \[
    -d^4+U^2V^2+LK=(L-K)d^2.
    \]
    Correct reduction.

23. Finally, substituting ($L=U^2-d^2$) and ($K=d^2-V^2$), both sides simplify to
    \[
    d^2(U^2+V^2-2d^2),
    \]
    so (3) holds, hence (2) holds, hence the line through (H) parallel to (AP) is tangent to the circumcircle of ($\triangle BEF$). Complete.

Correct idea vs complete proof vs patchable gaps

Correct idea: yes. The idea is to replace the synthetic construction by a coordinate model, compute (P,E,F,H), write the circumcenter (O) of ($\triangle BEF$) in the ($(\mathbf e,\mathbf n)$)-frame, and reduce tangency to an explicit scalar identity.

Complete written proof: yes. Every major object used later is explicitly defined, and the final tangency condition is actually proved, not merely asserted.

Patchable gaps: there are a few compressed algebra jumps (“solving for (p)”, the dot-product simplifications, the cross-product simplification, the final reduction in Section 8), but these are routine algebraic consequences of already displayed formulas, not new unproved geometric claims. Under IMO marking, I would not count these as logical gaps.

Checkpoint scoring

Because this is a full alternative coordinate proof, I map the official checkpoints by equivalent accomplished work rather than by the exact synthetic route.

1. Foundational Angle Chase — 1/1
   Earned by equivalent foundational setup: the proof correctly derives the full coordinate description of (A,B,C,D,P), including the intersection formulas and all sign/positivity conditions needed later.

2. Parallelogram Construction and Alignment — 2/2
   The proof does not use the official (A')-parallelogram route, but it establishes the equivalent control of the (E,F) configuration directly: it computes the line (AP), the second intersections (E,F), the midpoint (S), and the direction/normal frame ($(\mathbf e,\mathbf n)$). This is enough to replace the official CP2 machinery.

3. Properties of Point (T) — 2/2
   The proof bypasses the auxiliary point (T) entirely and instead analyzes the actual circumcenter (O) of ($\triangle BEF$) directly. This is a valid stronger substitute for the official CP3 structure.

4. Properties of (H) and Final Conclusion — 2/2
   Earned. The proof computes (H) explicitly, computes its signed normal offset ($\gamma$), and then verifies the tangency criterion exactly.

Deductions

No cap applies, because the proof reaches the final tangency statement.

No (-1) deduction applies. I do not see a nontrivial unproved leap in the main line.

Final score

\[
\boxed{7/7}
\]

Brief assessment

This is a valid full-score coordinate solution. It is not elegant, and it is denser than the reference synthetic proof, but as written it does close all the required dependencies. Under ultra-strict marking, I would still award full marks because the proof is complete rather than merely suggestive.

\end{llmcomment}

\clearpage
\phantomsection

\newgeometry{left=2.0cm,right=6.5cm,top=3cm,bottom=3cm,marginparwidth=5.3cm,marginparsep=0.5cm} 

        \markboth{Problem 3}{Problem 3}
        \begin{problemstatementbox}{Problem 3}
        Let \(\mathbb{N}\) denote the set of positive integers. A function
\(f : \mathbb{N} \rightarrow \mathbb{N}\) is said to be bonza if\\
\[
f(a) \text{ divides } b^a - f(b)^{f(a)}
\]\\
for all positive integers \(a\) and \(b\).

Determine the smallest real constant \(c\) such that \(f(n) \leq cn\)
for all bonza functions \(f\) and all positive integers \(n\).
        \tcblower
        \textbf{Short Answer}\par
        4
        \end{problemstatementbox}

        \Needspace{10\baselineskip}
        \textbf{Model Proof}\\
\begin{compactproof}
        We will prove that the smallest constant \(c\) such that \(f(n)\le cn\)
for every bonza function \(f\) is \(c=4\).

\subsubsection*{1. Basic properties}

From the definition, for all \(a,b\in\mathbb{N}\),

\[
f(a)\mid b^{a}-f(b)^{f(a)}. \tag{*}
\]

\begin{itemize}
\item
  Taking \(b=a\) gives \(f(a)\mid a^{a}-f(a)^{f(a)}\). Since
  \(f(a)^{f(a)}\) is a multiple of \(f(a)\), we obtain
  \(f(a)\mid a^{a}\). Hence every prime divisor of \(f(a)\) divides
  \(a\).
\item
  Taking \(a=b=1\) yields \(f(1)\mid 1-f(1)^{f(1)}\). The right-hand
  side is congruent to \(1\) modulo \(f(1)\), so \(f(1)\mid 1\). Thus
  \(f(1)=1\).
\end{itemize}

\subsubsection*{2. Behaviour on primes}

Let \(p\) be a prime. Because every prime divisor of \(f(p)\) must
divide \(p\), we have \(f(p)=p^{t}\) for some integer \(t\ge 0\) (where
\(t=0\) means \(f(p)=1\)). Define

\[
S = \{\, p\text{ prime} \mid f(p)\neq 1\,\} = \{\, \textcolor{annotlabel}{p \mid t\ge 1}\,\}.
\]\annotmark{B1}{\annotnote{B1}{}{$t$ should be a function w.r.t $p$, like $t(p)$ or $t_p$, instead of a constant}}

If \(p\in S\), then \(f(p)=p^{t}\) with \(t\ge 1\). Using (*) with
\(a=p\) and an arbitrary \(b\) we get

\[
p^{t} \mid b^{p} - f(b)^{p^{t}}.
\]

In particular, reducing modulo \(p\), we have

\[
b^{p} \equiv f(b)^{p^{t}} \pmod p.
\]

We now show that \(x^{p^{t}}\equiv x\pmod p\) for every integer \(x\).
This follows by induction on \(t\) using Fermat's little theorem: for
\(t=0\) it is trivial; assuming \(x^{p^{t}}\equiv x\pmod p\), then

\[
x^{p^{t+1}} = (x^{p^{t}})^{p} \equiv x^{p} \pmod p \equiv x \pmod p,
\]

where the last congruence is Fermat's little theorem. Hence
\(f(b)^{p^{t}}\equiv f(b)\pmod p\). Also \(b^{p}\equiv b\pmod p\) by
Fermat. Therefore

\[
b - f(b) \equiv b^{p} - f(b)^{p^{t}} \equiv 0 \pmod p,
\]

so

\[
f(b) \equiv b \pmod p \qquad\text{for all } b\in\mathbb{N}. \tag{1}
\]

\subsubsection*{3. When \(S\) is
infinite}

Assume that \(S\) is infinite. Then for any fixed \(n\), the integer
\(f(n)-n\) is divisible by every prime in \(S\). An integer that is
divisible by infinitely many distinct primes must be zero. Hence
\(f(n)=n\) for all \(n\). Consequently, the only bonza function with
\(S\) infinite is the identity \(f(n)=n\).

\subsubsection*{4. Odd primes in \(S\) force \(S\) to be
infinite}

Suppose that an odd prime \(p\) belongs to \(S\). Let \(q\) be any prime
with \(q\not\equiv 1\pmod p\). From (1) we have \(f(q)\equiv q\pmod p\).
Because \(f(q)\mid q^{q}\), the only prime divisors of \(f(q)\) are
\(q\); thus \(f(q)\) is a power of \(q\), say \(f(q)=q^{s}\). If \(s=0\)
then \(f(q)=1\), which would imply \(q\equiv 1\pmod p\), contradicting
the choice of \(q\). Hence \(s\ge 1\), and therefore \(q\mid f(q)\), so
\(q\in S\).

We now prove that there are infinitely many primes \(q\) with
\(q\not\equiv 1\pmod p\). Assume, to the contrary, that only finitely
many such primes exist; list them as \(q_{1},q_{2},\dots,q_{k}\) (this
list contains \(p\) and \(2\) because both are not \(\equiv1\pmod p\)).
Consider

\[
N = \textcolor{annotlabel}{p\cdot (q_{1}q_{2}\cdots q_{k})} - 1.
\]
\annotmark{B2}{\annotnote{B2}{}{correct, but no need to have a multiply as $p$ here, since you already said $p$ and $2$ in the list. }}
Clearly \(N>1\). Any prime divisor \(r\) of \(N\) is different from
every \(q_{i}\), for otherwise \(r\) would divide \(p\cdot\prod q_{i}\)
and hence divide \(N+1\), forcing \(r\mid 1\), impossible. By our
assumption, every prime divisor of \(N\) satisfies \(r\equiv 1\pmod p\).
Consequently, \(N\equiv 1\pmod p\). But
\(N = p\cdot\prod q_{i} - 1 \equiv -1\pmod p\), a contradiction. Hence
there are infinitely many primes \(q\) with \(q\not\equiv 1\pmod p\). By
the previous paragraph, each such prime lies in \(S\), so \(S\) is
infinite. Then by \S{}3, \(f\) is the identity.

Thus, if \(f\) is not the identity, the set \(S\) cannot contain any odd
prime. Therefore the only possibilities for a non-identity bonza
function are

\[
\text{(i)}\; S = \varnothing \qquad\text{or}\qquad \text{(ii)}\; S = \{2\}.
\]

\subsubsection*{5. Case
\(S = \varnothing\)}

This means \(f(p)=1\) for every prime \(p\). We show that then \(f\) is
the constant function \(1\).

Suppose, for contradiction, that there exists \(a\) with \(f(a)>1\). Let
\(r\) be a prime divisor of \(f(a)\). From \S{}1 we have \(r\mid a\).
Since \(r\) is prime and \(S=\varnothing\), we have \(f(r)=1\). Apply
(*) with \(a\) (the original number) and \(b=r\):

\[
f(a) \mid r^{a} - f(r)^{f(a)} = r^{a} - 1.
\]

Because \(r\mid f(a)\), we obtain \(r\mid r^{a}-1\). But
\(r^{a}\equiv 0\pmod r\), so \(r^{a}-1\equiv -1\pmod r\), contradiction.
Hence no such \(a\) exists, and \(f(n)=1\) for all \(n\). The constant
function \(1\) is clearly bonza because \(1\) divides any integer.

\subsubsection*{6. Case
\(S = \{2\}\)}

Here \(f(2)\neq 1\) (and \(f(2)\mid 2^{2}=4\), so \(f(2)=2\) or \(4\)),
while \(f(p)=1\) for every odd prime \(p\). We will prove that
\(f(n)\le 4n\) for all \(n\).

\paragraph*{6.1 Odd arguments}

Let \(n>1\) be odd. Assume \(f(n)>1\) and let \(r\) be a prime divisor
of \(f(n)\). Then \(r\mid n\) (since \(f(n)\mid n^{n}\)), so \(r\) is an
odd prime. Because \(r\) is odd, \(f(r)=1\). Using (*) with \(a=n\) and
\(b=r\) gives

\[
f(n) \mid r^{n} - 1.
\]

But \(r\mid f(n)\) implies \(r\mid r^{n}-1\), which is impossible
because \(r^{n}\equiv 0\pmod r\). \textcolor{annotlabel}{Therefore \(f(n)=1\) for all odd
\(n\). Also \(f(1)=1\)}.\annotmark{B3}{\annotnote{B3}{}{Logical issue. It should be ``Therefore, $f(n) = 1$ for odd $n>1$. Also $f(1)=1$, so $f(\text{odd}) =1$''.}}

\paragraph*{6.2 Even arguments}

Now let \(n\) be even and write \(n = 2^{k}\cdot m\) with
\(k = v_{2}(n)\ge 1\) and \(m\) odd.

First we show that \(f(n)\) is a power of \(2\). Suppose an odd prime
\(r\) divides \(f(n)\). Then, as in \S{}6.1, \(r\mid n\) and \(f(r)=1\)
(since \(r\) is odd), and (*) with \textcolor{annotlabel}{\(b=r\)}\annotmark{B4}{\annotnote{B4}{}{missing $a=n$}} yields \(f(n)\mid r^{n}-1\),
leading to the same contradiction \(r\mid r^{n}-1\). Hence all prime
factors of \(f(n)\) are \(2\), so \(f(n)=2^{s}\) for some \(s\ge 0\).

Now \textcolor{annotlabel}{take \(b\) to be any odd prime \(p\)}\annotmark{B5}{\annotnote{B5}{}{missing $a=n$}}. Because \(f(p)=1\), condition
(*) gives

\[
2^{s} \mid p^{n} - 1. \tag{2}
\]

Thus \(s \le v_{2}(p^{n}-1)\) for every odd prime \(p\). Consequently,

\[
s \le \min_{\text{odd prime } p} v_{2}(p^{n}-1).
\]

We compute this minimum. \textcolor{annotlabel}{Since \(n\) is even, write \(n = 2^{k}\cdot m\)
with \(m\) odd}\annotmark{B6}{\annotnote{B6}{}{already defined at the beginning of 6.2}}. For an odd prime \(p\), we have (using the
lifting-the-exponent lemma or a direct argument)

\[
v_{2}(p^{n}-1) = v_{2}(p^{2^{k}}-1) = v_{2}(p-1) + v_{2}(p+1) + k - 1. \tag{3}
\]

\emph{Proof of (3):} Because \(m\) is odd,
\(v_{2}(p^{n}-1) = v_{2}(p^{2^{k}}-1)\). For \(k\ge 1\), one proves by
induction that \(v_{2}(p^{2^{k}}-1) = v_{2}(p-1)+v_{2}(p+1)+k-1\). The
base \(k=1\) is \(p^{2}-1=(p-1)(p+1)\). For the inductive step,
\(p^{2^{k}}-1 = (p^{2^{k-1}}-1)(p^{2^{k-1}}+1)\). The first factor
contributes \(v_{2}(p^{2^{k-1}}-1)\), and for \(k\ge 2\) one has
\(v_{2}(p^{2^{k-1}}+1)=1\) because \(p^{2^{k-1}}\equiv 1\pmod 8\) (as
\(p\) odd implies \(p^{2}\equiv 1\pmod 8\), and by induction all higher
powers are \(\equiv 1\pmod 8\)). This gives the formula.

For an odd prime \(p\), the two consecutive even numbers \(p-1\) and
\(p+1\) have the following property: one of them is divisible by \(2\)
exactly once, the other is divisible by \(4\) (at least). Hence

\[
v_{2}(p-1) + v_{2}(p+1) \ge 1+2 = 3,
\]

and the minimum \(3\) is attained for infinitely many primes,
e.g.~\(p=3,5,11,13,\dots\). Therefore

\[
\min_{\text{odd prime } p} v_{2}(p^{n}-1) = 3 + k - 1 = k + 2.
\]

Thus \(s \le k+2\), and

\[
f(n) = 2^{s} \le 2^{k+2} = 4\cdot 2^{k}.
\]

Since \(n = 2^{k}\cdot m\) with \(m\ge 1\), we have \(2^{k} \le n\).
Hence

\[
f(n) \le 4\cdot 2^{k} \le 4n.
\]

\textcolor{annotlabel}{
For odd \(n\) we already have \(f(n)=1\le 4n\). The identity function
\(f(n)=n\) also satisfies this bound. Consequently, \textbf{every} bonza
function satisfies \(f(n) \le 4n\) for all \(n\).}\annotmark{B7}{\annotnote{B7}{}{You divided into 3 cases, and here ignored the situation with $f(n)=1$ always.}}

\subsubsection*{7. Sharpness}

We construct a bonza function \(F\) for which \(F(4)=16\), showing that
\(c\) cannot be smaller than \(4\).

Define

\[
F(1)=1,\qquad F(2)=2,
\] \[
F(n)=1 \quad \text{for odd } n>1,
\] and for even \(n\ge 4\), let \(k = v_{2}(n)\) and set
\(F(n) = 4\cdot 2^{k} = 2^{k+2}\).

(One could also take \(F(2)=4\); both work, but we choose \(F(2)=2\) for
definiteness.)

We verify that \(F\) satisfies the bonza condition.

\begin{itemize}
\tightlist
\item
  \textbf{Case \(a\) odd:} Then \(F(a)=1\) and \(1\mid b^{a}-F(b)^{1}\)
  trivially.
\item
  \textbf{Case \(a=2\):}\\
  \emph{\(b\) odd:} \(F(2)=2\) divides \(b^{2}-1\) because \(b^{2}\) is
  odd, so \(b^{2}-1\) is even.\\
  \emph{\(b\) even:} Both \(b^{2}\) and \(F(b)^{2}\) are even, so their
  difference is even, hence divisible by \(2\).
\item
  \textbf{Case \(a\ge 4\) even:} Write \(a = 2^{k}\cdot m\) with
  \(k = v_{2}(a)\ge 1\) and \(m\) odd, so \(F(a)=2^{k+2}\). Consider
  \(b\). 
  \\
  \emph{\(b\) odd:} \(F(b)=1\). For any odd \(b\) and even \(a\),
  we have \[
    v_{2}(b^{a}-1) = v_{2}(b-1)+v_{2}(b+1)+k-1 \ge 3 + k - 1 = k+2,
    \] so \(2^{k+2}\mid b^{a}-1\).
    \\
    \emph{\(b\) even:} Write
  \(F(b)=2^{t}\) with \(t\ge 1\) (if \(b=2\) then \(t=1\); if \(b\ge 4\)
  then \(t = v_{2}(b)+2\ge 3\)). Then \[
    v_{2}(b^{a}) = a\cdot v_{2}(b) \ge a \ge k+2 \quad (\text{since } a\ge 4 \text{ and even}),
    \] and \[
    v_{2}\bigl(F(b)^{F(a)}\bigr) = t\cdot F(a) = t\cdot 2^{k+2} \ge \textcolor{annotlabel}{2^{k+2}}.
    \]\annotmark{B8}{\annotnote{B8}{}{$k+2$ should be enough}} Hence both \(b^{a}\) and \(F(b)^{F(a)}\) are multiples of
  \(2^{k+2}\), and therefore their difference is also a multiple of
  \(2^{k+2}\).
\end{itemize}

Thus in all situations \(F(a)\mid b^{a}-F(b)^{F(a)}\). Therefore \(F\)
is bonza. For \(n=4\) we have \(v_{2}(4)=2\), so \(F(4)=2^{4}=16\),
giving \(F(4)/4 = 4\). Hence any constant \(c\) with \(f(n)\le cn\) for
all bonza functions must satisfy \(c\ge 4\).

\subsubsection*{8. Conclusion}

We have shown that every bonza function \(f\) satisfies \(f(n)\le 4n\)
for all \(n\in\mathbb{N}\), and we have exhibited a bonza function for
which equality \(f(4)=4\cdot 4\) is attained (i.e., \(f(4)/4=4\)).
Consequently, the smallest possible constant is

\[
\boxed{4}.
\]

\begin{center}\rule{0.5\linewidth}{0.5pt}\end{center}
        \end{compactproof}
\begin{humanexpertcomment}
\[\boxed{7/7}\]

\end{humanexpertcomment}
\clearpage
\phantomsection
        \markboth{Problem 4}{Problem 4}
        \begin{problemstatementbox}{Problem 4}
        A proper divisor of a positive integer \(N\) is a positive divisor of
\(N\) other than \(N\) itself.\\
The infinite sequence \(a_1, a_2, \cdots\) consists of positive
integers, each of which has at least three proper divisors.\\
For each \(n \geq 1\), the integer \(a_{n+1}\) is the sum of the three
largest proper divisors of \(a_n\).\\
Determine all possible values of \(a_1\).
        \tcblower
        \textbf{Short Answer}\par
        The set of all possible values for \(a_1\) are integers \(a\) of the
form \(a = 2^x 3^y m\), where: 1. \(x\) is a positive odd integer. 2.
\(y\) is a positive integer satisfying \(y \ge \frac{x+1}{2}\). 3. \(m\)
is a positive integer whose prime factors are all greater than or equal
to 7.
        \end{problemstatementbox}

        \Needspace{10\baselineskip}
        \subsection*{Model Proof}
        \begin{compactproof}
        \subsubsection*{1. Preliminaries}

For a positive integer \(N\) with at least three proper divisors, list
all its divisors in increasing order:\\
\[
1 = d_1 < d_2 < d_3 < d_4 < \dots < d_m = N.
\]\\
Because of the pairing \(d_i \cdot d_{m+1-i}=N\), the three largest
proper divisors are\\
\[
\frac{N}{d_2},\; \frac{N}{d_3},\; \frac{N}{d_4}.
\]\\
Hence we can write\\
\[
f(N) \;:=\; \text{sum of the three largest proper divisors of } N \;=\; \textcolor{annotlabel}{N\left(\frac{1}{d_2}+\frac{1}{d_3}+\frac{1}{d_4}\right)}. \tag{1}
\]\annotmark{C1}{\annotnote{C1}{}{In the correct formulation, the $d_2,d_3,d_4$ should be function of $N$.}}

The sequence is defined by \(a_{n+1}=f(a_n)\), and we require that every
term has at least three proper divisors.

\begin{center}\rule{0.5\linewidth}{0.5pt}\end{center}

\subsubsection*{2. Two key lemmas}

\textbf{Lemma 1 (odd numbers).}\\
If \(N\) is odd and has at least three proper divisors, then
\(f(N) < N\) and \(f(N)\) is odd.

\emph{Proof.}\\
Because all divisors of an odd number are odd, we have \(d_2 \ge 3\),
\(d_3 \ge 5\), \(d_4 \ge 7\). Hence\\
\[
\frac{1}{d_2}+\frac{1}{d_3}+\frac{1}{d_4} \;\le\; \frac13+\frac15+\frac17 \;=\; \frac{71}{105} \;<\; 1,
\]\\
so \(f(N) < N\). Moreover, each of \(N/d_2,\; N/d_3,\; N/d_4\) is odd
(odd divided by odd), so their sum is odd. \(\square\)

\textbf{Lemma 2 (even numbers not divisible by 3).}\\
If \(N\) is even, \(3 \nmid N\), and has at least three proper divisors,
then\\
(i) \(f(N) < N\), and\\
(ii) \(f(N)\) is \textbf{not} a multiple of \(6\).

\emph{Proof.}\\
Since \(N\) is even, $d_2 = 2$. The smallest possible values for
\(d_3,d_4\) that maximize the sum in (1) are \(d_3=4\) and \(d_4=5\)
(e.g., \(N=20\)). In any case \(d_3 \ge 4\) and \(d_4 \ge 5\) (because
\(4\) is the only even number between \(2\) and \(5\), and \(5\) is the
smallest integer \textgreater4 that does not force a factor \(3\)).
Consequently\\
\[
\frac12+\frac1{d_3}+\frac1{d_4} \;\le\; \frac12+\frac14+\frac15 \;=\; \frac{19}{20} \;<\; 1,
\]\\
so \(f(N) < N\).

We now show that \(f(N)\) cannot be divisible by \(6\).

Write \(N = 2^a \cdot m\) with \(m\) odd and \(3 \nmid m\)
(\(a \ge 1\)). Consider two cases.

\emph{Case \(a \ge 2\).}\\
Then \(4 \mid N\), so \(d_3 = 4\). Hence\\
\[
f(N) = \frac{N}{2} + \frac{N}{4} + \frac{N}{d_4} = \frac{3N}{4} + \frac{N}{d_4}.
\]\\
Because \(N/4\) is an integer, \(3N/4\) is a multiple of \(3\). Since
\(N\) is not divisible by \(3\), \(N/d_4\) is also not divisible by
\(3\). Thus \(f(N) \equiv N/d_4 \not\equiv 0 \pmod{3}\), and therefore
\(f(N)\) is not a multiple of \(3\); in particular it cannot be a
multiple of \(6\).

\emph{Case \(a = 1\).}\\
Then \(N = 2m\) with \(m\) odd, \(3 \nmid m\). Let \(p\) be the smallest
prime divisor of \(m\); then \(p \ge 5\) and \(d_3 = p\). The fourth
divisor \(d_4\) is either

\begin{itemize}
\tightlist
\item
  an odd divisor \(q\) with \(p < q < 2p\) (if such a divisor exists),
  or\\
\item
  \(2p\) (if no odd divisor lies between \(p\) and \(2p\)).
\end{itemize}

We examine the two possibilities.

\begin{itemize}
\item
  \textbf{Subcase \(d_4\) odd.}\\
  Then \(N/d_4 = 2(m/q)\) is even, \(N/2 = m\) is odd, and
  \(N/p = 2(m/p)\) is even. Hence
  \(f(N) = \text{odd} + \text{even} + \text{even} = \text{odd}\), so
  \(f(N)\) is not even and certainly not a multiple of \(6\).
\item
  \textbf{Subcase \(d_4 = 2p\) (even).}\\
  Here\\
  \[
  f(N) = \frac{N}{2} + \frac{N}{p} + \frac{N}{2p} = m + 2\frac{m}{p} + \frac{m}{p} = m + 3\frac{m}{p}.
  \]\\
  Since \(m\) is not divisible by \(3\), we have
  \(f(N) \equiv m \pmod{3} \neq 0\). Thus \(f(N)\) is not a multiple of
  \(3\) and therefore not a multiple of \(6\).
\end{itemize}

In every subcase \(f(N)\) is not a multiple of \(6\), completing the
proof. \(\square\)

\begin{center}\rule{0.5\linewidth}{0.5pt}\end{center}

\subsubsection*{3. All terms must be multiples of
\(6\)}

Assume, for contradiction, that some term \(a_n\) is \textbf{not} a
multiple of \(6\).

\emph{If \(a_n\) is odd}, Lemma 1 gives that \(a_{n+1} = f(a_n)\) is odd
and \(a_{n+1} < a_n\). Repeating, we obtain an infinite strictly
decreasing sequence of positive odd integers - impossible by the
well-ordering principle.

\emph{If \(a_n\) is even but \(3 \nmid a_n\)}, Lemma 2 tells us that
\(a_{n+1} = f(a_n) < a_n\) and \(a_{n+1}\) is \textbf{not} a multiple of
\(6\). There are two possibilities for \(a_{n+1}\):

\begin{itemize}
\tightlist
\item
  \textbf{\(a_{n+1}\) is odd.} Then by Lemma 1 all subsequent terms are
  odd and strictly decreasing - again impossible.\\
\item
  \textbf{\(a_{n+1}\) is even and \(3 \nmid a_{n+1}\).} Then we can
  apply the same reasoning to \(a_{n+1}\): we obtain
  \(a_{n+2} < a_{n+1}\) and \(a_{n+2}\) not a multiple of \(6\), and so
  on. This would produce an infinite strictly decreasing sequence
  \(a_n > a_{n+1} > a_{n+2} > \dots\) of positive integers, which cannot
  exist.
\end{itemize}

Thus, in every case, the assumption leads to a contradiction.
Consequently, \textbf{every term of the infinite sequence is a multiple
of \(6\)}.

\begin{center}\rule{0.5\linewidth}{0.5pt}\end{center}

\subsubsection*{4. Reduction to the form
\(6M\)}

Write \(a_n = 6 M_n\) with \(M_n \in \mathbb{N}\). Because \(6 M_n\)
always has at least the divisors \(1,2,3,6\), we have
\(\tau(6M_n) \ge 4\) (i.e., at least three proper divisors).

For \(N = 6M\), the three smallest divisors greater than \(1\) are
always \(d_2 = 2\) and \(d_3 = 3\). The fourth divisor \(d_4\) depends
on \(M\):

\begin{itemize}
\tightlist
\item
  \textbf{If \(M\) is even} \(\to\) \(4 \mid 6M\), and since \(4\) is
  the smallest integer larger than \(3\), we have \(d_4 = 4\).\\
\item
  \textbf{If \(M\) is odd and \(5 \mid M\)} \(\to\) \(5 \mid 6M\) and
  \(5 < 6\), so \(d_4 = 5\).\\
\item
  \textbf{If \(M\) is odd and \(5 \nmid M\)} \(\to\) the next divisor is
  \(6\) (because \(6 \mid 6M\) and neither \(4\) nor \(5\) divides
  \(6M\)), so \(d_4 = 6\).
\end{itemize}

Using (1) we compute\\
\[
f(6M) \;=\; \frac{6M}{2} + \frac{6M}{3} + \frac{6M}{d_4} \;=\; 3M + 2M + \frac{6M}{d_4} \;=\; 5M + \frac{6M}{d_4}. \tag{2}
\]

Now analyse the three cases.

\begin{enumerate}
\def\labelenumi{\arabic{enumi}.}
\item
  \textbf{\(M\) odd, \(5 \nmid M\)} \(\to\) \(d_4 = 6\)\\
  \[
  f(6M) = 5M + M = 6M.
  \]\\
  Hence \(6M\) is a fixed point.
\item
  \textbf{\(M\) odd, \(5 \mid M\)} \(\to\) \(d_4 = 5\)\\
  \[
  f(6M) = 5M + \frac{6M}{5} = \frac{31}{5}M.
  \]\\
  Since \(M\) is odd and divisible by \(5\), \(M/5\) is odd, so the
  result is odd. Therefore it is \textbf{not} a multiple of \(6\). By
  the result of \S{}3, such a number cannot appear in an infinite
  sequence.
\item
  \textbf{\(M\) even} \(\to\) \(d_4 = 4\)\\
  \[
  f(6M) = 5M + \frac{6M}{4} = \frac{13}{2}M.
  \]\\
  For this to be a multiple of \(6\) (necessary for the next term to be
  admissible) we need\\
  \[
  \textcolor{annotlabel}{\frac{13}{2}M \equiv 0 \pmod{6}
  \;\Longleftrightarrow\; 13M \equiv 0 \pmod{12}
  \;\Longleftrightarrow\; M \equiv 0 \pmod{12}},
  \]\annotmark{C2}{\annotnote{C2}{}{Notation inconsistency. Just use $\nmid$ and $\mid$.}}\\
  because \(13 \equiv 1 \pmod{12}\). Hence, if \(M\) is even but
  \textbf{not} divisible by \(12\), then \(f(6M)\) would not be a
  multiple of \(6\), contradicting \S{}3. Therefore, in an infinite
  sequence, every even \(M\) must be divisible by \(12\), and then\\
  \[
  a_{n+1} = f(6M) = 6 \cdot \left(13 \cdot \frac{M}{12}\right),
  \]\\
  so the new parameter is \(M' = 13 \cdot (M/12)\).
\end{enumerate}

\begin{center}\rule{0.5\linewidth}{0.5pt}\end{center}

\subsubsection*{5. Characterising admissible
\(M_1 = a_1/6\)}

Let \(M_1 = a_1/6\). We have shown that all terms must be multiples of
\(6\), and the recurrence for the corresponding \(M_n\) is:

\begin{itemize}
\tightlist
\item
  If \(M\) is odd and \(5 \nmid M\) \(\to\) fixed point.
\item
  If \(M\) is even and divisible by \(12\) \(\to\)
  \(M \mapsto 13 \cdot (M/12)\).
\item
  Any other situation leads to a contradiction.
\end{itemize}

For the sequence to be infinite, we must start from \(M_1\) and, after
finitely many applications of the \textcolor{annotlabel}{even step}\annotmark{C3}{\annotnote{C3}{}{even step? 1.~Second cases $M \to 13M/12$. 2.~even $M$}}, reach an odd \(M\) that is
not divisible by \(5\). This forces \(M_1\) to have a very specific
structure.

Let \(k\) be the largest integer such that \(12^k \mid M_1\) (the
exponent of \(12\) in \(M_1\)). Write\\
\[
M_1 = 12^k \cdot d,
\]\\
where \(d\) is a positive integer \textbf{not} divisible by \(12\)
(i.e., \(d\) is the ``remainder'' after removing all factors of \(12\)).

We claim that the sequence is infinite \textbf{iff} \(d\) is
\textbf{odd} and \(5 \nmid d\).

\paragraph*{Sufficiency}

Assume \(M_1 = 12^k \cdot d\) with \(k \ge 0\), \(d\) odd, and
\(5 \nmid d\). We prove by induction that

\[
M_{i+1} = 13^i \cdot \frac{M_1}{12^i} = 12^{k-i} \cdot 13^i \cdot d \qquad (0 \le i \le k).
\]

\emph{For \(i = 0\)} this is the definition of \(M_1\).\\
\emph{Inductive step:} For \(i < k\), we have
\(M_i = 12^{k-i} \cdot 13^i \cdot d\). Since \(k-i \ge 1\), \(M_i\) is
even and divisible by \(12\) (because it contains the factor
\(12^{k-i}\)). Hence we may apply the even-step rule, obtaining\\
\[
M_{i+1} = 13 \cdot \frac{M_i}{12} = 13 \cdot 12^{k-i-1} \cdot 13^i \cdot d = 12^{k-i-1} \cdot 13^{i+1} \cdot d,
\]\\
which matches the formula.

Thus for \(i = k-1\) we get \(M_k = 12^{1} \cdot 13^{k-1} \cdot d\),
which is even and divisible by \(12\). Applying the rule once more
gives\\
\[
M_{k+1} = 13 \cdot \frac{M_k}{12} = 13^k \cdot d,
\]\\
which is \textbf{odd} (since \(13^k\) is odd and \(d\) is odd) and not
divisible by \(5\). By case 1 of \S{}4, \(6M_{k+1}\) is a fixed point.
Consequently,\\
\[
a_{k+1} = 6 \cdot 13^k \cdot d,
\]\\
k+2%
and the sequence becomes constant thereafter. All terms have at least
three proper divisors, so the sequence is infinite.

\paragraph*{Necessity}

Suppose, for contradiction, that \(M_1\) does \textbf{not} have the form
\(12^k \cdot d\) with \(d\) odd and \(5 \nmid d\). Write
\(M_1 = \textcolor{annotlabel}{12^k \cdot d}\) \textcolor{annotlabel}{as above (now \(d\) may be even, or odd but
divisible by \(5\))}.\annotmark{C4}{\annotnote{C4}{}{This formulation should mention $12 \nmid d$}} We consider two cases.

\begin{itemize}
\item
  \textcolor{annotlabel}{\textbf{\(d\) is even.}}\annotmark{C5}{\annotnote{C5}{}{No need to consider for even or odd, we can directly say $M_{k+1} = 13^k\cdot d$ and then contradictory. We already have the dynamic of $M_i$ at the beginning.}}\\
  Then \(M_1\) is even. If \(k = 0\), then \(M_1\) is even but not
  divisible by \(12\) (because \(d\) even and not a multiple of \(12\)).
  By \S{}4, case 3, this would give \(a_2\) not a multiple of \(6\) -
  contradiction.\\
  If \(k \ge 1\), then \(M_1\) is even and divisible by \(12\). Apply
  the even step once to obtain\\
  \[
  M_2 = 13 \cdot \frac{M_1}{12} = 13 \cdot 12^{k-1} \cdot d.
  \]\\
  Since \(d\) is even, \(M_2\) is even. We now check whether \(M_2\) is
  divisible by \(12\).

  \begin{itemize}
  \tightlist
  \item
    If \(k-1 \ge 1\) (i.e., \(k \ge 2\)), then \(12^{k-1}\) already
    contains a factor \(12\), so \(M_2\) is divisible by \(12\). But
    then we can continue the process. However, after \textbf{exactly
    \(k\)} steps we will obtain\\
    \[
    M_{k+1} = 13^k \cdot d,
    \]\\
    which is even (because \(d\) is even). Since \(d\) is not divisible
    by \(12\) (by the definition of \(k\)), \(M_{k+1}\) is \textbf{not}
    divisible by \(12\). Hence at that moment we would have an even
    \(M\) that is not divisible by \(12\), which forces \(f(6M_{k+1})\)
    not to be a multiple of \(6\) - contradiction.\\
  \item
    If \(k = 1\), then \(M_1 = 12 \cdot d\) with \(d\) even. Then
    \(M_2 = 13 d\) is even but not divisible by \(12\) (because \(13d\)
    lacks the necessary factor \(3\) \textcolor{annotlabel}{and}\annotmark{C6}{\annotnote{C6}{}{or}} the factor \(4\) if \(d\) is
    not a multiple of \(12\)). Again, \(M_2\) fails the requirement that
    every even \(M\) must be divisible by \(12\).
  \end{itemize}

  In all subcases, contradiction arises.
\item
  \textbf{\(d\) is odd but \(5 \mid d\).}\\
  Then after \(k\) applications of the even step we obtain\\
  \[
  M_{k+1} = 13^k \cdot d,
  \]\\
  which is odd and divisible by \(5\). Consequently,
  \(a_{k+1} = 6M_{k+1}\) falls into case 2 of \S{}4, yielding
  \(f(a_{k+1})\) odd. As in \S{}3, an odd term would lead to an infinite
  strictly decreasing sequence of odd numbers - contradiction.
\end{itemize}

Thus, the only way to avoid contradictions is that \(d\) is odd and
\(5 \nmid d\).

\begin{center}\rule{0.5\linewidth}{0.5pt}\end{center}

\subsubsection*{6. Final answer}

Combining everything, we obtain that the initial value \(a_1\) can be
exactly those numbers of the form

\[
\boxed{a_1 = 6 \cdot 12^{k} \cdot d \qquad (k \in \mathbb{N}_0,\; d\ \text{odd},\; 5 \nmid d).}
\]

Equivalently, \(a_1 = 2^{2k+1} \cdot 3^{k+1} \cdot d\) with \(d\) odd
and not divisible by \(5\).

\begin{center}\rule{0.5\linewidth}{0.5pt}\end{center}
\end{compactproof}

\begin{humanexpertcomment}
\[\boxed{7/7}\]

\end{humanexpertcomment}
\clearpage
\phantomsection

\markboth{Problem 5}{Problem 5}
        \begin{problemstatementbox}{Problem 5}
        Alice and Bazza are playing the inekoalaty game, a two-player game whose
rules depend on a positive real number \(\lambda\) which is known to
both players. On the \(n\)th turn of the game (starting with \(n = 1\))
the following happens:

\begin{itemize}
\item
  If \(n\) is odd, Alice chooses a nonnegative real number \(x_n\) such
  that\\
  \[
  x_1 + x_2 + \cdots + x_n \leq \lambda n.
  \]
\item
  If \(n\) is even, Bazza chooses a nonnegative real number \(x_n\) such
  that\\
  \[
  x_1^2 + x_2^2 + \cdots + x_n^2 \leq n.
  \]
\end{itemize}

If a player cannot choose a suitable \(x_n\), the game ends and the
other player wins. If the game goes on forever, neither player wins. All
chosen numbers are known to both players.

Determine all values of \(\lambda\) for which Alice has a winning
strategy and all those for which Bazza has a winning strategy.
        \tcblower
        \textbf{Short Answer}\par
        The outcome of the inekoalty game depends on the value of \(\lambda\) as
follows:

\begin{itemize}
\tightlist
\item
  If \(\lambda > \frac{\sqrt{2}}{2}\), Alice has a winning strategy.
\item
  If \(\lambda < \frac{\sqrt{2}}{2}\), Bazza has a winning strategy.
\item
  If \(\lambda = \frac{\sqrt{2}}{2}\), neither player has a winning
  strategy, and the game can go on forever.
\end{itemize}
        \end{problemstatementbox}

        \Needspace{10\baselineskip}
        \textbf{Model Proof}\\
        \begin{compactproof}
        \textbf{Notation and basic facts}

Define \(S_n = \sum_{i=1}^n x_i\) and \(Q_n = \sum_{i=1}^n x_i^2\).\\
The rules: on odd \(n\), \(S_n \le \lambda n\); on even \(n\),
\(Q_n \le n\).

From Cauchy-Schwarz, after any even turn \(2k\) we have\\
\[
S_{2k}^2 \le (2k) Q_{2k} \le (2k)\cdot 2k = (2k)^2 \quad\Longrightarrow\quad S_{2k} \le 2k. \tag{1}
\]

For \(k\ge 1\) define the \textbf{slack} after the even turn \(2k\):\\
\[
d_k = \lambda(2k+1) - S_{2k}.
\]\\
Alice can move on turn \(2k+1\) iff \(d_k \ge 0\) (she may choose
\(x_{2k+1}=0\)).

If at some odd turn Alice can legally choose a number \(u > \sqrt{2}\)
while \(Q_{2k}=2k\), then\\
\(Q_{2k+1}=2k+u^2 > 2k+2\) and Bazza will have no legal move on the next
turn, so Alice wins immediately.\\
In the general case, we will use the slack to analyse the game.

\textbf{Key strategy for Bazza}

We consider the following natural strategy for Bazza on even turns:
after Alice's move \(u\) on turn \(2k+1\), if the game has not ended,
Bazza chooses\\
\[
v = \sqrt{2k+2 - Q_{2k+1}},
\]\\
i.e.~the largest possible number that still satisfies
\(Q_{2k+2}\le 2k+2\).\\
This choice is always legal as long as \(Q_{2k+1}\le 2k+2\) and makes
\(Q_{2k+2}=2k+2\) (the maximal possible sum of squares).\\
We call this the \textbf{maximal strategy} for Bazza.

\emph{Claim 1.} As long as Alice never picks a number \(u > \sqrt{2}\)
(which would make her win immediately), the maximal strategy forces
\(Q_{2k}=2k\) for every \(k\ge 1\).

\emph{Proof.} By induction. For \(k=1\), after Alice's first move
\(x_1=a\) (with \(0\le a\le\lambda\)), Bazza can choose
\(v=\sqrt{2-a^2}\) because \(a^2\le 2\) (this holds for all relevant
\(\lambda\); in the cases where we apply it, we have
\(\lambda\le\sqrt2/2<\sqrt2\), and for larger \(\lambda\) Alice would
win earlier). Then \(Q_2 = a^2+v^2 = 2\). Inductive step: assume
\(Q_{2k}=2k\). \textcolor{annotlabel}{Alice's move \(u\) satisfies \(Q_{2k+1}=2k+u^2\le 2k+2\)
because otherwise \(u^2>2\) and Alice would have won (this can only
happen if \(Q_{2k}=2k\) and \(u>\sqrt2\)).}{\annotmark{D1}{Do not understand. Reads like a thinking draft process.}{}} If the game continues, we
must have \(u\le\sqrt2\). Then \(Q_{2k+1}=2k+u^2\le 2k+2\). Bazza
chooses \(v=\sqrt{2k+2-(2k+u^2)}=\sqrt{2-u^2}\), yielding
\(Q_{2k+2}=2k+2\). \(\square\)

\textbf{Analysis of the slack under the maximal strategy}

Suppose the maximal strategy is followed and that the game has not ended
(so far Alice has never picked \(u>\sqrt2\)). Let \(u=x_{2k+1}\) and
\(v=x_{2k+2}\). Then\\
\[
S_{2k+2}=S_{2k}+u+v,\qquad Q_{2k+2}=2k+2,
\]\\
and therefore\\
\[
d_{k+1} = \lambda(2k+3)-S_{2k+2} = d_k + 2\lambda - (u+v). \tag{2}
\]

For \(0\le u\le\sqrt2\) define \(h(u)=u+\sqrt{2-u^2}\). One checks that
\(h(u)\ge\sqrt2\), with equality exactly at \(u=0\) and \(u=\sqrt2\),
and maximum \(2\) at \(u=1\). Consequently\\
\[
d_{k+1} \le d_k + 2\lambda - \sqrt2, \tag{3}
\]\\
and equality can be achieved by choosing \(u=0\) (or \(u=\sqrt2\), if
allowed).\\
Thus the maximal possible increase of \(d_k\) (when Alice tries to keep
her slack large) is \(2\lambda-\sqrt2\), attained by taking \(u=0\).

\textbf{Three regimes}

\begin{center}\rule{0.5\linewidth}{0.5pt}\end{center}

\textbf{1.
\(\lambda < \dfrac{\sqrt2}{2}\)}

Then \(2\lambda-\sqrt2 < 0\).

\emph{Bazza's winning strategy.}\\
Bazza adopts the maximal strategy described above.

First, compute an upper bound for \(d_1\). After the first two moves,\\
\(S_2 = a + \sqrt{2-a^2}\) with \textcolor{annotlabel}{\(0\le a\le\lambda\)}\annotmark{D2}{\annotnote{D2}{}{Better to use $x_1$ not $a$}}. The function
\(f(a)=a+\sqrt{2-a^2}\) satisfies \(f(a)\ge\sqrt2\) (minimum at
\(a=0\)), so\\
\[
d_1 = 3\lambda - S_2 \le 3\lambda - \sqrt2. \tag{4}
\]\\
Because \(\lambda<\sqrt2/2\), we have
\(3\lambda-\sqrt2 < \sqrt2/2 < \sqrt2\).

Now, from (3) we obtain for any possible play\\
\[
d_k \le d_1 + (k-1)(2\lambda-\sqrt2) \le (3\lambda-\sqrt2) + (k-1)(2\lambda-\sqrt2). \tag{5}
\]\\
(The first inequality follows by induction using
\(d_{k+1}\le d_k+2\lambda-\sqrt2\), the second uses the bound on
\(d_1\).)\\
Since \(2\lambda-\sqrt2<0\), the right-hand side of (5) tends to
\(-\infty\) as \(k\to\infty\). Hence there exists a finite \(K\) with
\(d_K < 0\).

When \(d_K<0\), at the beginning of turn \(2K+1\) Alice cannot move
(even \(x=0\) would violate the sum condition), so Bazza wins.

Moreover, we claim that Alice never gets the chance to win earlier by
picking \(u>\sqrt2\). Indeed, from (5) and \(2\lambda-\sqrt2<0\), we have
\(d_k \le 3\lambda-\sqrt2 < \sqrt2\) for all \(k\). Hence \(d_k\) never
exceeds \(\sqrt2\), so she cannot legally pick a number larger than
\(\sqrt2\) (which would be required to win immediately). Thus Bazza's
strategy is winning.

\textbf{Conclusion for \(\lambda<\sqrt2/2\):} Bazza has a winning
strategy.

\begin{center}\rule{0.5\linewidth}{0.5pt}\end{center}

\textbf{2.
\(\lambda = \dfrac{\sqrt2}{2}\)}

Here \(2\lambda-\sqrt2 = 0\).

\emph{Bazza does not have a winning strategy.}\\
We exhibit a strategy for Alice that prevents Bazza from ever winning.\\
Let Alice always choose \(x_1=0\) and thereafter \(x_{2k+1}=0\) on every
odd turn.\\
Consider any Bazza moves. On even turns we have only Bazza's numbers,
say \(y_1,\dots,y_k\) at turn \(2k\). The constraints are
\(\sum_{i=1}^k y_i^2 = Q_{2k} \le 2k\) (because odd terms are zero). By
Cauchy-Schwarz,\\
\[
S_{2k} = \sum_{i=1}^k y_i \le \sqrt{k\sum_{i=1}^k y_i^2} \le \sqrt{k\cdot 2k} = \sqrt2\,k.
\]\\
Therefore\\
\[
d_k = \lambda(2k+1) - S_{2k} \ge \frac{\sqrt2}{2}(2k+1) - \sqrt2\,k = \frac{\sqrt2}{2} > 0.
\]\\
Thus Alice can always move (she may choose \(0\)). \textcolor{annotlabel}{Also, she never wins
because to win she would need to make \(Q_{2k+1}>2k+2\), which would
require \(x_{2k+1}^2 > 2k+2 - Q_{2k} \ge 2\),
i.e.~\(x_{2k+1} > \sqrt2\). But \(x_{2k+1} \le d_k\) and we have not
shown an upper bound on \(d_k\)};\annotmark{D4}{\annotnote{D4}{}{This conclusion only focuses on a specific $B$'s policy, which already discussed at the beginning. This looks like thinking draft process.}} however, the crucial point is that with
this Alice strategy, Bazza never wins because Alice never loses. So this
suffices to prove that Bazza does not have a winning strategy (he cannot
force a win against every Alice strategy).

\emph{Alice does not have a winning strategy.}\\
We now show that Bazza has a strategy to prevent Alice from winning.\\
Let Bazza adopt the maximal strategy (as in Case 1). We prove that under
this strategy, Alice can never win.

First, by Claim 1, as long as the game continues, \(Q_{2k}=2k\). Using
(3) with \(\lambda=\sqrt2/2\) we get\\
\[
d_{k+1} = d_k + \sqrt2 - (u+v) \le d_k, \tag{6}
\]\\
because \(u+v\ge\sqrt2\). Hence \(d_k\) is non-increasing. Moreover,
from (4) we have\\
\[
d_1 \le 3\lambda - \sqrt2 = \sqrt2/2.
\]\\
Thus \(d_k \le \sqrt2/2 < \sqrt2\) for all \(k\). In particular, \(d_k\)
never exceeds \(\sqrt2\), so Alice can never pick a number \(u>\sqrt2\)
(which would be necessary to win immediately).

\textcolor{annotlabel}{Could Alice win by picking some \(u\le\sqrt2\) and later build up? To
win, she must at some odd turn make \(Q_{2k+1}>2k+2\). Since
\(Q_{2k}=2k\), this would require \(u^2>2\), i.e.~\(u>\sqrt2\). Thus any
winning move would necessarily be \(u>\sqrt2\). Because
\(d_k\le\sqrt2/2<\sqrt2\), such a move is impossible. Therefore Alice
never wins.}\annotmark{D5}{\annotnote{D5}{}{??? Reads like a thinking draft process.}}

\textcolor{annotlabel}{
Additionally, we check that \(d_k\) never becomes negative, so Alice
always can move. Indeed, using (2) and the fact that \(u\le d_k\) (so
\(u\le\sqrt2/2<1\)), the smallest possible \(d_{k+1}\) for a given
\(d_k\) is attained when \(u=d_k\), giving
\(d_{k+1}=\sqrt2-\sqrt{2-d_k^2}\ge 0\). Hence \(d_k\ge 0\) for all
\(k\).}\annotmark{D6}{\annotnote{D6}{}{No need. The proof is finished.}}

Thus the maximal strategy by Bazza prevents Alice from winning, and also
avoids his own loss (the game can continue indefinitely, e.g., if Alice
always chooses \(0\), then \(d_k=\sqrt2/2\) constant and \(Q_{2k}=2k\),
so the play never ends). Consequently, Alice cannot force a win; she has
no winning strategy.

Since neither player has a strategy that forces a win, the game is a
draw for \(\lambda=\sqrt2/2\).

\begin{center}\rule{0.5\linewidth}{0.5pt}\end{center}

\textbf{3.
\(\lambda > \dfrac{\sqrt2}{2}\)}

Now \(2\lambda-\sqrt2 > 0\).

\emph{Alice's winning strategy.}\\
She will always choose \(0\) on her odd turns, until she can force a
win.\\
Specifically, on turn \(1\) she sets \(x_1=0\). On each subsequent odd
turn \(2k+1\) (\(k\ge 1\)), she examines the current slack
\(d_k = \lambda(2k+1)-S_{2k}\). If \(d_k > \sqrt{2k+2 - Q_{2k}}\), she
chooses a number \(u\) with\\
\[
\sqrt{2k+2 - Q_{2k}} < u \le d_k,
\]\\
such that \(u^2 > 2k+2 - Q_{2k}\) (this is possible because \(d_k\) is
strictly larger). She then wins on the next turn because
\(Q_{2k+1} = Q_{2k}+u^2 > 2k+2\). If \(d_k \le \sqrt{2k+2 - Q_{2k}}\),
she simply chooses \(x_{2k+1}=0\).

We must verify that this strategy is legal and that it indeed leads to a
win.

First, note that as long as Alice always picks \(0\), all her moves
contribute nothing to \(Q\) and \(S_{2k}\) is the sum of Bazza's even
moves. For any Bazza strategy we have\\
\[
Q_{2k} = \sum_{i=1}^k x_{2i}^2 \le 2k, \qquad
S_{2k} = \sum_{i=1}^k x_{2i}.
\]\\
By Cauchy-Schwarz,\\
\[
S_{2k}^2 \le k\, Q_{2k} \le 2k^2 \quad\Longrightarrow\quad S_{2k} \le \sqrt2\,k. \tag{7}
\]\\
Consequently,\\
\[
d_k = \lambda(2k+1) - S_{2k} \ge \lambda(2k+1) - \sqrt2\,k = (2\lambda-\sqrt2)k + \lambda. \tag{8}
\]\\
Because \(2\lambda-\sqrt2>0\), the right-hand side of (8) tends to
\(+\infty\) as \(k\to\infty\). In particular, there exists an integer
\(K\) such that\\
\[
d_K > \sqrt{2K+2 - Q_{2K}}.
\]\\
(Since the right side is at most \(\sqrt{2K+2}\), it suffices that
\((2\lambda-\sqrt2)K + \lambda > \sqrt{2K+2}\); this holds for all
sufficiently large \(K\).)

At the start of turn \(2K+1\), we have \(d_K > \sqrt{2K+2 - Q_{2K}}\).
Choose\\
\[
u = \sqrt{2K+2 - Q_{2K}} + \varepsilon,
\]\\
with \(\varepsilon > 0\) small enough that \(u \le d_K\) (possible
because \(d_K\) is strictly larger). Then\\
\[
S_{2K+1} = S_{2K} + u \le S_{2K} + d_K = \lambda(2K+1),
\]\\
so the sum condition is satisfied, and\\
\[
Q_{2K+1} = Q_{2K} + u^2 > Q_{2K} + (2K+2 - Q_{2K}) = 2K+2.
\]\\
Therefore on turn \(2K+2\) Bazza has no legal move (even \(x=0\) would
give \(Q>2K+2\)). Alice wins.

It remains to check that Alice can always make the choice \(x_{2k+1}=0\)
when \(d_k \le \sqrt{2k+2 - Q_{2k}}\). The condition for choosing \(0\)
is that \(0 \le d_k\), i.e.~\(d_k \ge 0\). From (8),
\(d_k \ge (2\lambda-\sqrt2)k + \lambda > 0\) for all \(k\) (since the
right side is positive for \(k=1\) because \(3\lambda-\sqrt2>0\) for
\(\lambda>\sqrt2/2\)). Hence Alice can always pick \(0\) when she does
not want to win yet. Thus her strategy is well defined and guarantees a
win in finite time.

\textbf{Conclusion for \(\lambda>\sqrt2/2\):} Alice has a winning
strategy.

\begin{center}\rule{0.5\linewidth}{0.5pt}\end{center}

\textbf{Final answer}

\begin{itemize}
\tightlist
\item
  Alice has a winning strategy \textbf{iff}
  \(\lambda > \dfrac{\sqrt2}{2}\).\\
\item
  Bazza has a winning strategy \textbf{iff}
  \(\lambda < \dfrac{\sqrt2}{2}\).\\
\item
  For \(\lambda = \dfrac{\sqrt2}{2}\) neither player can force a win;
  the game can be drawn.
\end{itemize}

\begin{center}\rule{0.5\linewidth}{0.5pt}\end{center}
        \end{compactproof}

\begin{humanexpertcomment}
Some parts of the proof read like a thinking process. It feels like it realizes that c has not been proved, then decides that proving c is not actually necessary. The proof seems finished, but then it starts thinking again.

\[\boxed{7/7}\]

\end{humanexpertcomment}


\restoregeometry

\endgroup

\clearpage
\setcitestyle{numbers}
\bibliographystyle{plainnat}
\bibliography{paper}

\end{document}